%% file: main.tex
\documentclass{article}

\usepackage{PRIMEarxiv}
  
\usepackage[utf8]{inputenc} 
\usepackage[T1]{fontenc}    
\usepackage{hyperref}       
\usepackage{url}            
\usepackage{booktabs}       
\usepackage{longtable}      
\usepackage{amsfonts, amsmath, amssymb}       
\usepackage{nicefrac}       
\usepackage{microtype}      
\usepackage{lipsum}
\usepackage{fancyhdr}       
\usepackage{graphicx}       
\usepackage{subcaption}     
\usepackage{xcolor}         
\usepackage{algorithm}      
\usepackage{algpseudocode}  
\usepackage[most]{tcolorbox}  
\graphicspath{{media/}{figures/}}     

\definecolor{problemcolor}{HTML}{ffb000}
\definecolor{requirementscolor}{HTML}{fe6100}
\definecolor{evaluationcolor}{HTML}{648fff}
\definecolor{datacolor}{HTML}{dc267f}

\title{
AgenticSciML: Collaborative Multi-Agent Systems for Emergent Discovery in Scientific Machine Learning
}

\author{
  Qile Jiang \vspace{1mm} \\
  Division of Applied Mathematics \\
  Brown University  \\
  Providence, RI, USA \\
  \texttt{qile\_jiang@brown.edu} \\
    \And
  George Karniadakis \thanks{Corresponding author: \texttt{george\_karniadakis@brown.edu}}  \vspace{1mm} \\
  Division of Applied Mathematics \\
  Brown University  \\
  Providence, RI, USA \\
  \texttt{george\_karniadakis@brown.edu} 
}

\begin{document}
\maketitle

\begin{abstract}
Scientific Machine Learning (SciML) integrates data-driven inference with physical modeling to solve complex problems in science and engineering. However, the design of SciML architectures, loss formulations, and training strategies remains an expert-driven research process, requiring extensive experimentation and problem-specific insights. Here we introduce AgenticSciML, a collaborative multi-agent system in which over 10 specialized AI agents collaborate to propose, critique, and refine SciML solutions through structured reasoning and iterative evolution. The framework integrates structured debate, retrieval-augmented method memory, and ensemble-guided evolutionary search, enabling the agents to generate and assess new hypotheses about architectures and optimization procedures. Across physics-informed learning and operator learning tasks, the framework discovers solution methods that outperform single-agent and human-designed baselines by up to four orders of magnitude in error reduction. The agents produce novel strategies—including adaptive mixture-of-expert architectures, decomposition-based PINNs, and physics-informed operator learning models—that do not appear explicitly in the curated knowledge base. These results show that collaborative reasoning among AI agents can yield emergent methodological innovation, suggesting a path toward scalable, transparent, and autonomous discovery in scientific computing.

\end{abstract}


\section*{Significance Statement}
Scientific machine learning (SciML) has transformed modeling of physical and biological systems, but  constructing effective SciML models still requires substantial expert effort in choosing architectures, enforcing physical constraints, balancing losses, and diagnosing failure modes. This paper introduces AgenticSciML, a collaborative multi-agent reasoning framework that automates these steps through structured proposal, critique, refinement, and evaluation. Unlike AutoML, NAS, or symbolic regression systems, AgenticSciML does not search within a fixed hypothesis class, but discovers new modeling strategies through debate among 10 domain-specialized agents supported by a persistent knowledge base. Across PDE-constrained learning and inverse problems, the framework yields models and training procedures that were not present in the knowledge base nor in standard SciML formulations, demonstrating the potential for emergent model-form innovation. 

\include{Introduction}

\section{Methods}\label{sec:methods}

\begin{figure}[h!]
    \centering
    \includegraphics[width=0.98\linewidth]{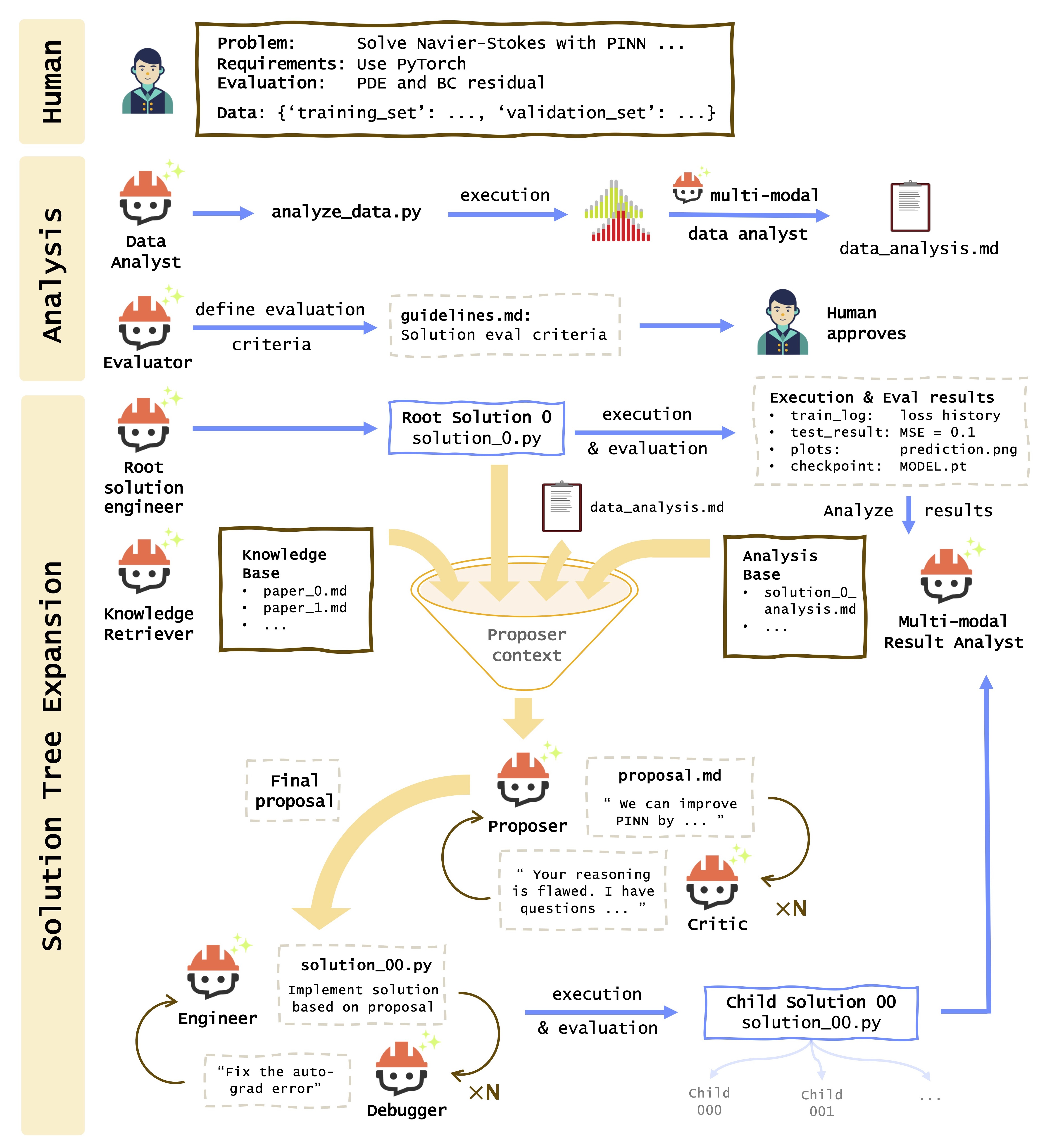}
    \caption{Framework for the evolutionary multi-agent SciML system. \textbf{Phase 1}: A Human user provides inputs including problem, requirements, evaluation criteria, and data. \textbf{Phase 2}: Agents analyze user inputs and define solution evaluation criteria, with human approval. \textbf{Phase 3}: Specialized agents (Knowledge Retrievers, Proposers, Critics, Engineers, and Debuggers) collaboratively propose, implement, and execute new solutions. The Result Analyst agent evaluates each solution, which then becomes eligible for mutation in the next iteration.}
    \label{fig:framework}
\end{figure}

Our evolutionary multi-agent system operates in three phases (Figure~\ref{fig:framework}): user input, data analysis and evaluation criteria creation, and solution evolution. A team of 10+ agents is deployed with specialized roles. Key steps are detailed below. 

\subsection*{Structured User Input}

To start, the human user provides structured input written in Markdown and JSON formats to the agentic system: (1) \texttt{Problem.md}: a statement of the problem to solve; (2) \texttt{Requirements.md}: constraints on implementation (frameworks, libraries, hardware); (3) \texttt{Evaluation.md}: metrics and procedures for scoring solutions; (4) \texttt{Data\_config.json} (optional): training or validation datasets with descriptions, file paths, and loading instructions. 

The problem and requirement statements are to provide context. The evaluation criteria define what success looks like for this problem, such as a low mean squared error on a validation set or low PDE residuals. Datasets are optional, and the user can choose to provide none, some, or all data needed for training or validation. Agents will use user-provided data when available, but can also generate synthetic data if needed. The user provides minimal natural language prompts, and the agents will handle all technical details. See Appendix \ref{sec:appendix_problems} for examples.

\subsection*{Exploratory Data Analysis}

If training data is provided, a multi-modal (text and vision) data analyst agent autonomously generates Python code to perform exploratory data analysis on the training dataset. It is prompted to quantitatively and visually examine both mathematical properties (discontinuities, sharp gradients, multi-scale phenomena, etc.) and data quality (outliers, distribution characteristics, etc.). The agent then analyzes the generated plots using its vision capabilities and produces a text-only report (\texttt{data\_analysis.md}) summarizing key findings, so that downstream text-only agents can always reference this analysis. 

This step is important because human scientists often start by visualizing and analyzing data to gain intuitions. The data analyst agent mimics this process and provides such data intuition to all downstream agents. Note that no analysis is performed on validation data to avoid data leakage.

\subsection*{Evaluation Contract Generation}

Before proceeding to solution generation, an evaluator agent formalizes an evaluation contract that all candidate solutions must satisfy. The goal is to ensure consistent evaluation across diverse solutions and produces a scalar score for ranking. Note that throughout this work, the score is usually defined as a loss to minimize such as mean squared error (MSE) or relative $L_2$ error, so lower is better.

The evaluator agent generates two components that define the solution interface. First, \texttt{evaluate.py} is a complete test script that loads existing model checkpoints, generates or loads validation data, computes metrics, and returns a score. Second, \texttt{guidelines.md} outlines some implementation details for the engineer agent to follow, such as model checkpoint formats and expected input-output data shapes, so that all engineered models can be evaluated through the same procedure. Finally, since the human defines the success criteria, human approval is sought before the plan is set.

\subsection*{Root Solution Generation}

To start the solution tree expansion, a root solution is generated by a single-agent root engineer. 

The root engineer has access to all the user inputs and evaluation guidelines. It produces a baseline solution, named \texttt{solution\_0}. The solution is executed and evaluated autonomously. The root solution serves as our \textit{single-agent baseline}: it is generated by a single LLM call with no access to the knowledge base, no multi-agent debate or critique, and no prior solution context. This design choice isolates the contribution of multi-agent collaboration, knowledge retrieval, and evolutionary search from the raw capability of the underlying single-agent LLM. The results, including training logs and any generated plots, are analyzed by a multi-modal result analyst agent, which writes a detailed textual report (\texttt{analysis\_0.md}) saved to the Analysis Base for future reference.

\subsection*{Ensemble-Guided Parent Selection}

During each iteration of tree expansion, several existing solutions are selected from the solution tree to serve as parents for mutation. Each parent will produce one child solution through the mutation process described later. Multiple parents are mutated in parallel to generate multiple children per iteration.

\begin{figure}[h!]
    \centering
    \includegraphics[width=0.75\linewidth]{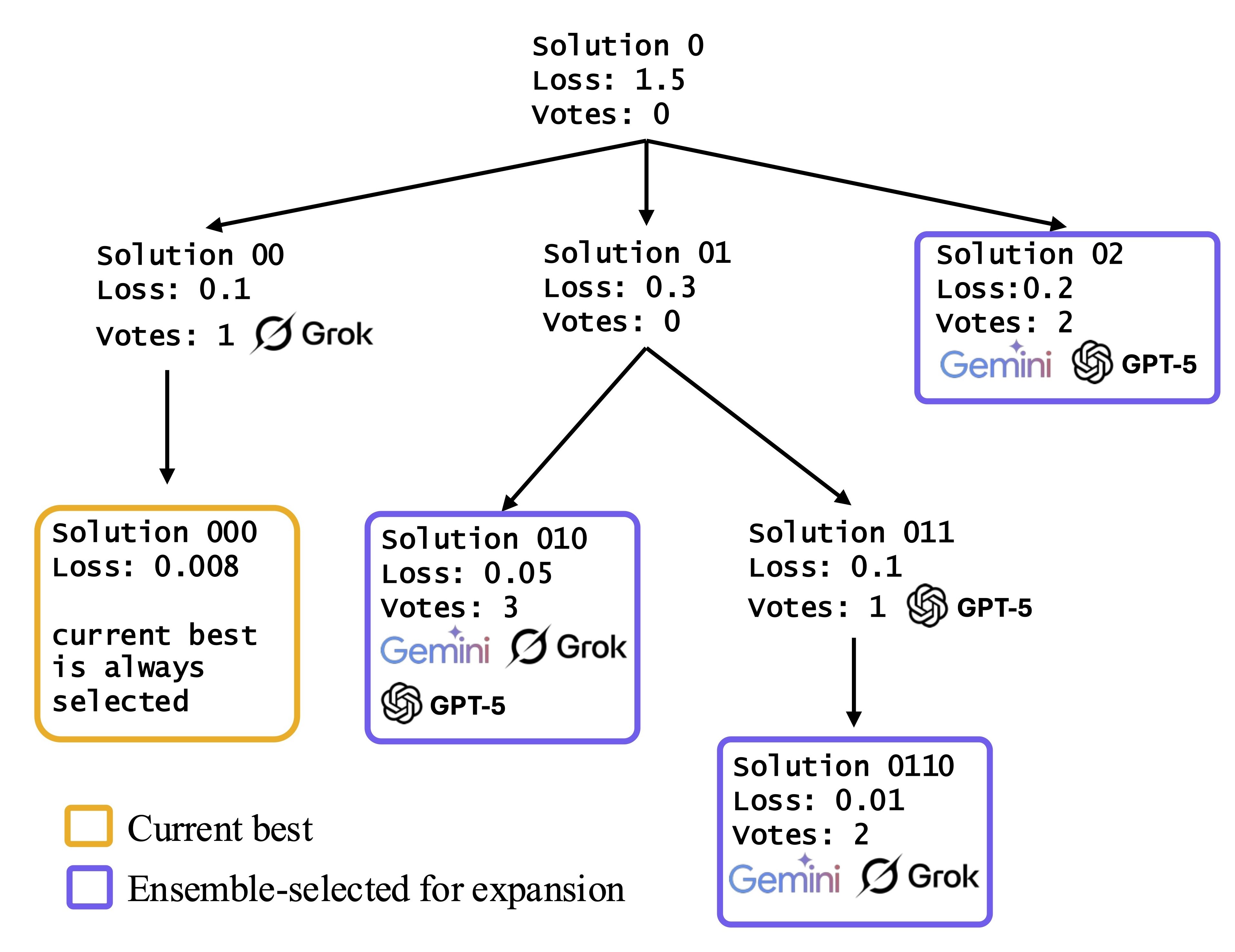}
    \caption{Example showing the selection mechanism for solution tree expansion. The best performing solution on the tree is always selected for mutation in the next iteration (exploitation). Additional solutions are selected based on multi-agent majority voting (exploration). This balances the search between refining the best-known solution and exploring alternative approaches that are deemed promising by the agent ensemble. All selections and mutations are performed in parallel.}
    \label{fig:tree_illustration}
\end{figure}

In the early stage (total solutions $\leq$ mutation batch size), all solutions mutate in parallel. In the mature stage (total solutions $>$ mutation batch size), the best-scoring solution is always selected (exploitation), while additional parents are chosen through majority voting by a diverse ensemble of selector agents (exploration). Figure~\ref{fig:tree_illustration} shows an example of the selection mechanism.

The selectors are prompted to holistically evaluate solution code, analysis reports, and scores to select 1-2 solutions that give promising results and 1-2 solutions that are less promising but have potential for improvement. Each selector agent votes independently, and the solutions with the most votes are selected as additional parents. Solutions reaching maximum children per node (10) are excluded from selection to prevent over-exploitation. 

\subsection*{Solution Mutation through Multi-Agent Debate and Engineering}

After parents are selected, child solutions are generated through three stages: planning, engineering, and evaluation. 

\textbf{Planning:} This is the critical stage for scientific reasoning and innovation. First, a retriever agent critically evaluates the parent's solution and result analysis to determine if any entries from a curated Knowledge Base of SciML techniques would be helpful. If so, it retrieves 0-1 entries to share with the proposer and critic agents. More information on the Knowledge Base can be found in Supplementary Materials \ref{sec:kb}.

Next, several reports from the Analysis Base are also retrieved. These include the parent's analysis report, sibling reports (same parent), and uncle reports (grandparent's other children). This provides broader context on what strategies have worked or failed in neighboring solutions.

All these contexts are provided to the proposer and critic agents, who then engage in structured $N$-round debate to formulate a concrete implementation proposal for the child solution. For the first $N-2$ rounds, the proposer is asked to analyze the problem deeply and think out loud. It is \textit{not} asked to propose a new strategy yet -- it focuses solely on reasoning. The critic then challenges the proposer's reasoning, pointing out flaws, gaps, or alternative perspectives. In the $(N-1)$-th round, the proposer starts to synthesize its reasoning into a concrete implementation plan, which the critic evaluates for feasibility and soundness. Finally, in the $N$-th round, the proposer produces a final implementation-ready proposal without further critique.

\textbf{Engineering:} The engineer agent receives the parent's full code, final proposal, and other problem specifications. It implements the exact proposed solution by modifying the parent's codebase and does not need to reason about high-level strategies. Any errors are handled by the debugger agent until successful execution or maximum debugging iteration reached. 

\textbf{Evaluation:} The multi-modal result analyst agent is again employed to analyze the child solution's results, both textual logs and generated plots, and write a detailed report, which is added into the Analysis Base. The score of this solution is also computed for ranking.

\subsubsection*{Iteration}

The above tree mutation process is repeated for a fixed number of iterations or until a stopping criterion is met. At the end, the best-scoring solution on the tree is returned as the champion solution.

\subsection*{Algorithm Summary}

Algorithm~\ref{alg:agenticsciml} consolidates the three-phase procedure. The LLM assigned to each agent role varies by experiment and is summarized in Tables~\ref{tab:func_approx_config}--\ref{tab:cylinder_config}. By default: Gemini 2.5 Pro/Flash serves as Proposer, Retriever, Data Analyst, and Result Analyst; Claude Haiku 4.5 as Root Engineer and Engineer; GPT-5 Mini as Critic and Debugger; and GPT-5 Mini, Grok-4 Fast, and Gemini as the Selector ensemble.

\begin{algorithm}[h!]
\caption{AgenticSciML: Evolutionary Multi-Agent SciML Framework}
\label{alg:agenticsciml}
\begin{algorithmic}[1]
\Require Problem $P$, Requirements $R$, Evaluation criteria $E$, optional Data $D$
\Ensure Champion SciML solution $s^*$

\Statex \textbf{Phase 1: Initialization}
\If{$D$ is provided}
    \State $r_{\text{data}} \gets \textsc{DataAnalyst}(D)$ \Comment{Multimodal EDA; saves text-only report for downstream agents}
\EndIf
\State $(\texttt{eval.py},\, \texttt{guidelines.md}) \gets \textsc{Evaluator}(P, R, E)$ \Comment{Formalize testing contract}
\State Await human approval of $(\texttt{eval.py},\, \texttt{guidelines.md})$ \Comment{Human-in-the-loop checkpoint}

\Statex \textbf{Phase 2: Root Solution}
\State $s_0 \gets \textsc{RootEngineer}(P, R, \texttt{guidelines.md})$ \Comment{Single-agent baseline: no KB, no debate}
\State $\mathrm{score}_0 \gets \textsc{Execute}(s_0,\, \texttt{eval.py})$
\State $a_0 \gets \textsc{ResultAnalyst}(s_0,\, \mathrm{score}_0)$ \Comment{Multimodal analysis report}
\State Initialize tree $\mathcal{T} \gets \{(s_0,\, \mathrm{score}_0,\, a_0)\}$, Analysis Base $\mathcal{A} \gets \{a_0\}$

\Statex \textbf{Phase 3: Evolutionary Tree Expansion}
\For{$t = 1, 2, \ldots, T_{\max}$}
    \State $\mathcal{P} \gets \bigl\{\arg\min_{s \in \mathcal{T}}\,\mathrm{score}(s)\bigr\}$ \Comment{Always include best solution (exploitation)}
    \State $\mathcal{P} \gets \mathcal{P} \cup \textsc{SelectorEnsemble}(\mathcal{T},\, K{-}1)$ \Comment{$K{-}1$ additional parents by majority vote (exploration)}
    \For{each parent $p \in \mathcal{P}$} \Comment{All parents mutated in parallel}
        \State $\mathrm{kb} \gets \textsc{Retriever}(p,\, \mathcal{T},\, \mathrm{KB})$ \Comment{Retrieve 0--1 relevant KB entry based on parent's weaknesses}
        \State $\mathrm{ctx} \gets \mathcal{A}.\text{get}(p)$ \Comment{Retrieve parent, sibling, and uncle analysis reports}
        \For{round $r = 1$ to $N$} \Comment{Structured $N$-round debate ($N=4$ in all experiments)}
            \If{$r \leq N-2$} \Comment{Reasoning rounds: deep analysis without proposing}
                \State $\mathrm{reasoning}_r \gets \textsc{Proposer}(\mathrm{ctx},\, \mathrm{kb},\, \mathrm{critique}_{r-1})$
                \State $\mathrm{critique}_r \gets \textsc{Critic}(\mathrm{reasoning}_r)$ \Comment{Challenge flaws, gaps, and alternatives}
            \ElsIf{$r = N-1$} \Comment{Synthesis round: form concrete implementation plan}
                \State $\mathrm{plan} \gets \textsc{Proposer.Synthesize}(\mathrm{reasoning}_{1 \ldots N-2},\, \mathrm{kb})$
                \State $\mathrm{critique}_{N-1} \gets \textsc{Critic.Evaluate}(\mathrm{plan})$ \Comment{Assess feasibility and soundness}
            \Else \Comment{$r = N$: finalization; no further critique}
                \State $\mathrm{proposal} \gets \textsc{Proposer.Finalize}(\mathrm{plan},\, \mathrm{critique}_{N-1})$
            \EndIf
        \EndFor
        \State $c \gets \textsc{Engineer}(p.\mathrm{code},\, \mathrm{proposal},\, \texttt{guidelines.md})$ \Comment{Implement child by modifying parent code}
        \State \textbf{repeat} $c \gets \textsc{Debugger}(c)$ \textbf{until} no errors \textbf{or} max retries reached
        \State $\mathrm{score}_c \gets \textsc{Execute}(c,\, \texttt{eval.py})$
        \State $a_c \gets \textsc{ResultAnalyst}(c,\, \mathrm{score}_c,\, a_p,\, \mathrm{kb})$ \Comment{Multimodal analysis; added to $\mathcal{A}$}
        \State $\mathcal{T} \gets \mathcal{T} \cup \{(c,\, \mathrm{score}_c,\, a_c)\}$
    \EndFor
\EndFor
\State \Return $s^* = \arg\min_{s \in \mathcal{T}}\,\mathrm{score}(s)$ \Comment{Champion solution}
\end{algorithmic}
\end{algorithm}

\section{Results} \label{sec:results}

To demonstrate the capabilities of our multi-agent SciML system, we evaluate its performance on a challenging set of benchmark problems on physics-informed machine learning and operator learning. The problems include:

\begin{enumerate}
  \item \textbf{Discontinuous Function Approximation (\ref{sec:func_approx}):} Learning a piecewise oscillatory function with discontinuities from data using neural networks.
  \item \textbf{Solving Poisson Equation on L-shaped Domain with PINN (\ref{sec:poisson_l}):} Using Physics-Informed Neural Networks (PINNs) to solve the Poisson equation on an L-shaped domain. 
  \item \textbf{Solving Burger's Equation using PINN (\ref{sec:burger}):} Using PINNs to solve the time-dependent nonlinear Burger's equation with periodic boundary conditions and a challenging initial condition.
  \item \textbf{Antiderivative Operator Learning (\ref{sec:antideriv}):} Learning a neural operator that maps input functions to their antiderivatives, generalizing to unseen functions.
  \item \textbf{Multiple-input Operator Learning for Reaction-Diffusion Equation (\ref{sec:multiop}):} Learning a neural operator that maps multiple input functions (diffusion coefficient and source term) to the solution of a reaction-diffusion equation over time.
  \item \textbf{Reconstruction of 2D Cylinder Wake Vorticity Field from Sparse and Noisy Observations (\ref{sec:cylinder_wake}):} Using neural networks to reconstruct the vorticity field behind a cylinder from only 4 sensors and noisy measurements.
\end{enumerate}

These problems are designed to encourage original thinking from the agents and be solved with novel SciML techniques. Detailed setups and results for each problem are provided in Supplementary Materials. All problems use the same knowledge base and similar agent parameters. Below, we summarize key findings across all experiments.

\subsection{Multi-Agent Collaboration Outperforms Single-Agent Baselines}

\begin{figure}[h!]
\centering
\includegraphics[width=0.8\linewidth]{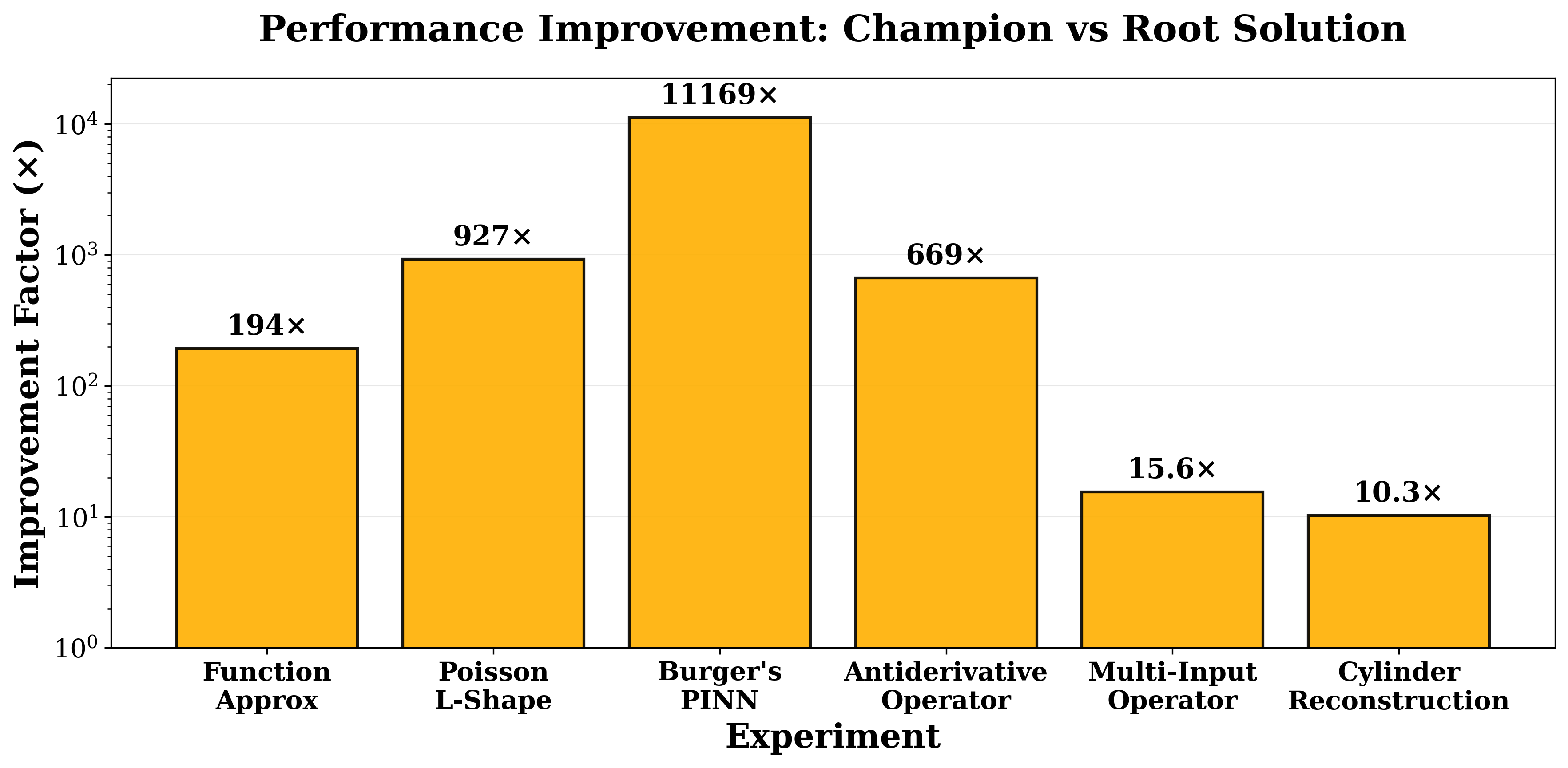}
\caption{Performance improvements achieved by the multi-agent system over single-agent baselines across all benchmark problems. The y-axis shows the improvement factor (root score / champion score) on a logarithmic scale. In every case, the multi-agent system discovers solutions that outperform the single-agent baseline, with improvement factors ranging from 10$\times$ to over 11,000$\times$. Each bar corresponds to one of the six benchmark problems (x-axis). The improvement factor is defined as the ratio of the root solution score to the champion solution score; a higher value indicates greater benefit from multi-agent collaboration, knowledge retrieval, and evolutionary search over the single-agent baseline.}
\label{fig:performance_improvements}
\end{figure}

For all problems, a single root engineer agent generates the root solution as a baseline to improve upon. With today's state-of-the-art LLMs, single-agent systems can already produce reasonably competent solutions for many SciML problems. However, the multi-agent system discovers solutions that significantly outperform the single-agent baseline across all benchmark problems, with improvement factors ranging from 10$\times$ to over 11,000$\times$. Figure~\ref{fig:performance_improvements} summarizes the performance improvements. The improvement factor is measured as the ratio of the root solution score to the champion solution score after multi-agent evolution.

These results critically demonstrate that the intelligence of the multi-agent system emerges from collaborative reasoning, knowledge integration, and evolutionary tree search, rather than solely relying on the capabilities of individual LLMs.

\subsection{Novel Solution Strategies Emerge Through Evolution}

We define a solution strategy as an \textit{emergent discovery} if it is not present in any KB entry but is instead synthesized by the agents through reasoning about retrieved techniques, problem structure, and prior experimental results. The KB provides methodological inspiration; emergence occurs when agents adapt, combine, or derive new approaches that extend beyond what any single KB entry explicitly describes.

We observe that the multi-agent system discovers novel solution strategies that differ significantly from the initial root solution. Even though knowledge base entries are retrieved and considered, the final solutions are not direct copies of existing methods. Instead, agents creatively adapt and combine techniques to devise new approaches tailored to each problem. Some novel strategies for each problem include:

\begin{enumerate}
  \item \textbf{Discontinuous Function Approximation (\ref{sec:func_approx_novelty}):} A Mixture-of-Experts architecture with specialized experts for different branches of the piecewise function, controlled by a learnable gating mechanism. The trainable gating parameter is inspired by the trainable parameter in adaptive activation functions \cite{jagtap2020adaptive}.
  \item \textbf{Poisson Equation on L-shaped Domain (\ref{sec:poisson_novelty}):} Decomposition of the target solution into a known particular solution plus a component learned by a PINN. The solution also uses importance sampling of collocation points to bias towards the corner's singularity.
  \item \textbf{Burger's Equation (\ref{sec:burger_novelty}):} Dividing training into 3 steps: pre-train only on the initial and boundary condition, then train on the gradient-enhanced PDE residual \cite{yu2022gradient} with self-adaptive weights \cite{mcclenny2023self}, and finally fine-tune with residual-based adaptive refinement of collocation points using double-precision L-BFGS. 
  \item \textbf{Antiderivative Operator Learning (\ref{sec:antideriv_novelty}):} A DeepONet with a linear, bias-free branch net to enforce linearity of the antiderivative operator. 
  \item \textbf{Multiple-input Operator Learning (\ref{sec:multiop_novelty}):} 2D FNO trained with a derivative-enhanced loss, inspired by \cite{wen2022u}. 
  \item \textbf{Cylinder Wake Reconstruction (\ref{sec:cylinder_novelty}):} Modified U-FNO architecture \cite{wen2022u} with CNO-inspired \cite{raonic2023convolutional} bandlimit-preserving filters in the decoder to mitigate aliasing during upsampling.
\end{enumerate}

In particular, solutions to Problems 2 and 4 are mainly based on analysis of the underlying mathematical properties of the problem. Solutions to Problems 1, 5, and 6 come from combining and refining existing methods from the literature. Table~\ref{tab:emergent_discovery} maps all KB entries retrieved during each experiment to the emergent strategy of the champion solution and compares what was retrieved versus what was synthesized.

\begin{table}[h!]
\centering
\caption{KB entries retrieved across all evolutionary iterations (deduplicated by reference) and the emergent discovery of the champion solution for each experiment. Strategies are emergent if not present in any retrieved KB entry. See Section~\ref{sec:appendix_problems} for full details.}
\label{tab:emergent_discovery}
\small
\begin{tabular}{p{0.13\linewidth}p{0.43\linewidth}p{0.36\linewidth}}
\toprule
\textbf{Problem} & \textbf{KB Entries Retrieved} & \textbf{Emergent Discovery (Champion)} \\
\midrule
Func.\ Approx. &
  Adaptive Act.~\cite{jagtap2020adaptive}; Local Adaptive Act.~\cite{jagtap2020locally}; MoE-PINN~\cite{bischof2022mixture}; Self-Adaptive PINN~\cite{mcclenny2023self}; gPINN~\cite{yu2022gradient} &
  MoE with learnable sigmoid gating; bounded sharpness $k$ via $\sigma(k_{\text{raw}})$ \\
\midrule
Poisson &
  gPINN~\cite{yu2022gradient}; hp-VPINN (L-shaped)~\cite{kharazmi2021hp}; Self-Adaptive PINN~\cite{mcclenny2023self} &
  Decomposition $u\!=\!u_{nn}\!+\!u_p$; power-law importance sampling at the corner \\
\midrule
Burger's &
  gPINN~\cite{yu2022gradient}; Self-Adaptive PINN~\cite{mcclenny2023self}; MoE-PINN~\cite{bischof2022mixture}; Local Adaptive Act.~\cite{jagtap2020locally} &
  3-phase training: BC/IC pretraining $\to$ gPINN + adaptive weights $\to$ RAR + L-BFGS \\
\midrule
Antiderivative &
  DeepONet~\cite{lu2021learning}; PI-DeepONet~\cite{wang2021learning}; Adaptive Act.~\cite{jagtap2020adaptive}; Self-Adaptive PINN~\cite{mcclenny2023self}; gPINN~\cite{yu2022gradient} &
  Linear bias-free branch net enforcing operator linearity \\
\midrule
Multi-input Op. &
  U-FNO~\cite{wen2022u}; gPINN~\cite{yu2022gradient}; CNO~\cite{raonic2023convolutional} &
  1D inputs expanded into 2D spatiotemporal grids; hard BC/IC enforcement on FNO output \\
\midrule
Cylinder Wake &
  U-FNO~\cite{wen2022u}; CNO~\cite{raonic2023convolutional} &
  CNO-inspired bandlimit-preserving Gaussian filter in U-FNO decoder upsampling \\
\bottomrule
\end{tabular}
\end{table}

\subsection{Ensemble Voting Analysis}

\begin{figure}[h!]
  \centering
  \includegraphics[width=0.95\linewidth]{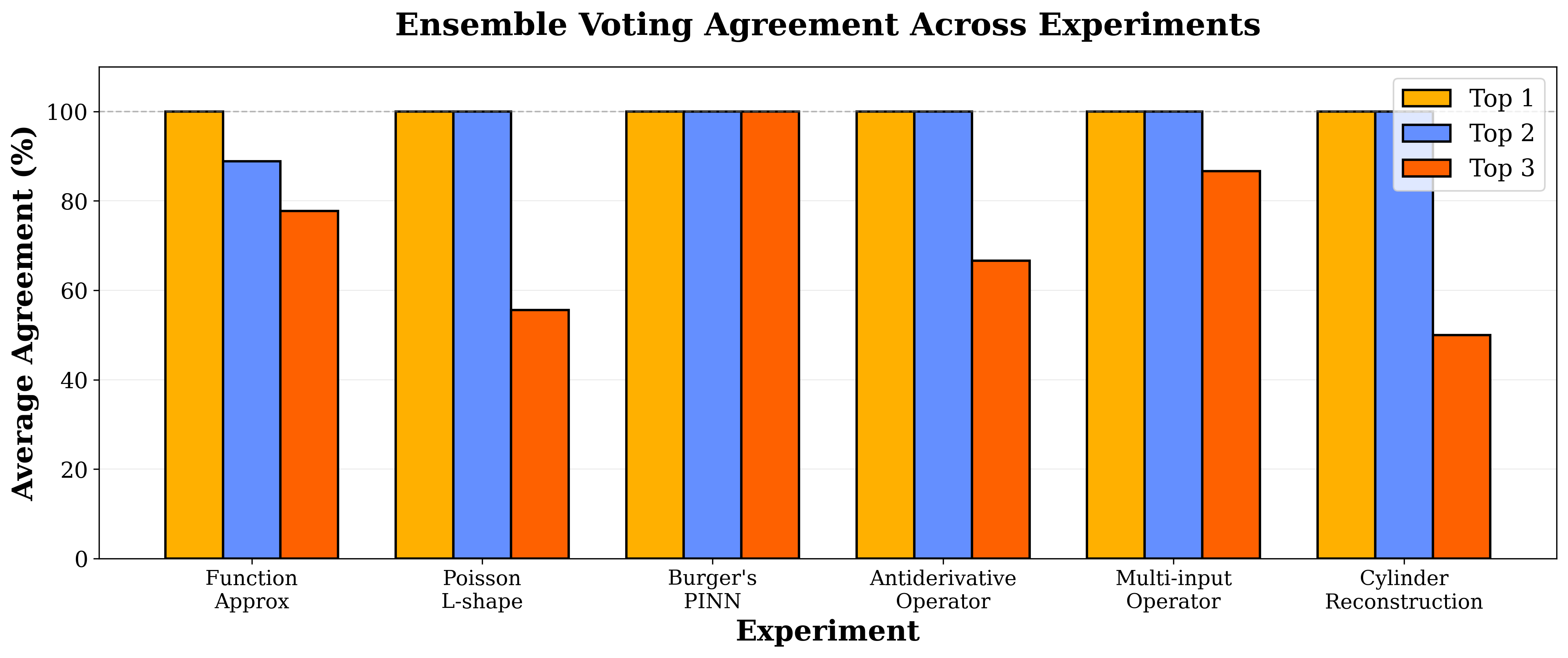}
  \caption{Ensemble voting agreement across all experiments for top 3 selections. The selectors consistently agree on the first choice and mostly on the second, while the third choice shows disagreement, since different selector agents have different opinions on which solutions have the most potential for improvement. The x-axis groups selections by nomination rank (1st, 2nd, 3rd), and the y-axis shows the fraction of experiments in which all three selector models (GPT-5 Mini, Grok-4 Fast, Gemini 2.5 Pro/Flash) cast identical votes. High agreement at rank 1 reflects convergence on the best-performing (exploitation) candidate; lower agreement at rank 3 reflects healthy diversity in exploratory choices.}
  \label{fig:ensemble_agreement}
\end{figure}

The solution tree mutation is guided by an ensemble of selector agents. Throughout this work, Gemini-2.5-pro, GPT-5-mini, and Grok-4-fast are used as the selector ensemble due to their diverse perspectives. In all experiments, each selector nominates three candidate solutions for mutation. Figure \ref{fig:ensemble_agreement} shows the ensemble's voting agreement for the top three nominations across all experiments.

The selectors consistently agree on the first choice and mostly on the second, as these typically correspond to the best-performing solutions favored for exploitation. In contrast, the third selection shows a healthy disagreement between agents. This is because the third selection is usually an exploratory choice, where agents diverge on which solution has the potential for improvement. The voting mechanism thus introduces stochasticity and diversity into the evolutionary tree search.

\subsection{Agent Contribution Analysis}

\begin{figure}[h!]
  \centering
  \includegraphics[width=0.95\linewidth]{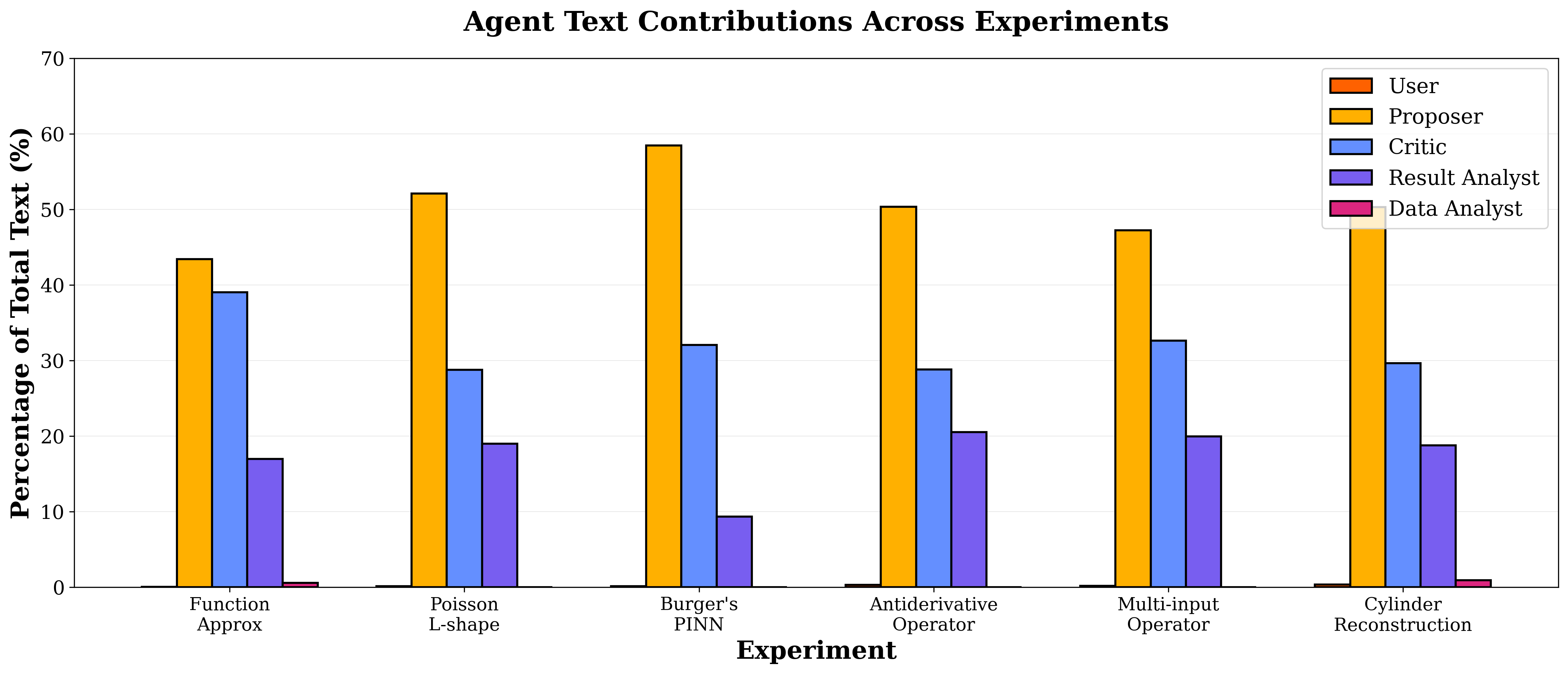}
  \caption{Agent contribution analysis across all experiments by word count. The proposer contributes the most, while the user's contribution is minimal. Each bar group corresponds to one experiment (x-axis), and the stacked colored segments show total word count per agent type (y-axis). The Proposer dominates due to its extended multi-round reasoning steps. Engineering and selector agents are excluded. Across all six experiments, the human user contributes less than 0.3\% of total generated text, demonstrating the system's high degree of autonomy.}
  \label{fig:agent_contributions}
\end{figure}

Figure~\ref{fig:agent_contributions} summarizes the contributions of key planning agents by word count. The contributions from engineers and debuggers are not included. We observe that the proposer agent consistently contributes the most words, since it is instructed to think out loud and write down all intermediate reasoning steps. The human user contributes less than 0.3\% of the total text, showing the system's ability to autonomously handle problems with minimal human input.

\subsection{Token and Cost Analysis}
\label{sec:cost_analysis}

The scalability and efficiency of the proposed framework are demonstrated through the following analysis of the associated computational costs and wall-clock execution time.

Total LLM API cost of one end-to-end experiment ranges from \$2.07 (Burgers' PINN, 3 evolution iterations) to \$11.30 (function approximation, 6 evolution iterations), showing that the multi-agent evolutionary framework is practical at roughly single-digit dollar cost per experiment and readily scalable to larger-scale research projects. Across all six problems, the proposer agent dominates LLM expenditure in most runs (36--50\% of total LLM cost in four of six experiments), since it generates and debates multiple candidate solutions per iteration. The result analyst also incurs a large token cost, owing to multimodal inputs that include training loss curves and plots. In most experiments, GPU training time substantially exceeds LLM API calls' time (e.g., 5.6\,h GPU vs.\ 1.7\,h LLM for the Poisson problem; 10.7\,h vs.\ 2.1\,h for multi-input operator learning), showing that the execution time bottleneck is solution-specific model training rather than multi-agent coordination.

All experiments were run on NVIDIA A6000 GPUs. Detailed per-agent breakdowns are tabulated in the Supplementary section.




\section{Summary and Discussion}
\label{sec:summary}

We introduced \textbf{AgenticSciML}, a collaborative multi-agent framework for constructing scientific machine learning (SciML) models through structured proposal, critique, refinement, and evaluation. In contrast to automated hyperparameter tuning, one-shot prompting, or neural architecture search pipelines, AgenticSciML coordinates distinct reasoning roles among agents and maintains a persistent methodological memory, enabling the system to explore the space of modeling strategies rather than merely parameter configurations. The resulting designs are not limited to variations of standard templates; instead, the system produces new forms of SciML architectures and training procedures that reflect nontrivial modeling decisions, such as adaptive domain decomposition in PINNs, constraint-conditioned branches in operator networks, or dynamically modulated loss schedules based on residual structure. These outcomes are characteristic of expert manual development in scientific computing and indicate that structured agentic collaboration can support emergent model-form innovation.

The approach shares conceptual motivation with recent multi-agent scientific planning frameworks such as the Virtual Lab \cite{swanson2025virtual} and ChemCrow, an agent-based framework for molecular synthesis planning \cite{bran2023chemcrow}, as well as with neural architecture search and evolutionary optimization methods \cite{liu2018darts,real2019regularized}. However, those systems typically optimize within fixed hypothesis classes or focus on experimental decision-making, whereas SciML model design requires reasoning across architectural choices, physical constraints, solver behavior, training dynamics, and numerical stability. Similarly, symbolic regression methods \cite{cranmer2020discovering,udrescu2020ai} recover interpretable expressions but do not construct PDE-constrained learning architectures. Recent automated PINN and operator-learning frameworks \cite{zhang2023auto} streamline training workflows but do not modify the underlying modeling paradigm. AgenticSciML differs in aiming for the \emph{construction} of new solution strategies rather than the optimization of predefined ones, with collaborative critique and retrieval enabling iterative refinement informed by prior results.

We also note some current limitations of this work. The quality of emergent solutions depends in part on the structure and coverage of the knowledge base, and improving retrieval grounding remains an important direction. Debate and justification among agents are mediated by LLMs and thus may not always reflect physically rigorous reasoning unless paired with numerical verification signals. Furthermore, the evolutionary search layer introduces computational overhead proportional to evaluation cost, suggesting that future work should prioritize tighter integration with differentiable solvers and low-fidelity model proxies. Extending the framework to multi-physics systems, turbulence, data assimilation, and real experimental workflows will also require adaptation in both evaluation procedures and memory representation.

Future directions include incorporating classic solvers, e.g., finite element and spectral methods, as well as hybrid methods based on PINNs or neural operators combined with classical methods. This will provide scalability and improved accuracy.
In addition, we plan to develop stronger \emph{physics-grounded reasoning signals}, such as adjoint-based consistency checks or solver-in-the-loop validation; exploring \emph{hierarchical agent coordination} in which meta-agents learn to orchestrate debate and search strategies; and formalizing the relationship between collaborative agentic dynamics and the emergence of new SciML strategies. More broadly, our results suggest that multi-agent reasoning systems offer a path toward scalable exploration of the combinatorial solution spaces characteristic of modern scientific computing that includes both classical solvers and neural PDEs and operators. 

\section*{Code Availability}

Code and all experiment results will be available on GitHub \url{https://github.com/Qile-J/AgenticSciML}.

\section*{Funding}
This work was supported by the Vannevar Bush Faculty Fellowship award from ONR (N00014-22-1-2795). The funder played no role in study design, data collection, analysis and interpretation of data, or the writing of this manuscript.


\clearpage
\newpage
\section*{Supplementary Materials}
\addcontentsline{toc}{section}{Supplementary Materials}
\renewcommand{\thesection}{S\arabic{section}}
\renewcommand{\thefigure}{S\arabic{figure}}
\renewcommand{\thetable}{S\arabic{table}}
\setcounter{section}{0}
\setcounter{figure}{0}
\setcounter{table}{0}

\section{Problems and Results} \label{sec:appendix_problems}

\subsection{Discontinuous Function Approximation} \label{sec:func_approx}

In this problem, the multi-agent system is asked to build a machine learning model to approximate a piecewise oscillatory function with discontinuities based on provided training data. The target function is defined as:
\begin{align*}
  f(x) =
  \begin{cases}
    x^{2} - 2e^{-3000(x + 0.5)^{2}}, & -1 \leq x \le 0,\\
    \sin(10\pi x) + 1, & 0 < x \leq 1.
  \end{cases}
\end{align*}

However, the agents do \textit{not} have access to the underlying mathematical form of the function. They only observe the training datasets containing 200 input-output pairs. A validation set of 500 points is also provided for model evaluation but is not leaked during training.

\subsubsection{Problem Setup}

The user provided the following structured prompt to the system:

\begin{tcolorbox}[colback=problemcolor!10, colframe=problemcolor, title=Problem]
Build any model of your choice for data regression using the provided dataset.
\end{tcolorbox}

\begin{tcolorbox}[colback=requirementscolor!10, colframe=requirementscolor, title=Requirements]
Use PyTorch.
\end{tcolorbox}

\begin{tcolorbox}[colback=evaluationcolor!10, colframe=evaluationcolor, title=Evaluation]
The model should be scored on the MSE based on the provided validation set.
\end{tcolorbox}

\begin{tcolorbox}[colback=datacolor!10, colframe=datacolor, title=Data, breakable]
\textbf{training\_set:}
\begin{itemize}
\item filename: train\_data.npz
\item description: Training points for regression. Shapes: x\_train (200, 1), u\_train (200, 1).
\item loading\_instructions: Use np.load(`train\_data.npz'). Access via data[`x\_train'] and data['u\_train']
\end{itemize}

\textbf{validation\_set:}
\begin{itemize}
\item filename: val\_data.npz
\item description: Validation points for evaluation. Shapes: x\_val (500, 1), u\_val (500, 1).
\item loading\_instructions: Use np.load(`val\_data.npz'). Access via data[`x\_val'] and data[`u\_val']
\end{itemize}
\end{tcolorbox}

\subsubsection{Agents Setup}

Table~\ref{tab:func_approx_config} summarizes the agent configurations and evolutionary parameters used in this experiment. Gemini is used as the proposer due to its long context window capable of handling extensive retrieved knowledge and analysis reports. Claude is used as the root engineer and main engineering agent due to its strong coding capabilities. The temperature parameter is set lower for agents requiring deterministic outputs (e.g., root engineer, evaluator, engineer, debugger) and higher for creative agents (e.g., proposer, critic, Analyst).

\begin{table}[h!]
\centering
\caption{Agent configurations for function approximation experiment}
\label{tab:func_approx_config}
\small
\begin{tabular}{lll}
\toprule
\textbf{Agent} & \textbf{Model} & \textbf{Temperature} \\
\midrule
root engineer & Claude Haiku 4.5 & 0.0 \\
\midrule
data analyst & Gemini 2.5 Pro & 0.5 \\
evaluator & Claude Haiku 4.5 & 0.0 \\
retriever & Gemini 2.5 Pro & 0.0 \\
proposer & Gemini 2.5 Pro & 0.9 \\
critic & GPT-5 Mini & 0.5 \\
engineer & Claude Haiku 4.5 & 0.0 \\
debugger & GPT-5 Mini & 0.0 \\
result analyst & Gemini 2.5 Pro & 0.5 \\
selector & GPT-5 Mini, Grok-4 Fast, Gemini 2.5 Pro & - \\
\midrule
\multicolumn{3}{l}{\textbf{Evolutionary Parameters}} \\
\midrule
Max Iterations & 6 & \\
Parallel Mutations & 4 & \\
\bottomrule
\end{tabular}
\end{table}

\subsubsection{Results}

\begin{figure}[h!]
\centering
\includegraphics[width=0.75\linewidth]{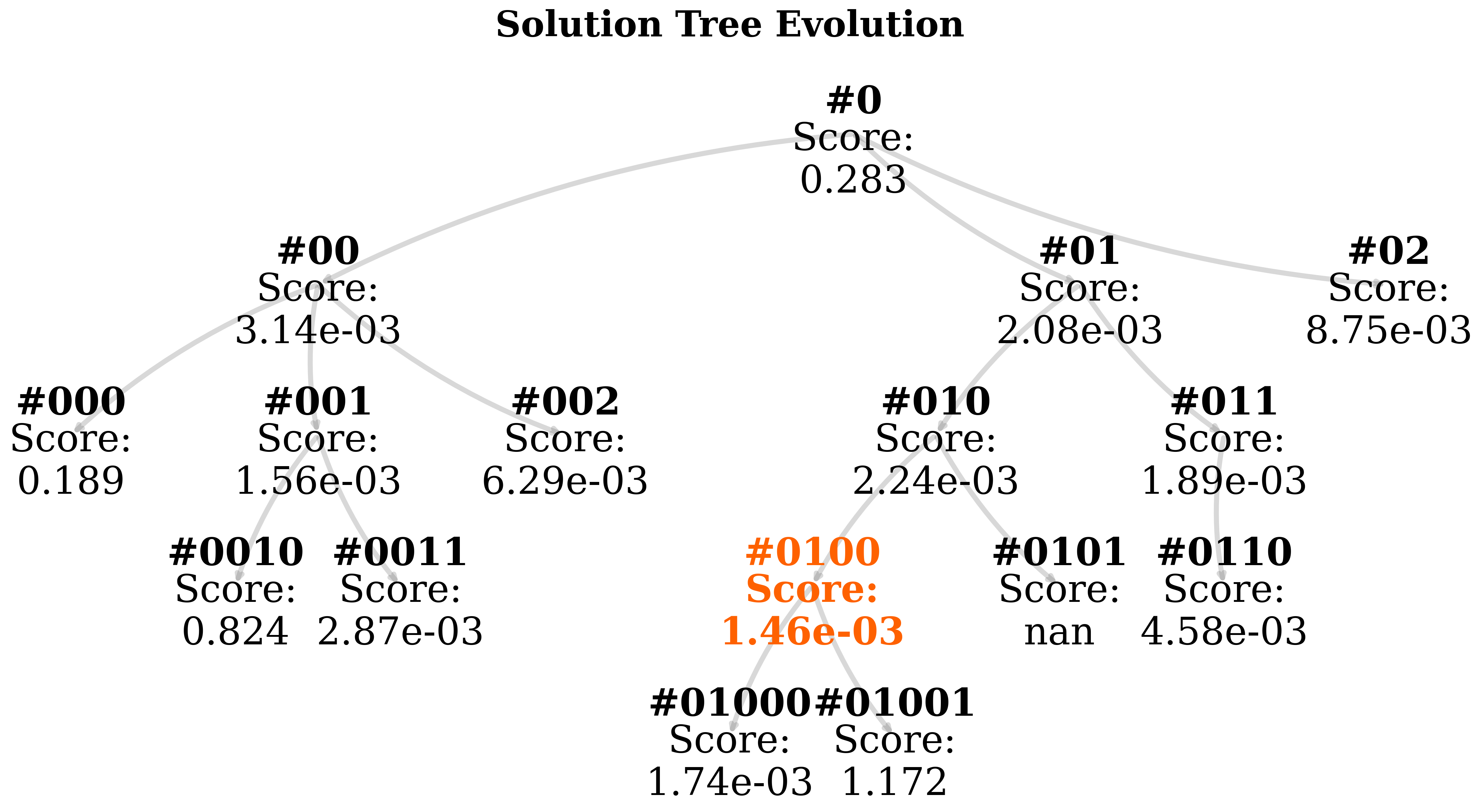}
\caption{Evolution of solution tree for function approximation problem. Each node of the tree represents a solution implemented and evaluated by the agents. The score shows the MSE error on the validation dataset. The best solution is colored in orange.}
\label{fig:func_approx_tree}
\end{figure}

\begin{figure}[h!]
\centering
\begin{subfigure}{0.48\linewidth}
\includegraphics[width=\linewidth]{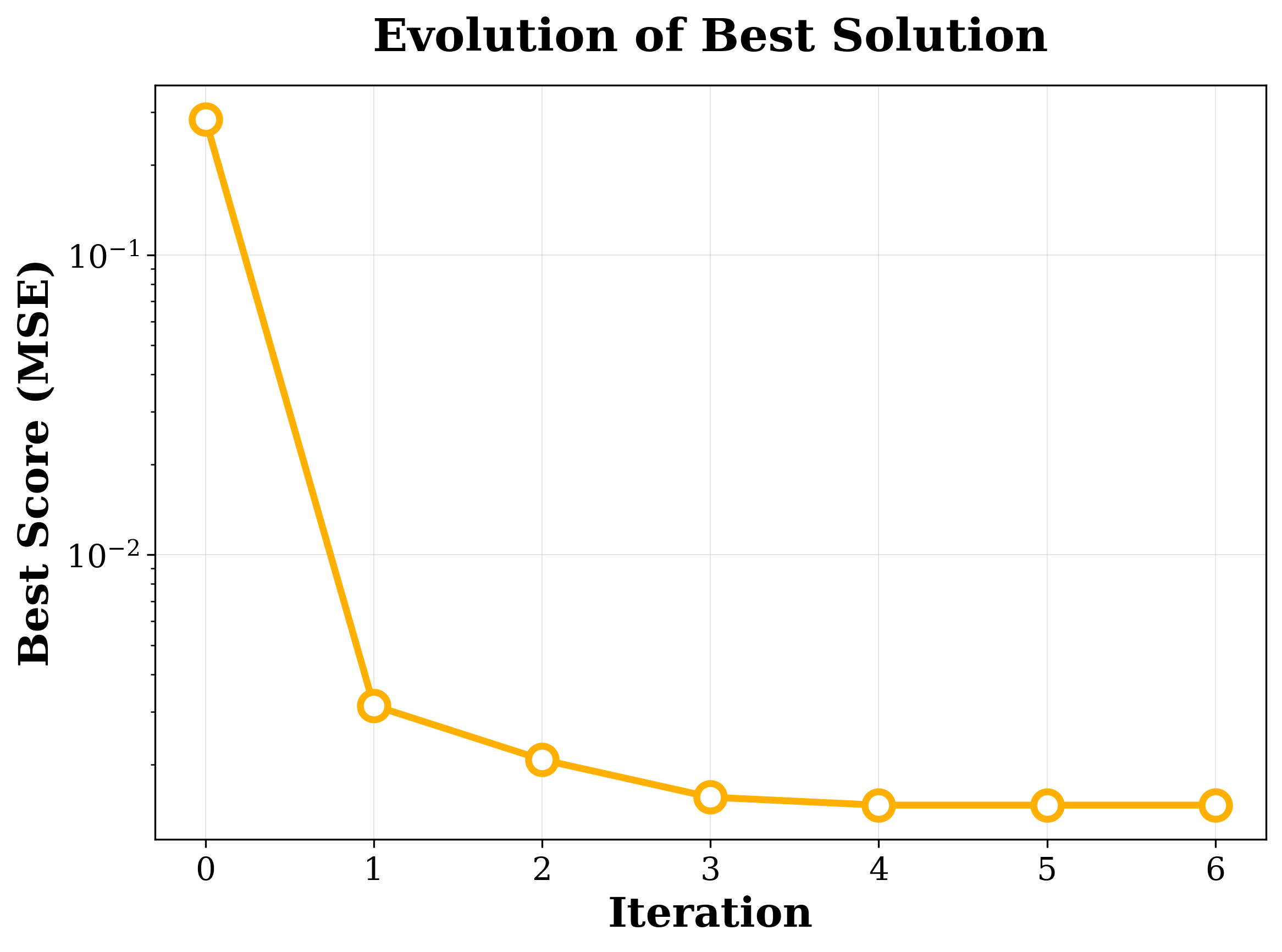}
\caption{Evolution of best MSE over tree expansion iterations}
\end{subfigure}
\hfill
\begin{subfigure}{0.48\linewidth}
\includegraphics[width=\linewidth]{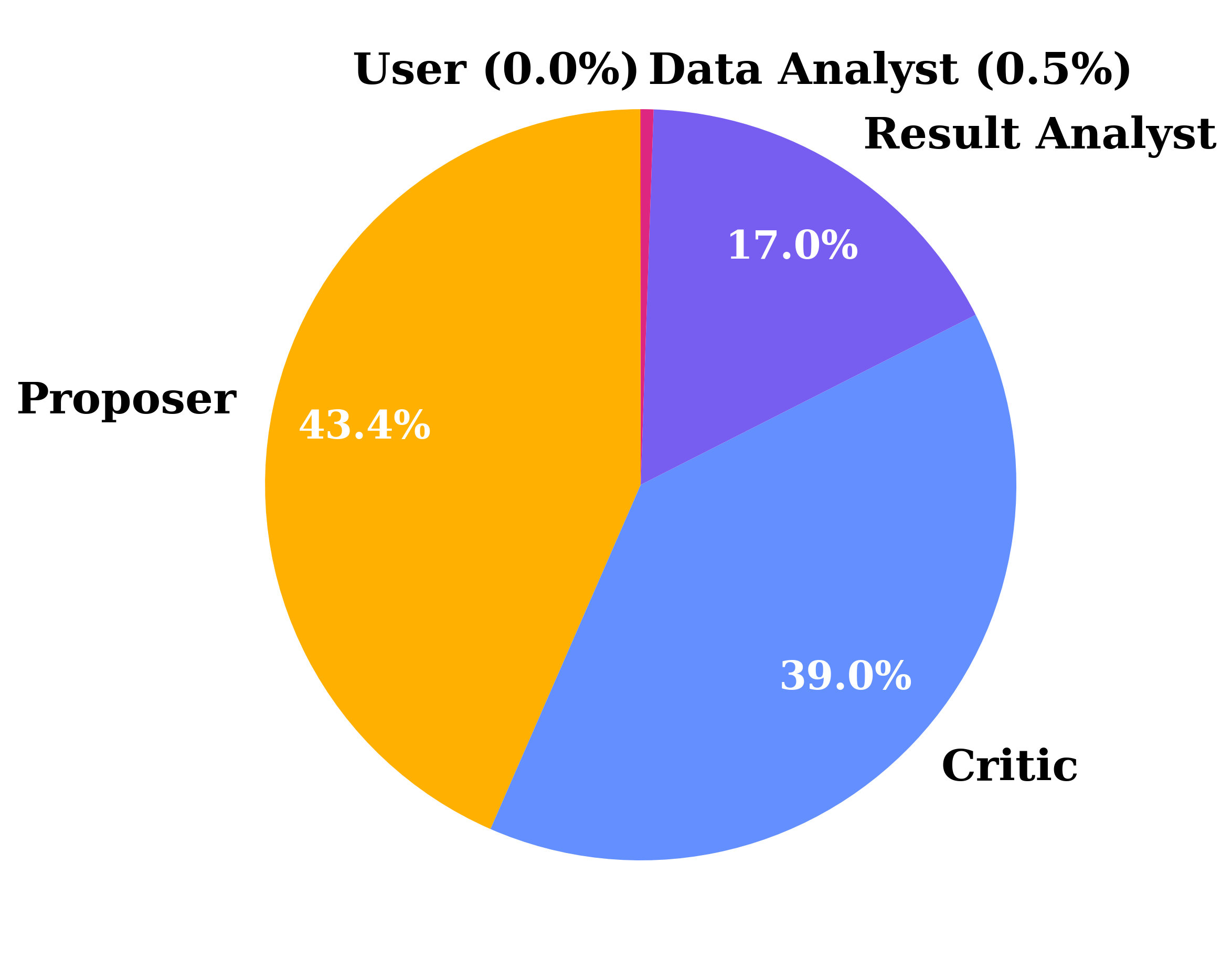}
\caption{Agent contributions by word count}
\end{subfigure}
\caption{Multi-agent evolution metrics for the discontinuous function approximation problem. \textbf{Left:} Evolution of best solution score over iterations. \textbf{Right:} Text contributions by agent type, showing the proposer contributes the most while the human user contributes minimal text. Engineering agents and selector agents are excluded in this plot.}
\label{fig:func_approx_evolution}
\label{fig:func_approx_contrib}
\end{figure}

The evolutionary process generated 16 solutions over 6 iterations. The score used for ranking in this problem is defined as the MSE on the validation set. Figure~\ref{fig:func_approx_tree} visualizes the solution evolution tree. Figure~\ref{fig:func_approx_evolution} tracks the best MSE obtained over evolution iterations. The best MSE decreases as the solution tree grows deeper before reaching a plateau. Figure~\ref{fig:func_approx_contrib} shows key planning agents' contributions by word count. The proposer agent contributes the most words, followed by critic and result analyst, while the user contributed minimal text. Non-planning agents, such as engineers and debuggers, are excluded in this plot. 

\begin{figure}[h!]
\centering
\includegraphics[width=0.95\linewidth]{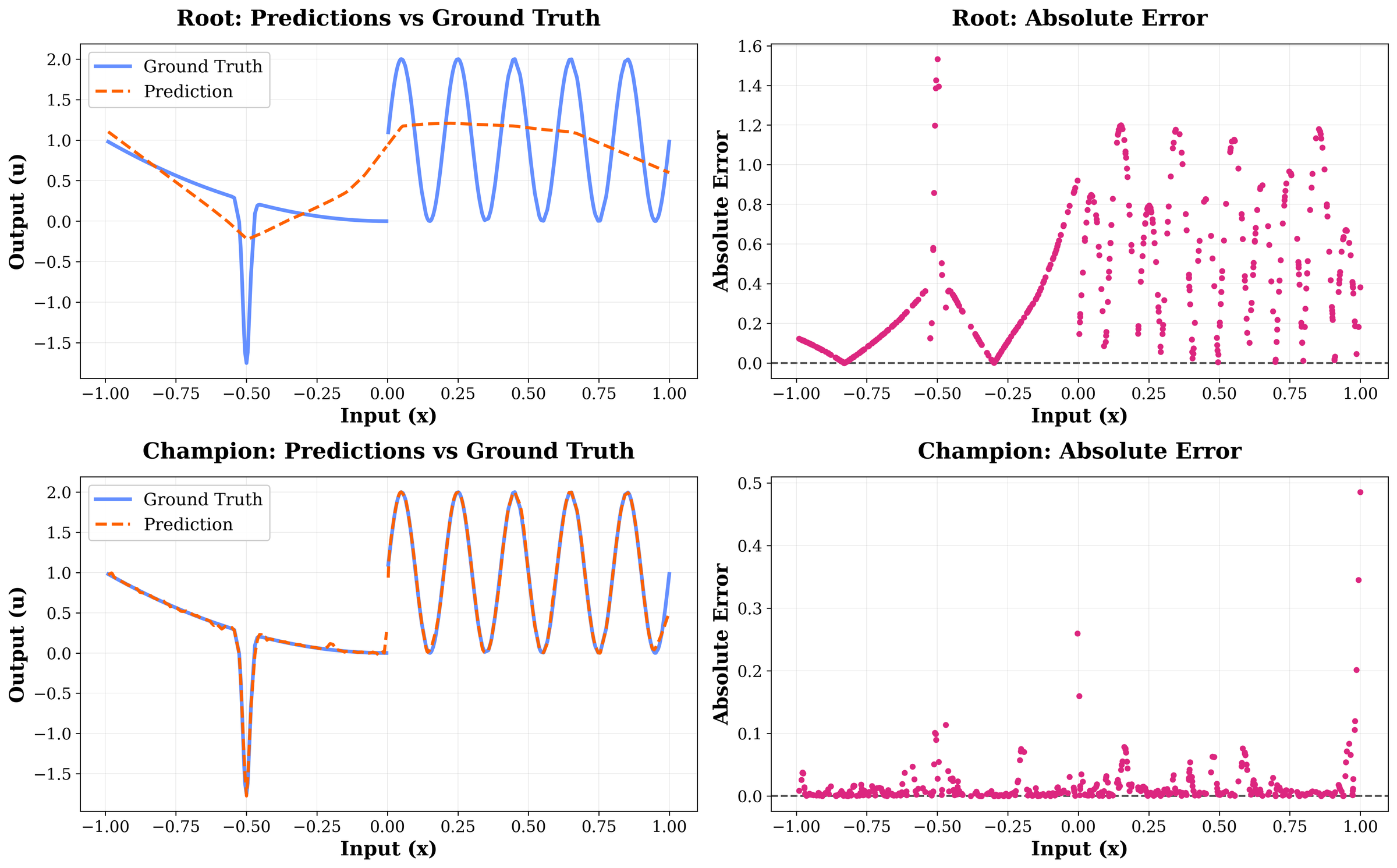}
\caption{Comparison of root and champion solutions for function approximation on the validation dataset. Top row: root solution, MSE = $0.283$. Bottom row: champion, MSE = $1.46 \times 10^{-3}$. Champion solution exhibits significant improvement from the root.}
\label{fig:func_approx_compare}
\end{figure}

Figure~\ref{fig:func_approx_compare} compares the predictions and absolute errors of the root and champion solutions. The root model fails to capture the oscillations and discontinuity, while the champion's solution shows significant improvements on both pieces of the function. Note that these plots are generated by the evaluator agent and are validated and reformatted by the authors for inclusion in this paper. The result analyst agent is able to visually analyze these plots and generate insights.

\subsubsection{Novelty of the Champion Solution} \label{sec:func_approx_novelty}

The champion solution (0100) achieved a test MSE of $1.46 \times 10^{-3}$, representing a 194$\times$ improvement over the root solution ($0.283$). 
The winning strategy  employs a Mixture-of-Experts (MoE) architecture with two specialized experts: a simple MLP for the region $x < 0$ and a Fourier-feature-augmented MLP for the oscillatory region $x > 0$. Expert selection was controlled by a parametric sigmoid gating function
\begin{align*}
  g(x) = \sigma(k \cdot (x - x_0))
\end{align*}
where $\sigma$ is the sigmoid function, $x_0$ is the trainable transition location (the final learned location is $0.04$, close to the discontinuity at $0$), and $k$ controls the gating sharpness. The final output combined experts as
\begin{align*}
  u_{\text{pred}}(x) = g(x) \cdot E_1(x) + (1-g(x)) \cdot E_2(x)
\end{align*} 
The parent solution (010) also used a similar MoE architecture. However, the training optimization suffered from numerical instability leading to divergence of loss. To mitigate this, the champion's solution defines a learnable parameter $k_{\text{raw}}$ that is then transformed via a bounded mapping to get $k = k_{\min} + (k_{\max} - k_{\min}) \cdot \sigma(k_{\text{raw}})$. This constrains $k \in [k_{\min}, k_{\max}] = [1.0, 50.0]$. A small weight decay ($10^{-5}$) on $k_{\text{raw}}$ prevents drift to extreme values. 

Noticeably, during development, the proposer agent reasons that 
\begin{quote}
  ``$k$ should be a trainable parameter. Let $k_{raw}$ be a learnable parameter, and define the sharpness as $k = \exp(k_{raw})$ to ensure positivity. This allows the optimizer to discover the optimal trade-off between a soft, stable mixture and a sharp, accurate partition. This approach, \underline{inspired by adaptive activations}, directly addresses the instability by incorporating the sharpness into the optimization loop, allowing for smoother, data-driven adjustments rather than forced, exponential increases."
\end{quote}
The above reasoning shows that the proposer agent is able to draw inspiration from the retrieved knowledge base entry on adaptive activation functions \cite{jagtap2020adaptive} and adapt it to the current problem context, even though the knowledge base entry does not directly discuss mixture-of-experts or gating functions.

In addition, the solution also implemented a weighted loss function to emphasize the discontinuity ($x=0$). The loss is given as
\begin{align*}
  \mathcal{L} = \frac{\sum_{i=1}^{N} w_i (u_{\text{pred}}(x_i) - u_{\text{true}}(x_i))^2}{\sum_{i=1}^{N} w_i},
\end{align*}
where $w_i = 1 + w_{\max} \cdot \exp(-x_i^2 / (2\sigma^2))$ with $w_{\max}=4.0$ and $\sigma=0.1$. 

The optimizer also employs separate AdamW parameters for MLP experts and gating parameters: expert networks use learning rate $5 \times 10^{-4}$ with weight decay $10^{-4}$, while gating parameters ($x_0$, $k_{\text{raw}}$) use learning rate $10^{-4}$ with weight decay $10^{-5}$. An entropy penalty $\mathcal{L}_{\text{ent}} = -\mathbb{E}[g \log g + (1-g)\log(1-g)]$ with weight ramping from 0 to $5 \times 10^{-4}$ after epoch 200 encourages crisp expert separation. The solution also uses gradient clipping (norm 1.0) and ReduceLROnPlateau scheduling (patience 100, factor 0.5) to prevent instability that the parent solution suffers from. Training runs for 2000 epochs with batch size 32.

\subsubsection{Token and Cost Analysis}

\begin{table}[htbp]
\centering
\caption{Token and cost analysis for Discontinuous Function Approximation (data-driven regression, 6 iterations). The result analyst consumes the most tokens (48\%) due to multimodal inputs; LLM API time exceeds GPU training time because the network is small and fast to train.}
\label{tab:cost_func_approx}

\begin{tabular}{lrrrr}
\toprule
\textbf{Agent (Model)} & \textbf{API Calls} & \textbf{Tokens (M)} & \textbf{Cost (\$)} & \textbf{Time (s)} \\
\midrule
\multicolumn{5}{l}{\textit{Overall}} \\
\quad Total             & 164 & 6.26         & 11.30         & 9{,}290 \\
\quad LLM (multi-agent) & 164 & 6.26  & 11.30 & 5{,}643  \\
\quad GPU Training      & --  & --           & --            & 3{,}647 \\
\midrule
\multicolumn{5}{l}{\textit{Agent Breakdown}} \\
\quad Proposer (Gemini-2.5-Pro)  & 57 & 1.50 (24\%) & 4.10 (36\%) & 2{,}391 (42\%) \\
\quad Analyst (Gemini-2.5-Pro)   & 19 & 3.03 (48\%) & 4.51 (40\%) &   817 (14\%) \\
\quad Engineer (Claude-Haiku)    & 25 & 0.52  (8\%) & 1.33 (12\%) & 1{,}078 (19\%) \\
\quad Retriever (Gemini-2.5-Pro) & 19 & 0.47  (8\%) & 1.01  (9\%) &   502  (9\%) \\
\quad Critic (GPT-4o-Mini)       & 38 & 0.67 (11\%) & 0.32  (3\%) &   763 (14\%) \\
\quad Debugger (GPT-4o-Mini)     &  6 & 0.07  (1\%) & 0.03  (0\%) &    91  (2\%) \\
\bottomrule
\end{tabular}

\end{table}

The proposer accounts for 36\% of LLM cost and 42\% of LLM wall-clock time. The analyst incurs the highest token count (48\%) due to multimodal prompts combining training logs and evaluation plots. Uniquely, LLM API time (1.57\,h) exceeds GPU training time (1.01\,h), as the neural networks for this problem are small and fast to train. The debugger is invoked only 6 times, indicating stable generated code.

\clearpage

\subsection{Solving Poisson Equation on L-shaped Domain with PINN} \label{sec:poisson_l}

This problem tests the system's ability to discover effective PINN architectures and loss balancing strategies for solving PDEs on irregular geometries.

\subsubsection{Problem Setup}

The user provided the following structured prompt to the system:

\begin{tcolorbox}[colback=problemcolor!10, colframe=problemcolor, title=Problem]
Solve the Poisson equation using Physics-Informed Neural Networks (PINNs):
$$\nabla^2 u(x, y) = 1, \quad (x, y) \in \Omega$$
Domain: L-shaped region $\Omega = [-1, 1]^2 \setminus [0, 1]^2$.
Boundary Conditions: Dirichlet $u = 0$ on $\partial\Omega$.
\end{tcolorbox}

\begin{tcolorbox}[colback=requirementscolor!10, colframe=requirementscolor, title=Requirements]
Use PyTorch.
\end{tcolorbox}

\begin{tcolorbox}[colback=evaluationcolor!10, colframe=evaluationcolor, title=Evaluation]
Score = $\frac{1}{2}$(PDE Residual MSE + BC MSE), where PDE residual is evaluated at 5000 interior points and BC error at 500 points per boundary side.
\end{tcolorbox}

\subsubsection{Agents Setup}

Table~\ref{tab:poisson_config} summarizes the agent configurations and evolutionary parameters used in this experiment.

\begin{table}[h!]
\centering
\caption{Agent configurations for Poisson L experiment}
\label{tab:poisson_config}
\small
\begin{tabular}{lll}
\toprule
\textbf{Agent} & \textbf{Model} & \textbf{Temperature} \\
\midrule
root engineer & GPT-5 Mini & 0.0 \\
\midrule
data analyst & Gemini 2.5 Pro & 0.1 \\
evaluator & Claude Haiku 4.5 & 0.0 \\
retriever & Gemini 2.5 Pro & 0.0 \\
proposer & Gemini 2.5 Pro & 0.8 \\
critic & GPT-5 Mini & 0.5 \\
engineer & Claude Haiku 4.5 & 0.0 \\
debugger & GPT-5 Mini & 0.0 \\
result analyst & Gemini 2.5 Pro & 0.1 \\
selector & GPT-5 Mini, Grok-4 Fast, Gemini 2.5 Pro & - \\
\midrule
\multicolumn{3}{l}{\textbf{Evolutionary Parameters}} \\
\midrule
Max Iterations & 6 & \\
Parallel Mutations & 4 & \\
\bottomrule
\end{tabular}
\end{table}

\subsubsection{Results}

\begin{figure}[h!]
\centering
\includegraphics[width=0.75\linewidth]{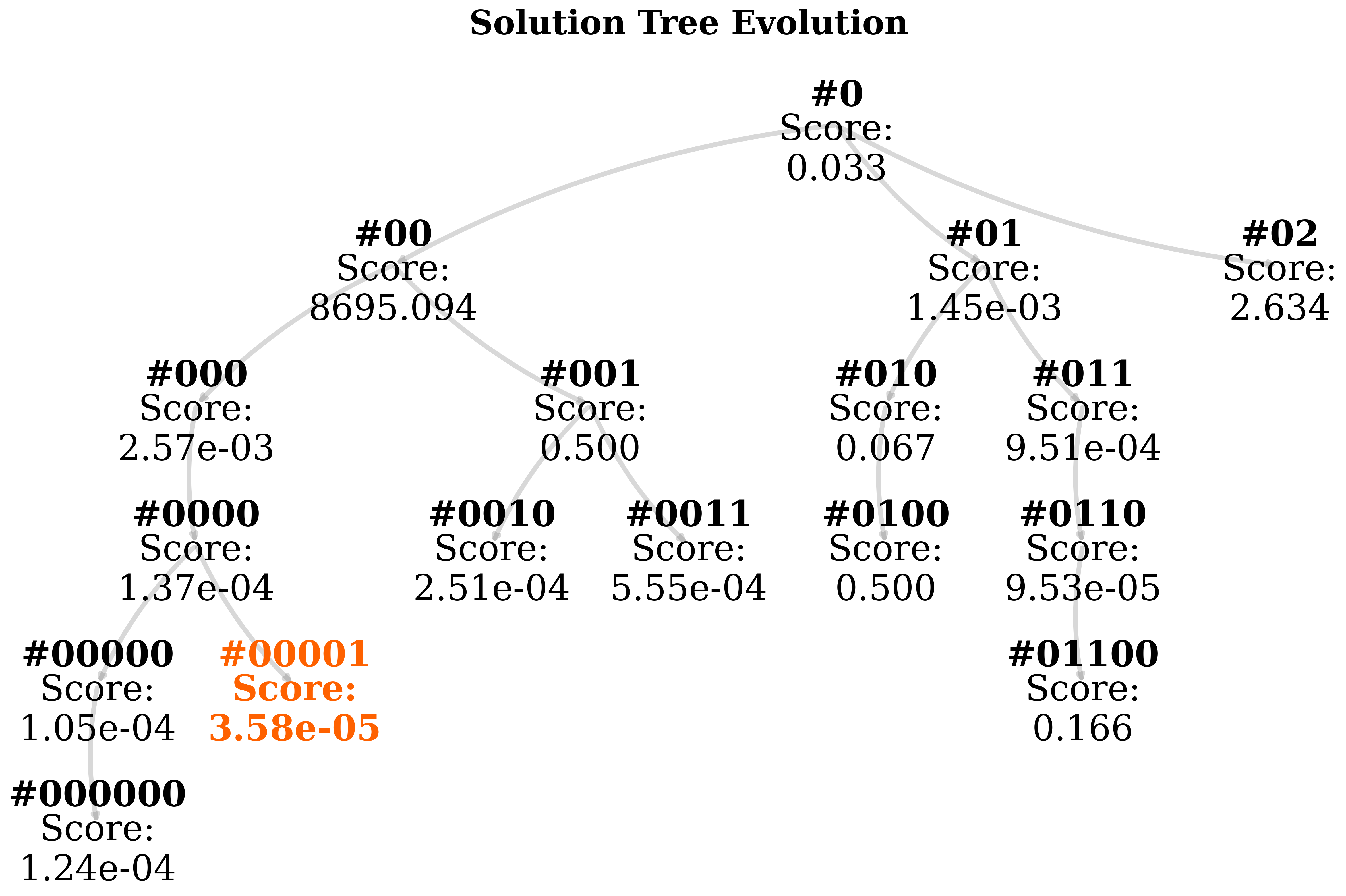}
\caption{Evolution of solution tree for Poisson equation on L-shaped domain. Each node of the tree represents a solution implemented and evaluated by the agents. The score shows the combined PDE residual and boundary condition MSE. The best solution is colored in orange.}
\label{fig:poisson_tree}
\end{figure}

\begin{figure}[h!]
\centering
\begin{subfigure}{0.48\linewidth}
\includegraphics[width=\linewidth]{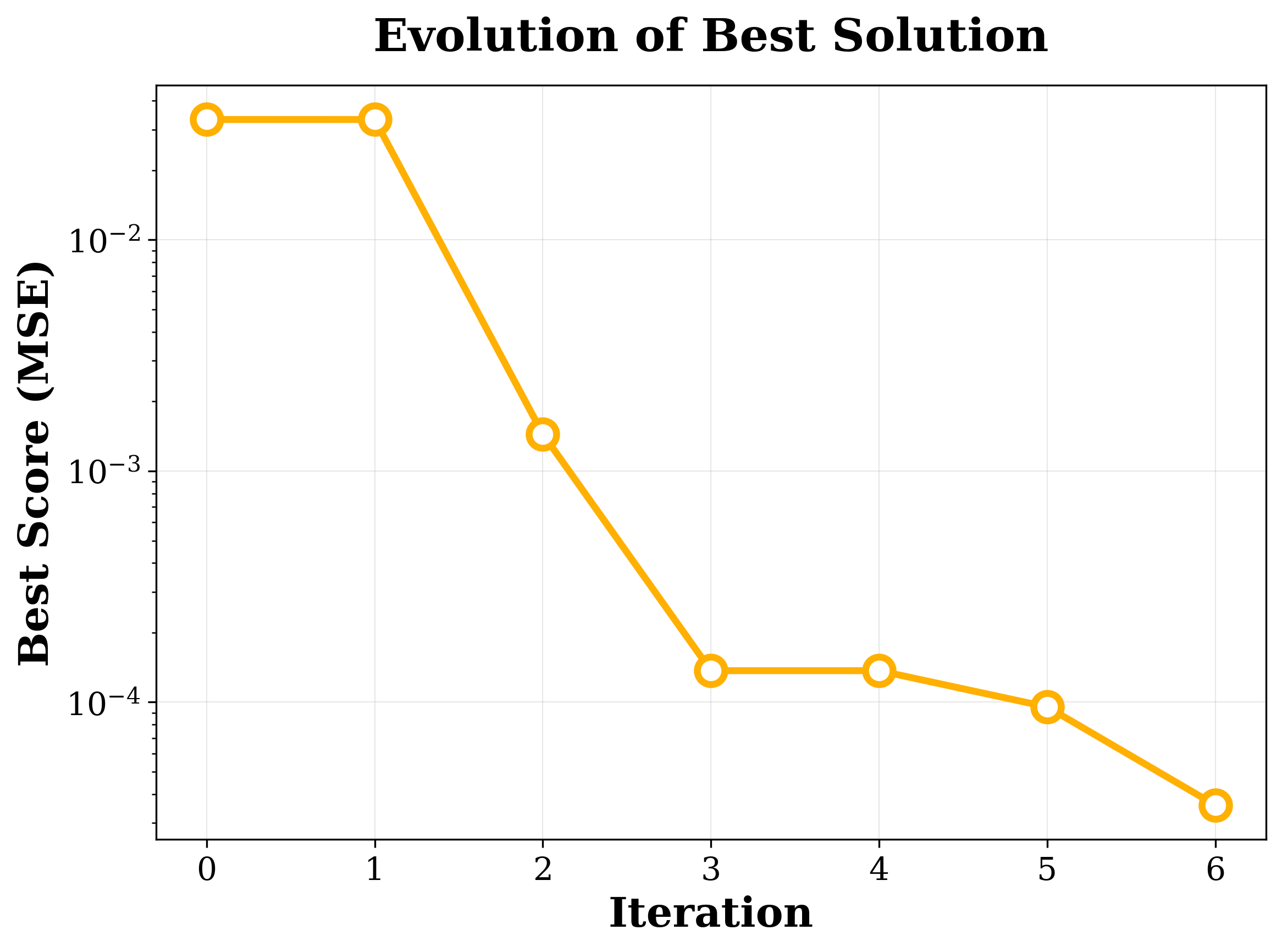}
\caption{Evolution of best solution}
\end{subfigure}
\hfill
\begin{subfigure}{0.48\linewidth}
\includegraphics[width=\linewidth]{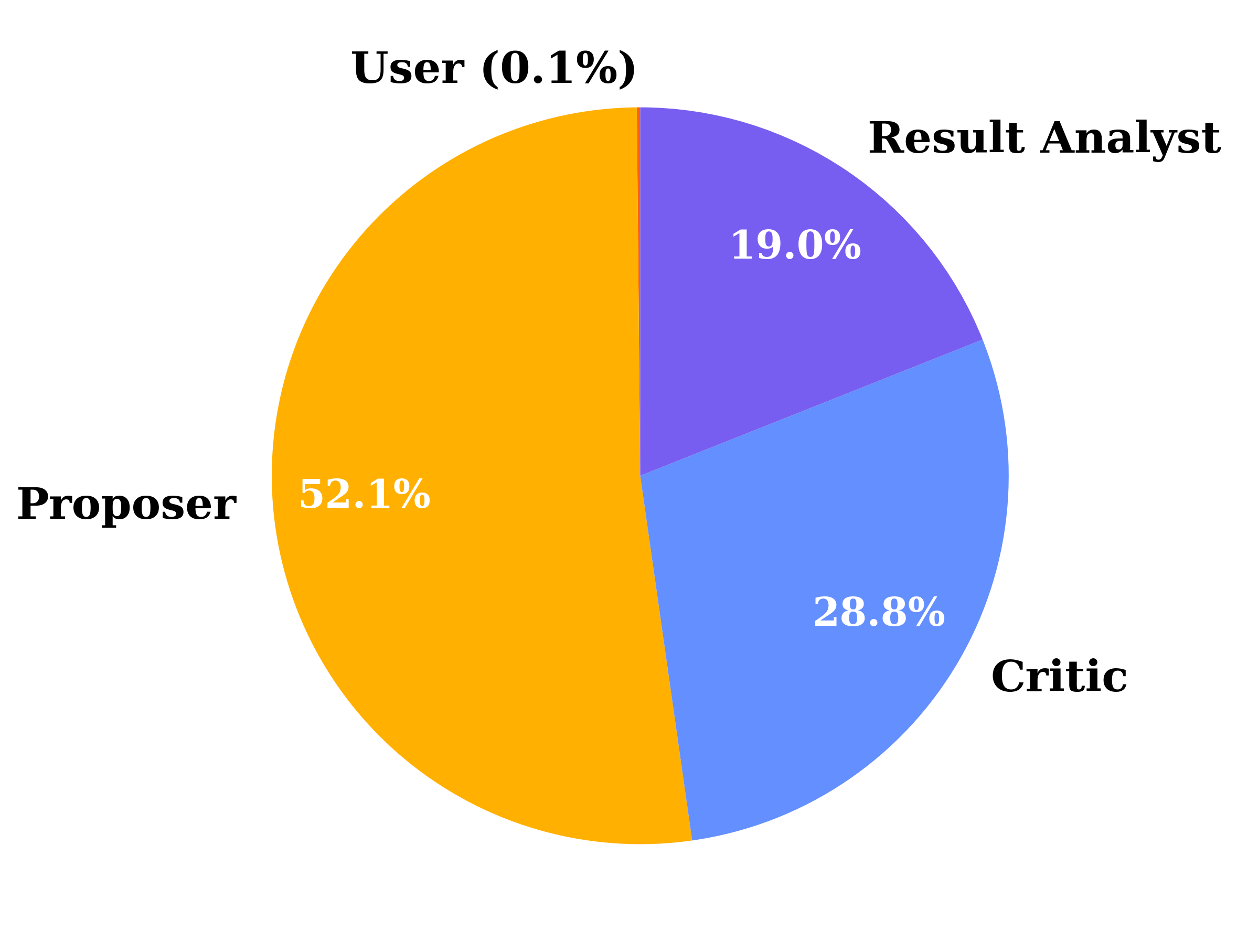}
\caption{Agent text contributions}
\end{subfigure}
\caption{Multi-agent evolution metrics for the Poisson equation on L-shaped domain. \textbf{Left:} Evolution of best solution score over iterations. \textbf{Right:} Text contributions by agent type, showing the proposer contributes the most while the human user contributes minimal text. Engineering agents and selector agents are excluded in this plot.}
\label{fig:poisson_evolution}
\label{fig:poisson_contrib}
\end{figure}

\begin{figure}[h!]
\centering
\begin{subfigure}{0.45\linewidth}
\includegraphics[width=\linewidth]{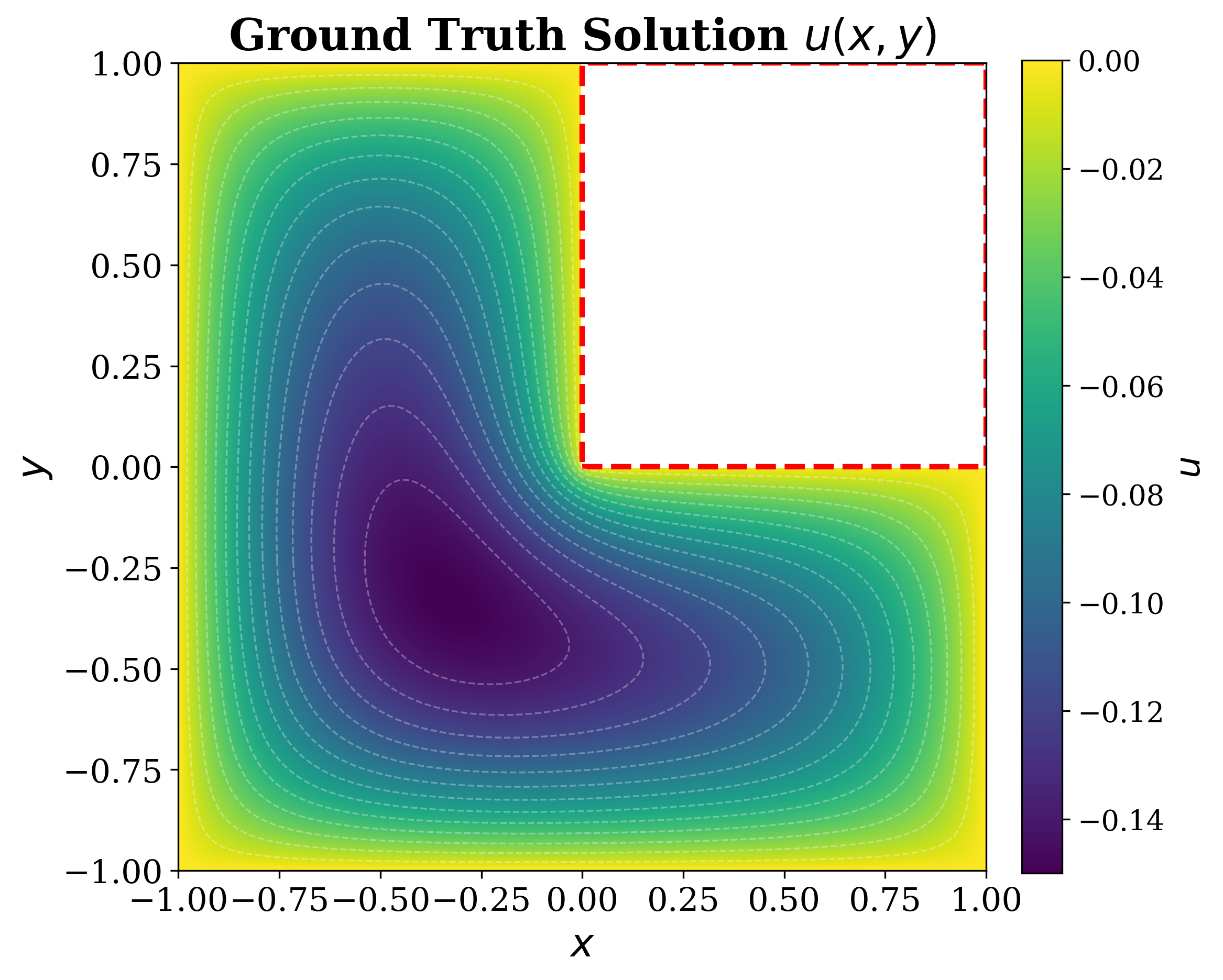}
\caption{Ground truth solution}
\end{subfigure}

\vspace{0.5cm}

\begin{subfigure}{0.9\linewidth}
\includegraphics[width=\linewidth]{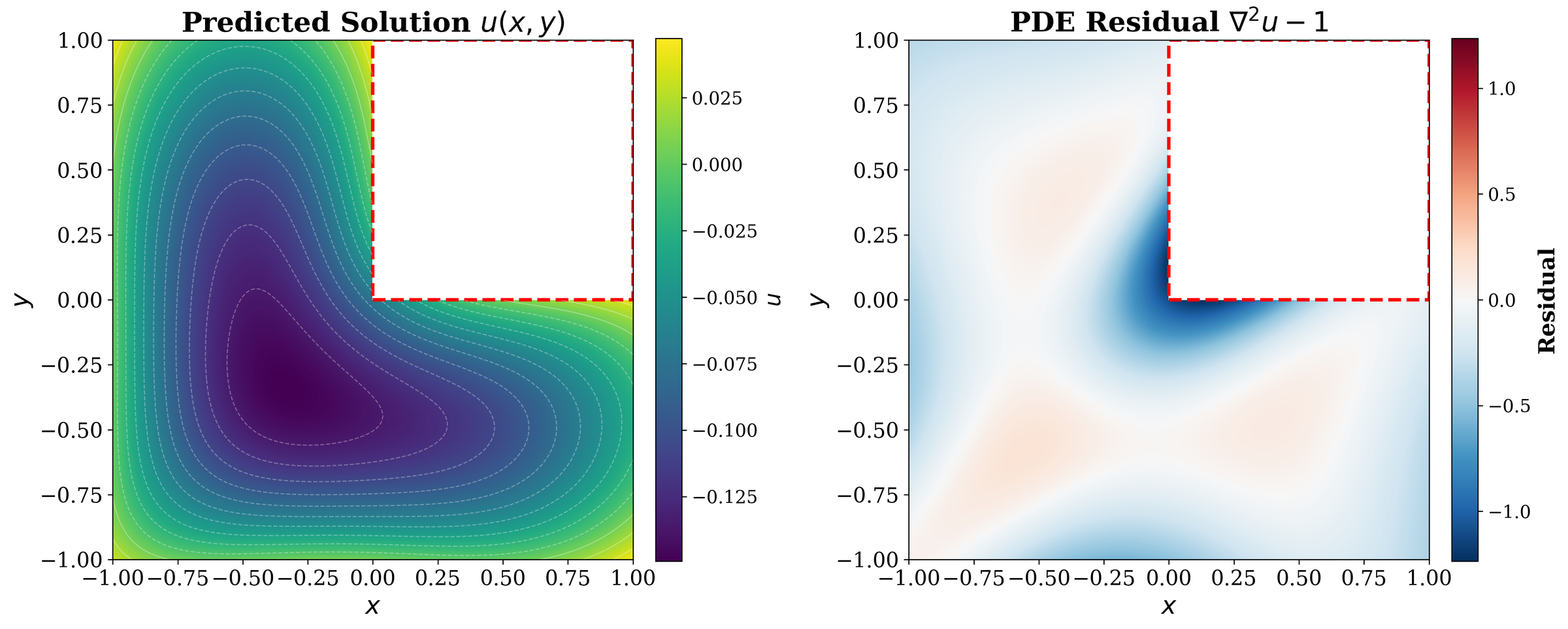}
\caption{Root solution (Score = 0.0332)}
\end{subfigure}

\vspace{0.5cm}

\begin{subfigure}{0.9\linewidth}
\includegraphics[width=\linewidth]{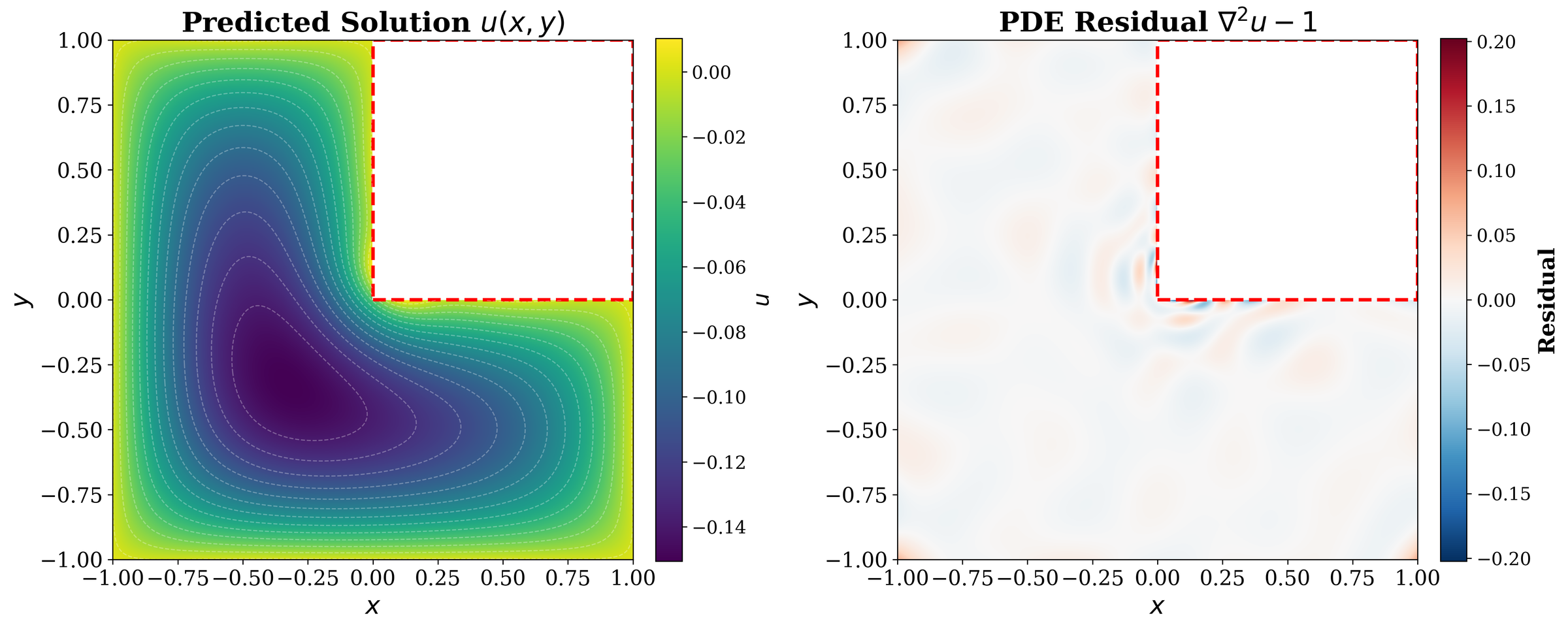}
\caption{Champion solution (Score = 3.58e-05)}
\end{subfigure}
\caption{Comparison of ground truth, root, and champion solutions for Poisson equation on L-shaped domain. \textbf{(a)} Ground truth solution from finite difference method. \textbf{(b)} Root solution, Score = $3.32 \times 10^{-2}$. \textbf{(c)} Champion solution, Score = $3.58 \times 10^{-5}$. The score shows the combined PDE residual and boundary condition MSE. The Champion solution exhibits significant improvement from the root.}
\label{fig:poisson_compare}
\end{figure}

The evolutionary process generated 17 solutions over 6 iterations. Figure~\ref{fig:poisson_tree} visualizes the solution tree showing parent-child relationships across all 17 solutions. The champion solution (00001) achieved a test score of $3.58 \times 10^{-5}$, representing a 927$\times$ improvement over the root solution ($3.32 \times 10^{-2}$). Figure~\ref{fig:poisson_evolution} shows the evolution of the best solution score across iterations and shows the text contributions from different agent types, with the proposer agent contributing the most. Figure~\ref{fig:poisson_compare} compares the ground truth solution obtained from finite difference method with 500$\times$500 grid resolution against the root and champion PINN solutions. The root solution exhibits large errors especially around the corner at $(0,0)$, while the champion solution achieves much reduced errors nearly uniformly distributed across the domain.

\subsubsection{Novelty of the Champion Solution} \label{sec:poisson_novelty}

The agents planned to decompose the solution into multiple components. The proposer agent's initial plan is to decompose the target solution as $u(x,y) = u_{nn}(x,y) + u_p(x,y) + u_{s1}(x,y) + u_{s2}(x,y)$. Here $u_{nn}$ is a 5-layer MLP, $u_p = (x^2+y^2)/4$ is an analytical particular solution satisfying $\nabla^2 u_p = 1$, and $u_{s1}, u_{s2}$ are analytical singular basis functions $A r^{2/3}\sin(\frac{2}{3}(\theta - \pi/2))$ and $B r^{4/3}\sin(\frac{4}{3}(\theta - \pi/2))$ with trainable amplitudes $A$ and $B$. The goal is to have the singular terms capture the singularity at the origin, while $u_{nn}$ learns the smooth remainder. 

\begin{figure}[h!]
\centering
\includegraphics[width=0.95\linewidth]{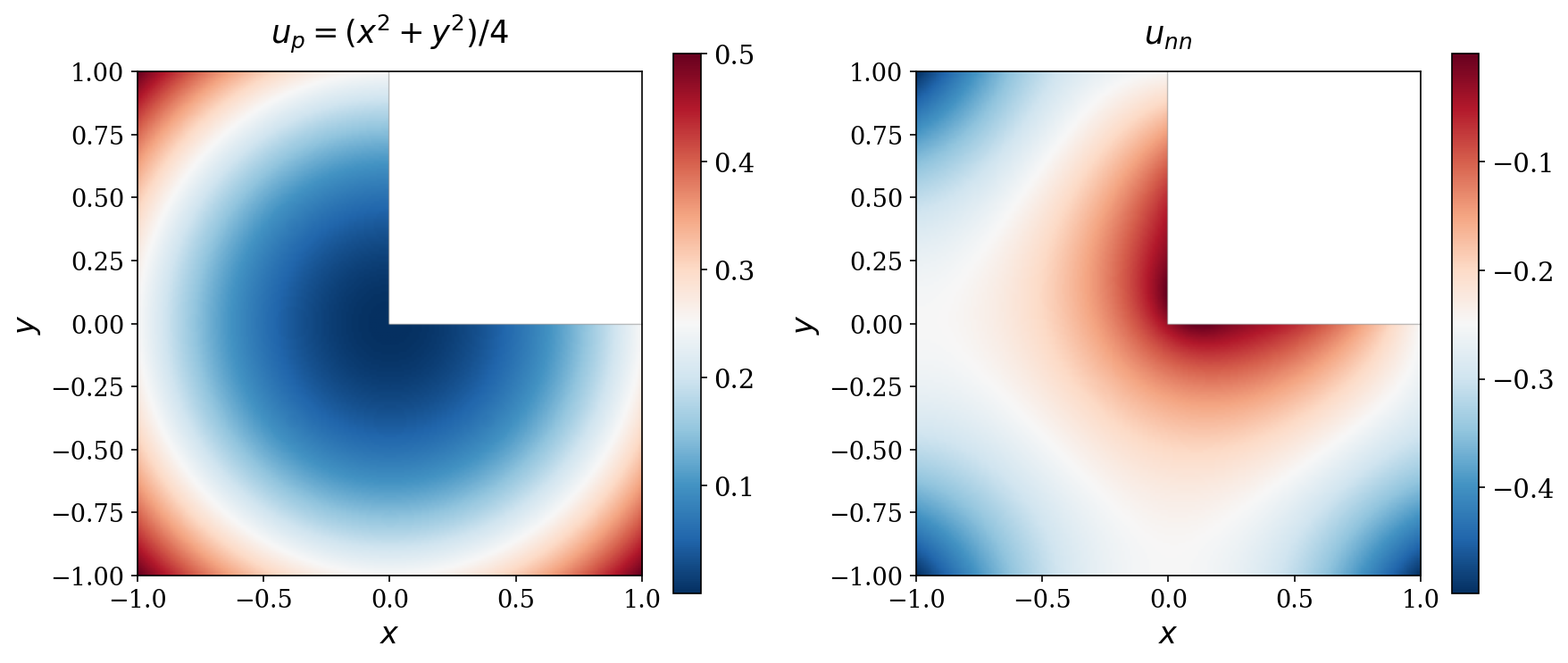}
\caption{Decomposition of the champion solution showing the two active components. \textbf{Left:} $u_p = (x^2 + y^2)/4$, the particular solution satisfying $\nabla^2 u_p = 1$. \textbf{Right:} $u_{nn}$, the MLP-learned component.}
\label{fig:poisson_decomposition}
\end{figure}

However, the engineer agent made a critical bug in the implementation: a PyTorch \texttt{.detach()} call cuts off gradient flow to $A$ and $B$, preventing them from being learned, so $A$ and $B$ are kept as the initial values $A=B=0$ in the final model. Therefore, this 4-way decomposition strategy is not fully realized in practice. The actual learned decomposition is $u(x,y) = u_{nn}(x,y) + u_p(x,y)$. The two components are visualized in Figure~\ref{fig:poisson_decomposition}. 

In addition to the decomposition, the agents improved sampling of collocation points. The critic agent noted that 
\begin{quote}
  ``Uniform sampling of interior points is inefficient. The difficulty of the problem is concentrated at the origin, yet this region is sampled with the same probability as any other. This starves the model of the critical information needed to properly resolve the singularity."
\end{quote}

The agents thus implemented a 70/30 hybrid sampling scheme: 70\% of collocation points are sampled uniformly over the domain, while 30\% use an importance sampler. The importance sampler uses inverse transform sampling in polar coordinates centered at the origin. Specifically, the radius $r$ is drawn from a truncated power-law distribution $p(r) \propto r^{-\gamma}$ on $[r_{\min}, r_{\max}]$ with $\gamma = 0.7$, $r_{\min} = 10^{-3}$, and $r_{\max} = \sqrt{2}$ (the maximum distance from origin to domain corner). For a uniform random variable $U \in [0,1]$, the inverse transform formula is:
\begin{align*}
  r = \left[U \cdot (r_{\max}^{1-\gamma} - r_{\min}^{1-\gamma}) + r_{\min}^{1-\gamma}\right]^{1/(1-\gamma)} = \left[U \cdot (r_{\max}^{0.3} - r_{\min}^{0.3}) + r_{\min}^{0.3}\right]^{1/0.3}
\end{align*}
The angle $\theta$ is sampled uniformly in $[\pi/2, 2\pi]$ to cover the three quadrants of the L-shaped domain, then converted to Cartesian coordinates $(x, y) = (r\cos\theta, r\sin\theta)$ and filtered to reject points outside the L-shaped domain. This power-law distribution with $\gamma = 0.7 < 1$ biases points near the origin. 

\begin{figure}[h!]
\centering
\includegraphics[width=0.95\linewidth]{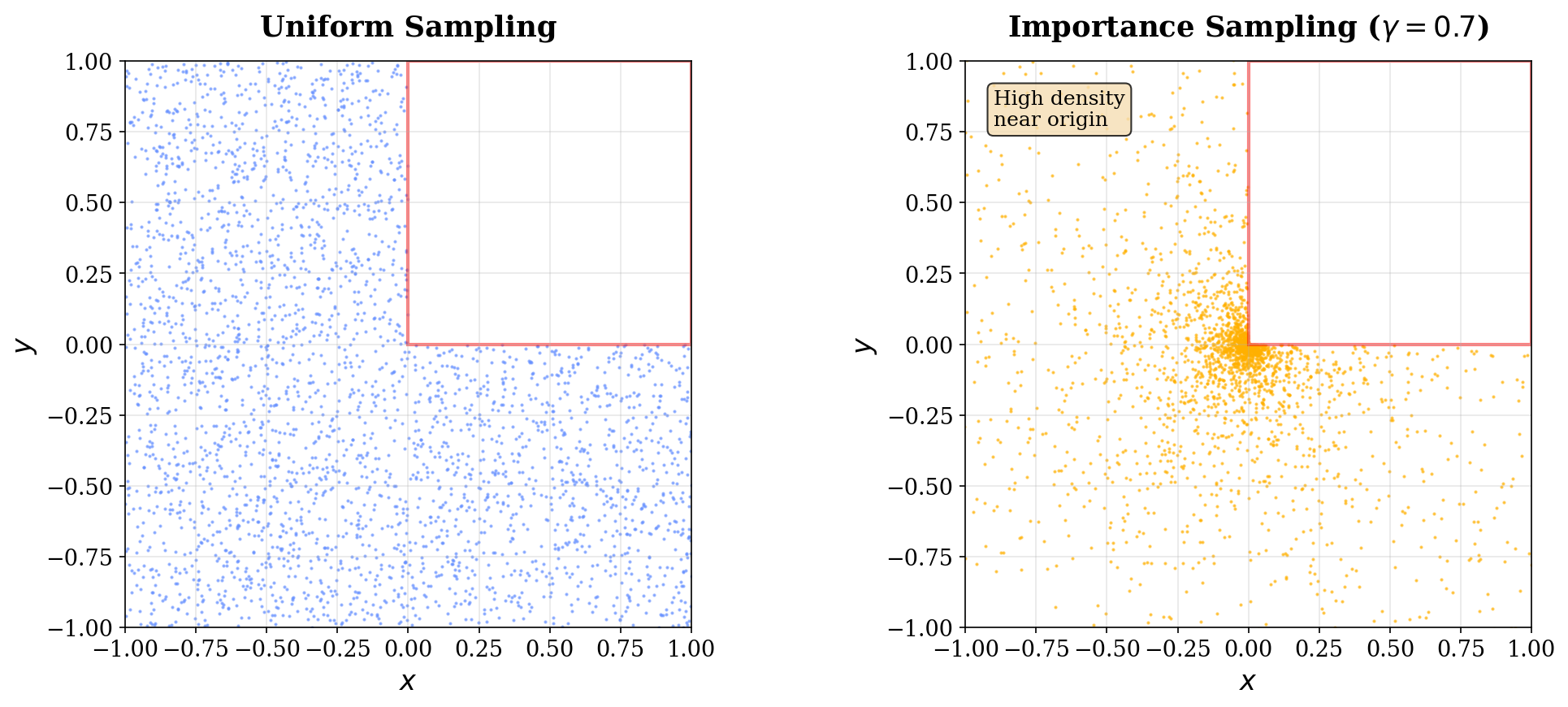}
\caption{Uniform and importance sampling of collocation points. \textbf{Left:} Uniform sampling. \textbf{Right:} Importance sampling with power-law distribution $p(r) \propto r^{-0.7}$ concentrates points near the origin. During each epoch of training, 70\% of collocation points are sampled uniformly and 30\% from the importance sampler.}
\label{fig:poisson_sampling}
\end{figure}

The code implements \texttt{n\_uniform = int(4096 * 0.7) = 2867} uniform points and \texttt{n\_importance = 4096 - 2867 = 1229} importance-sampled points. During training, each epoch samples a fresh batch of collocation points according to this scheme. Figure~\ref{fig:poisson_sampling} visualizes the two samples.

Finally, the agents improved the optimization by first training with Adam optimizer for 12,000 epochs using a CosineAnnealingLR scheduler. After Adam, it switches to L-BFGS training for 3,000 epochs, where the collocation points are fixed and do not change between epochs.

\subsubsection{Token and Cost Analysis}

\begin{table}[htbp]
\centering
\caption{Token and cost analysis for Poisson Equation on L-shaped Domain (PINN, 6 iterations). GPU training time dominates at 5.6\,h vs.\ 1.7\,h LLM time; the proposer accounts for 44\% of LLM cost.}
\label{tab:cost_poisson}

\begin{tabular}{lrrrr}
\toprule
\textbf{Agent (Model)} & \textbf{API Calls} & \textbf{Tokens (M)} & \textbf{Cost (\$)} & \textbf{Time (s)} \\
\midrule
\multicolumn{5}{l}{\textit{Overall}} \\
\quad Total             & 156 & 5.54         & 10.68         & 26{,}413 \\
\quad LLM (multi-agent) & 156 & 5.54 & 10.68 &  6{,}174 \\
\quad GPU Training      & --  & --           & --            & 20{,}239 \\
\midrule
\multicolumn{5}{l}{\textit{Agent Breakdown}} \\
\quad Proposer (Gemini-2.5-Pro)  & 57 & 1.71 (31\%) & 4.72 (44\%) & 2{,}723 (44\%) \\
\quad Analyst (Gemini-2.5-Pro)   & 19 & 2.12 (38\%) & 3.31 (31\%) &   757 (12\%) \\
\quad Engineer (Claude-Haiku)    & 21 & 0.47  (8\%) & 1.20 (11\%) & 1{,}022 (17\%) \\
\quad Retriever (Gemini-2.5-Pro) & 19 & 0.50  (9\%) & 1.08 (10\%) &   546  (9\%) \\
\quad Critic (GPT-4o-Mini)       & 38 & 0.72 (13\%) & 0.36  (3\%) & 1{,}091 (18\%) \\
\quad Debugger (GPT-4o-Mini)     &  2 & 0.03  (0\%) & 0.01  (0\%) &    35  (1\%) \\
\bottomrule
\end{tabular}

\end{table}

GPU training time (5.62\,h) dominates wall-clock cost, exceeding LLM API time (1.72\,h) by 3.3$\times$, consistent with the iterative residual minimization in PINN training. The proposer accounts for 44\% of LLM cost and the analyst for 38\% of total tokens. The debugger is called only twice across 156 LLM calls, indicating stable generated code.

\clearpage

\subsection{Solving Burger's Equation using PINN} \label{sec:burger}

In this test, the multi-agent system is tasked with building a PINN to solve the viscous Burger's equation with a challenging initial condition that leads to sharp gradients in the solution. 

\subsubsection{Problem Setup}

The user provided the following structured prompt to the system:

\begin{tcolorbox}[colback=problemcolor!10, colframe=problemcolor, title=Problem]
Build a PINN to solve the Burger's equation with periodic boundary conditions:
$$u_t + u u_x - \frac{0.01}{\pi} u_{xx} = 0, \quad x \in [0, 1], \quad t \in [0, 1]$$
$$u(0, x) = \sin(2 \pi x) - \cos(4 \pi x)$$
\end{tcolorbox}

\begin{tcolorbox}[colback=requirementscolor!10, colframe=requirementscolor, title=Requirements]
Use PyTorch. Build a PINN - do NOT use numerical methods.
\end{tcolorbox}

\begin{tcolorbox}[colback=evaluationcolor!10, colframe=evaluationcolor, title=Evaluation]
Find the PDE residual mean squared error (MSE) over a dense set of collocation points in the domain (must satisfy the Burger's equation). Find the boundary condition MSE over a dense set of points on the boundary (must satisfy the periodic boundary conditions). Find initial condition error.
$$\text{Score} = \frac{1}{3}(\text{PDE Residual MSE} + \text{IC Error MSE} + \text{BC Error MSE})$$
\end{tcolorbox}

\subsubsection{Agents Setup}

Table~\ref{tab:burger_config} summarizes the agent configurations and evolutionary parameters used in this experiment.

\begin{table}[h!]
\centering
\caption{Agent configurations for Burger's PINN experiment}
\label{tab:burger_config}
\small
\begin{tabular}{lll}
\toprule
\textbf{Agent} & \textbf{Model} & \textbf{Temperature} \\
\midrule
root engineer & Claude Haiku 4.5 & 0.0 \\
\midrule
data analyst & Gemini 2.5 Pro & 0.1 \\
evaluator & Claude Haiku 4.5 & 0.0 \\
retriever & Gemini 2.5 Pro & 0.0 \\
proposer & Grok-4 Fast Reasoning & 0.5 \\
critic & GPT-5 Mini & 0.3 \\
engineer & Claude Haiku 4.5 & 0.0 \\
debugger & GPT-5 Mini & 0.0 \\
result analyst & Gemini 2.5 Pro & 0.1 \\
selector & GPT-5 Mini, Grok-4 Fast, Gemini 2.5 Pro & - \\
\midrule
\multicolumn{3}{l}{\textbf{Evolutionary Parameters}} \\
\midrule
Max Iterations & 6 & \\
Parallel Mutations & 4 & \\
\bottomrule
\end{tabular}
\end{table}

\subsubsection{Results}

\begin{figure}[h!]
\centering
\begin{subfigure}{0.45\linewidth}
\includegraphics[width=\linewidth]{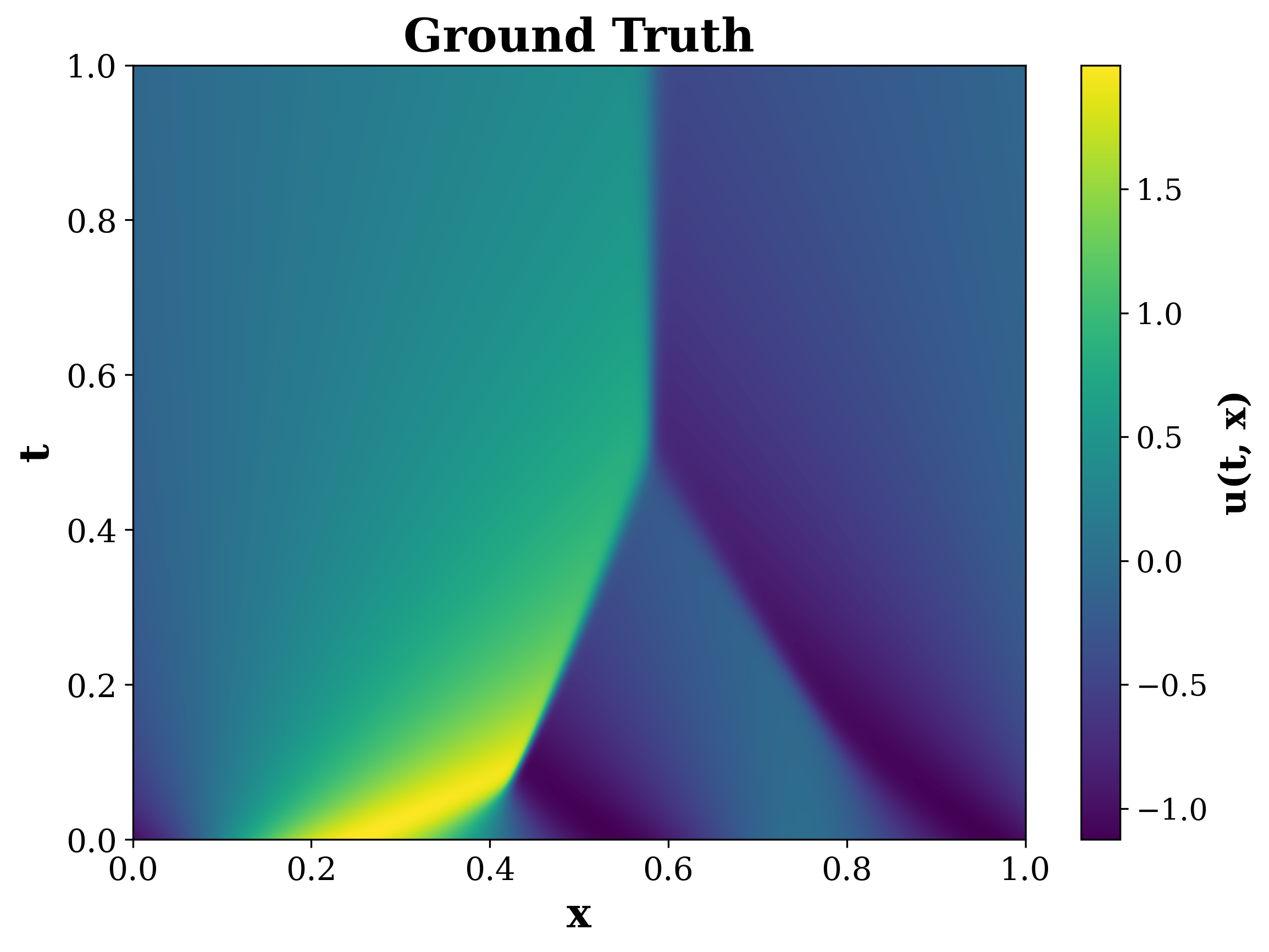}
\caption{Ground truth solution}
\end{subfigure}

\vspace{0.5cm}

\begin{subfigure}{0.9\linewidth}
\includegraphics[width=\linewidth]{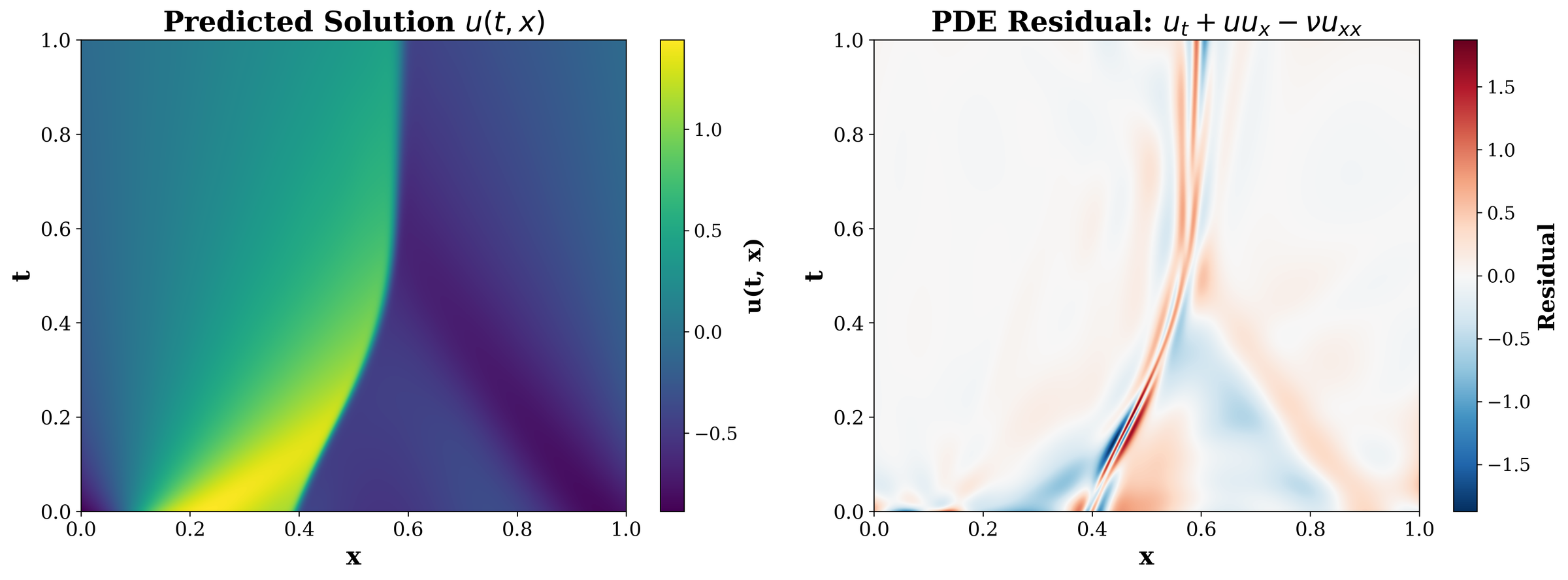}
\caption{Root solution (Score = 0.0449)}
\end{subfigure}

\vspace{0.5cm}

\begin{subfigure}{0.9\linewidth}
\includegraphics[width=\linewidth]{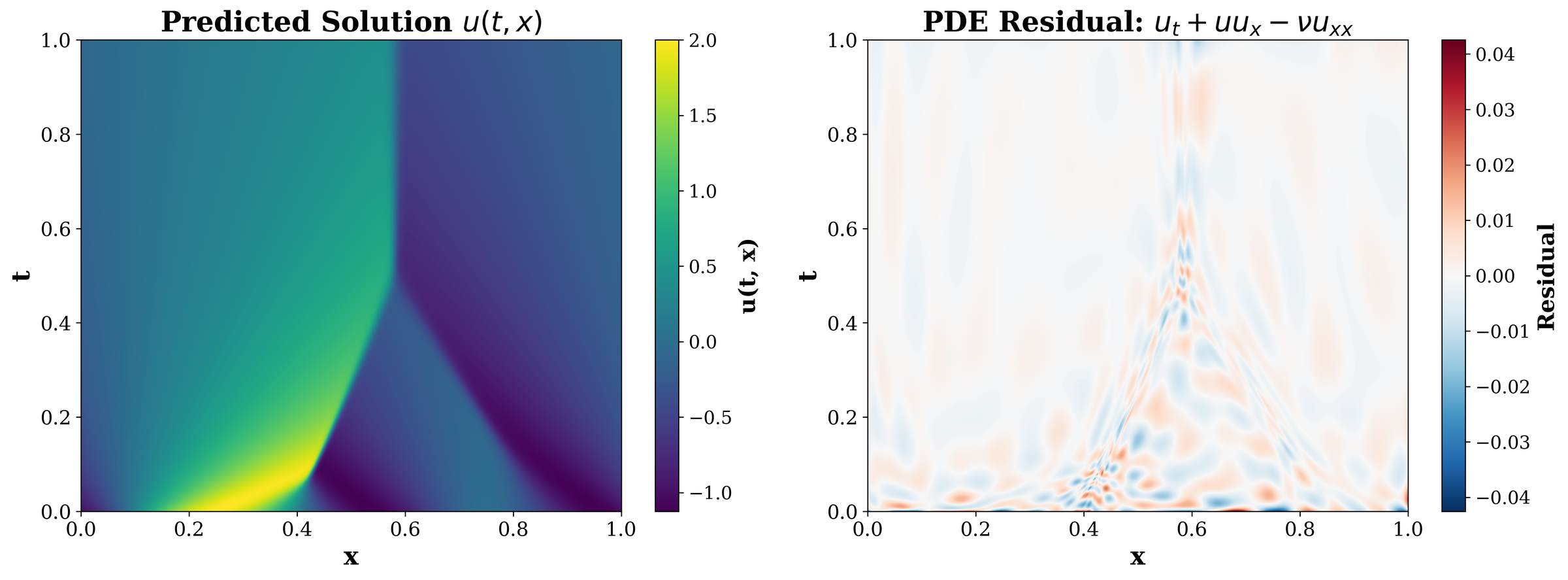}
\caption{Champion solution (Score = $4.02\times10^{-6}$)}
\end{subfigure}
\caption{Comparison of ground truth, root, and champion solutions for Burger's equation PINN. The score shows the combined PDE residual, initial condition, and boundary condition errors.}
\label{fig:burger_compare}
\end{figure}

\begin{figure}[h!]
\centering
\includegraphics[width=0.75\linewidth]{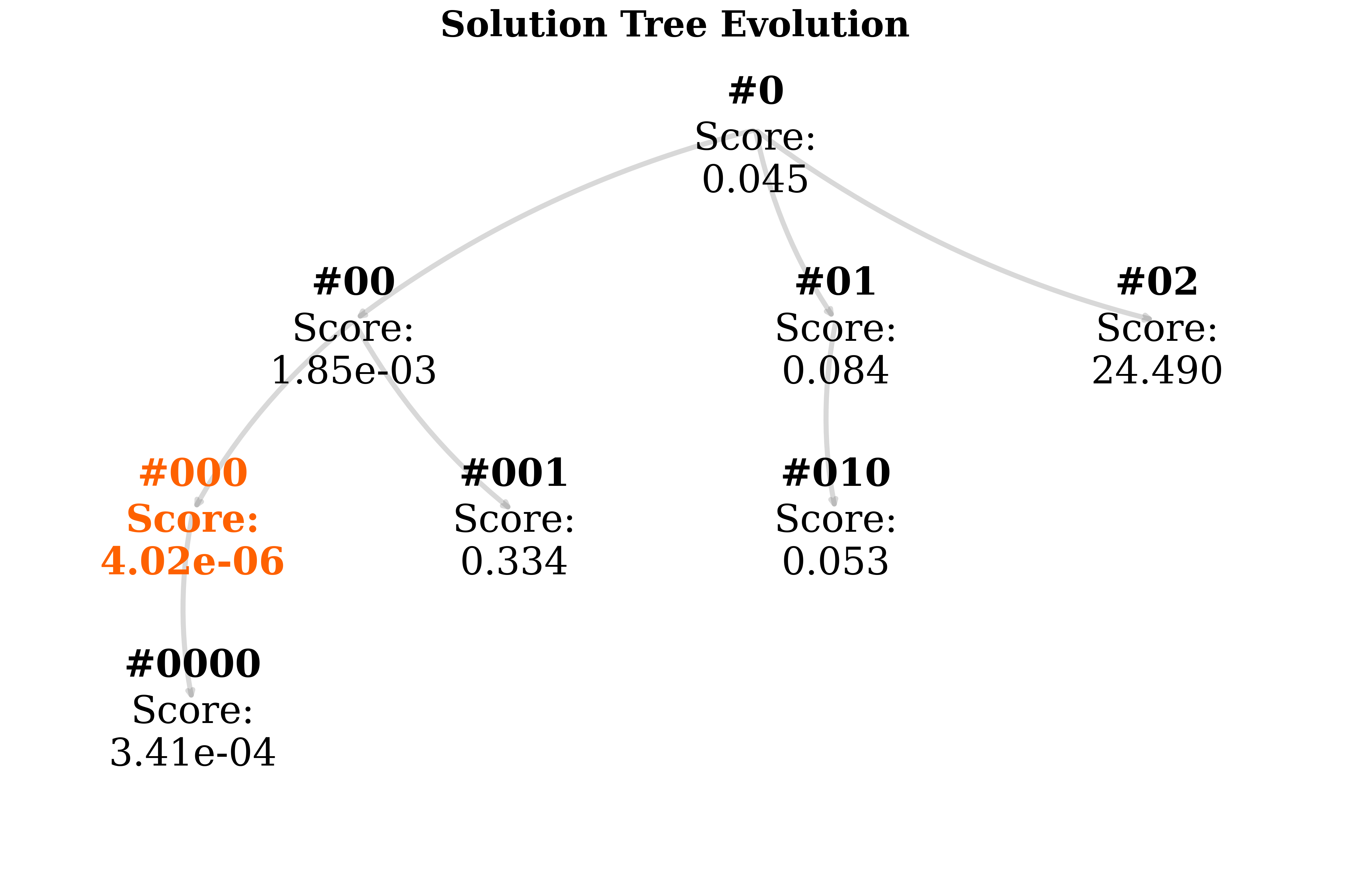}
\vspace{-1cm}
\caption{Evolution of solution tree for Burger's equation PINN. Each node of the tree represents a solution implemented and evaluated by the agents. The score shows the combined PDE residual, initial condition, and boundary condition errors. The best solution is colored in orange.}
\label{fig:burger_tree}
\end{figure}

\begin{figure}[h!]
\centering
\begin{subfigure}{0.48\linewidth}
\includegraphics[width=\linewidth]{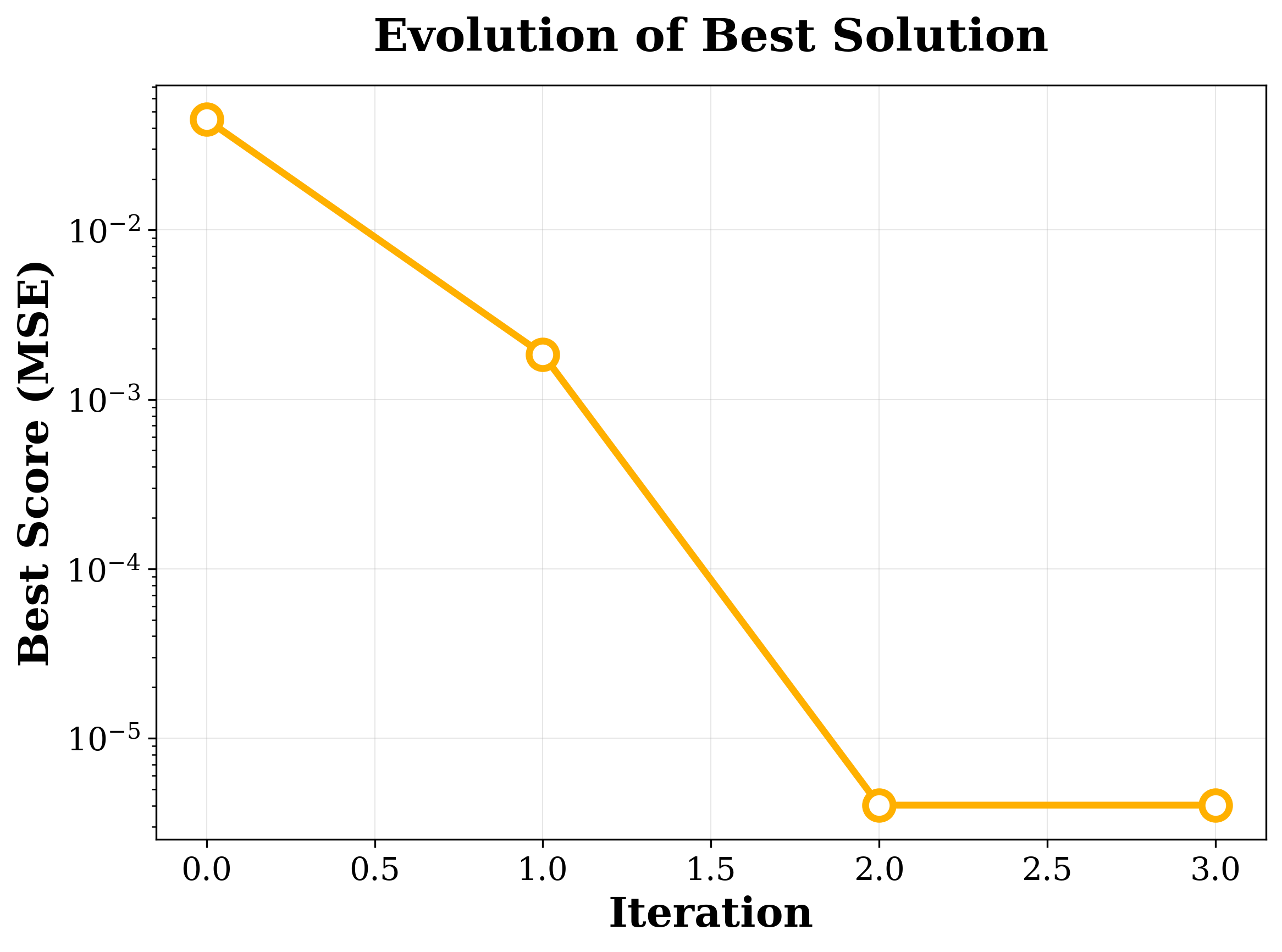}
\caption{Evolution of best solution}
\end{subfigure}
\hfill
\begin{subfigure}{0.48\linewidth}
\includegraphics[width=\linewidth]{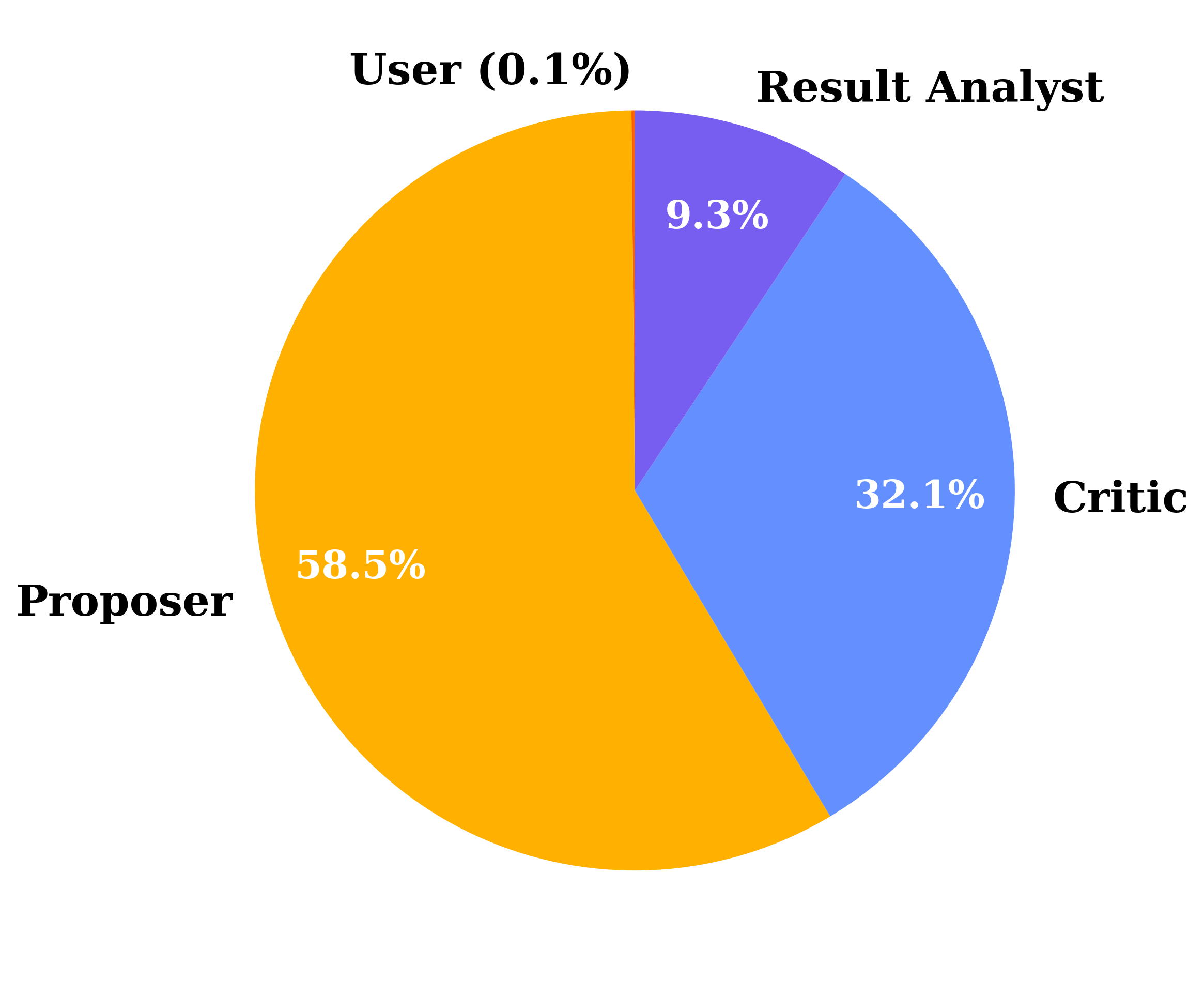}
\caption{Agent text contributions}
\end{subfigure}
\caption{Multi-agent evolution metrics for Burger's equation PINN. \textbf{Left:} Evolution of best solution score over iterations. \textbf{Right:} Text contributions by agent type, showing the proposer contributes the most while the human user contributes minimal text. Engineering agents and selector agents are excluded in this plot.}
\label{fig:burger_evolution}
\label{fig:burger_contrib}
\end{figure}

The evolutionary process generated 8 solutions over 3 iterations. The score used for ranking in this problem is defined as the combined PDE residual, initial condition, and boundary condition MSE. The champion solution achieved an 11,169$\times$ improvement from the root solution's score. Figure~\ref{fig:burger_compare} compares the ground truth solution obtained from FFT-based solver against the root and champion PINN solutions. The root solution is able to capture the general form of the solution, but details near sharp gradients are missed. The champion solution closely matches the ground truth even in regions with steep gradients. Figure~\ref{fig:burger_tree} visualizes the solution tree. Figure~\ref{fig:burger_evolution} shows the evolution of the best solution score across iterations with rapid convergence and agent text contributions.

\subsubsection{Novelty of the Champion Solution} \label{sec:burger_novelty}

The agents proposed an adaptive gradient-enhanced PINN based on the retrieved knowledge base entry ``Gradient-Enhanced Physics-Informed Neural Networks (gPINN)'' \cite{yu2022gradient}. The network architecture is a 6-block ResNet with Fourier embedding. The input $(t,x)$ is first embedded as $[t, \sin(2\pi x), \cos(2\pi x), x]$ to exploit the periodic boundary conditions.

Training proceeds in three phases. First, the PINN undergoes pretraining to fit only the initial and boundary conditions using Adam with learning rate $5 \times 10^{-3}$. No PDE residuals are enforced. 

Next, the PINN is trained with full loss for an additional 12,000 epochs. Here, the PDE residual $f = u_t + uu_x - \nu u_{xx}$ is differentiated to give $f_x = u_{tx} + u_x^2 + u u_{xx} - \nu u_{xxx}$. The loss function is $\mathcal{L} = \mathcal{L}_{\text{PDE}} + 10^{-4} \mathcal{L}_{\text{grad}} + \mathcal{L}_{\text{IC}} + \mathcal{L}_{\text{BC}}$ where:
\begin{align*}
\mathcal{L}_{\text{PDE}} &= \frac{1}{2N_{\text{coll}}} \sum_{i=1}^{N_{\text{coll}}} \lambda(t_i, x_i) \cdot f(t_i, x_i)^2, \\
\mathcal{L}_{\text{grad}} &= \frac{1}{2N_{\text{coll}}} \sum_{i=1}^{N_{\text{coll}}} \lambda(t_i, x_i) \cdot f_x(t_i, x_i)^2, \\
\mathcal{L}_{\text{IC}} &= \frac{1}{2N_{\text{IC}}} \sum_{j=1}^{N_{\text{IC}}} \lambda(0, x_j) \cdot [u(0, x_j) - u_0(x_j)]^2, \\
\mathcal{L}_{\text{BC}} &= \frac{1}{N_{\text{BC}}} \sum_{k=1}^{N_{\text{BC}}} [u(t_k, 0) - u(t_k, 1)]^2.
\end{align*}
Inspired by \cite{mcclenny2023self}, $\lambda(t,x)$ are per-point adaptive weights, which are parameterized by a small 3-layer MLP $\lambda_{\text{net}}$. For each point $(t,x)$, the network outputs $\lambda(t,x) = \text{clamp}(\text{softplus}(\lambda_{\text{net}}(t,x)) \cdot \alpha, 0.1, 100)$ where $\alpha = 1/\text{mean}(\lambda)$ normalizes the average weight to 1. The $\lambda_{\text{net}}$ is optimized via gradient ascent using a separate Adam optimizer that maximizes the weighted losses while the main model minimizes them. 

Finally, the solution uses residual-based adaptive refinement (RAR) to further finetune the model on high-error regions. Each RAR iteration generates candidate points via Latin Hypercube Sampling, evaluate $\max(|f|, |f_x|)$ at these candidates in batches of 1000, and add the top-50 highest-error points to the collocation set. After adding points, the model undergoes 1000 Adam epochs with $\lambda_{\text{net}}$ trainable followed by 500 L-BFGS iterations with $\lambda_{\text{net}}$ frozen. The L-BFGS optimization uses strong Wolfe line search and double precision.

\subsubsection{Token and Cost Analysis}

\begin{table}[htbp]
\centering
\caption{Token and cost analysis for Burgers' Equation PINN (physics-informed neural network, 3 iterations). The Grok-4 proposer generates 31\% of tokens at only 6\% of cost; the engineer accounts for 40\% of LLM cost.}
\label{tab:cost_burger}

\begin{tabular}{lrrrr}
\toprule
\textbf{Agent (Model)} & \textbf{API Calls} & \textbf{Tokens (M)} & \textbf{Cost (\$)} & \textbf{Time (s)} \\
\midrule
\multicolumn{5}{l}{\textit{Overall}} \\
\quad Total             &  79 & 1.60         &  2.07         & 5{,}198 \\
\quad LLM (multi-agent) &  79 & 1.60 &  2.07 & 2{,}101 \\
\quad GPU Training      & --  & --           & --            & 3{,}097 \\
\midrule
\multicolumn{5}{l}{\textit{Agent Breakdown}} \\
\quad Proposer (Grok-4)          & 24 & 0.50 (31\%) & 0.12  (6\%) &   481 (23\%) \\
\quad Analyst (Gemini-2.5-Pro)   &  7 & 0.17 (11\%) & 0.45 (22\%) &   275 (13\%) \\
\quad Engineer (Claude-Haiku)    & 15 & 0.33 (21\%) & 0.83 (40\%) &   647 (31\%) \\
\quad Retriever (Gemini-2.5-Pro) &  9 & 0.22 (14\%) & 0.49 (23\%) &   253 (12\%) \\
\quad Critic (GPT-4o-Mini)       & 16 & 0.30 (19\%) & 0.15  (7\%) &   347 (17\%) \\
\quad Debugger (GPT-4o-Mini)     &  8 & 0.08  (5\%) & 0.04  (2\%) &    97  (5\%) \\
\bottomrule
\end{tabular}

\end{table}

The Grok-4 proposer generates a large token volume (31\%) at very low per-token cost (6\% of total), while the engineer accounts for 40\% of total LLM cost. The debugger is called 8 times across only 3 iterations, indicating that the generated code required notable repair. At \$2.07 total LLM cost this is the least expensive run.

\clearpage

\subsection{Antiderivative Operator Learning} \label{sec:antideriv}

To test the agents' understanding of operator learning and neural operators, this problem requires building a neural operator that maps input functions to their antiderivatives.

\subsubsection{Problem Setup}

The user provided the following structured prompt to the system:

\begin{tcolorbox}[colback=problemcolor!10, colframe=problemcolor, title=Problem]
Learn a neural operator that maps input functions to their antiderivatives. Given an input function $f: [0, 1] \to \mathbb{R}$ sampled at 100 equispaced points, find the operator $\mathcal{G}$ such that $F'(x) = f(x)$.

\textit{Input:} Functions $f(x)$ sampled at 100 points.
\textit{Output:} Antiderivative functions $F(x)$ at the same 100 points.
\textit{Goal:} Generalize to unseen functions. Build a machine learning model, not a numerical integrator.
\end{tcolorbox}

\begin{tcolorbox}[colback=requirementscolor!10, colframe=requirementscolor, title=Requirements]
Use PyTorch.
\end{tcolorbox}

\begin{tcolorbox}[colback=evaluationcolor!10, colframe=evaluationcolor, title=Evaluation]
Average relative L2 error over 4000 validation samples (1000 each at 4 frequency levels: high, medium-high, medium-low, low).
$$\text{Score} = \frac{1}{N}\sum_{i=1}^N \frac{\|F_{\text{pred}} - F_{\text{true}}\|_2}{\|F_{\text{true}}\|_2}$$
\end{tcolorbox}

\begin{tcolorbox}[colback=datacolor!10, colframe=datacolor, title=Data, breakable]
\textbf{validation\_set:}
\begin{itemize}
\item filename: val\_data.pkl
\item description: 4 datasets containing functions of different properties. Each dataset contains 1000 function pairs (f, F) where F is the antiderivative of f. Each function is sampled at 100 equispaced points on [0, 1]. Total: 4000 samples.
\item loading\_instructions: import pickle\\
with open('val\_data.pkl', 'rb') as f:\\
\hspace{1em}data = pickle.load(f)\\
\# data = \{'0.05': \{...\}, '0.1': \{...\}, '0.2': \{...\}, '0.5': \{...\}\}\\
\# Each key contains: \{'values': (1000, 100), 'antiderivative': (1000, 100)\}\\
\# Extract specific dataset: dataset = data['0.05']\\
\# Or combine all: import numpy as np; all\_values = np.vstack([data[k]['values'] for k in data.keys()])
\end{itemize}
\end{tcolorbox}

The validation set is provided to the evaluator to help with model assessment, but it is not accessible to other agents during solution development. The validation set contains functions sampled from Gaussian Random Fields (GRFs) with four different length scales (0.05, 0.1, 0.2, 0.5), but this information is not provided to the agents. No training dataset is provided; the agents must generate their own training samples.

\subsubsection{Agents Setup}

Table~\ref{tab:antideriv_config} summarizes the agent configurations and evolutionary parameters used in this experiment.

\begin{table}[h!]
\centering
\caption{Agent configurations for antiderivative operator experiment}
\label{tab:antideriv_config}
\small
\begin{tabular}{lll}
\toprule
\textbf{Agent} & \textbf{Model} & \textbf{Temperature} \\
\midrule
root engineer & GPT-5 Mini & 0.0 \\
\midrule
data analyst & Gemini 2.5 Pro & 0.1 \\
evaluator & Claude Haiku 4.5 & 0.0 \\
retriever & Gemini 2.5 Pro & 0.0 \\
proposer & Gemini 2.5 Pro & 0.3 \\
critic & GPT-5 Mini & 0.1 \\
engineer & Claude Haiku 4.5 & 0.0 \\
debugger & GPT-5 Mini & 0.0 \\
result analyst & Gemini 2.5 Pro & 0.1 \\
selector & GPT-5 Mini, Grok-4 Fast, Gemini 2.5 Pro & - \\
\midrule
\multicolumn{3}{l}{\textbf{Evolutionary Parameters}} \\
\midrule
Max Iterations & 5 & \\
Parallel Mutations & 4 & \\
\bottomrule
\end{tabular}
\end{table}

\subsubsection{Results}

\begin{figure}[h!]
\centering
\includegraphics[width=0.75\linewidth]{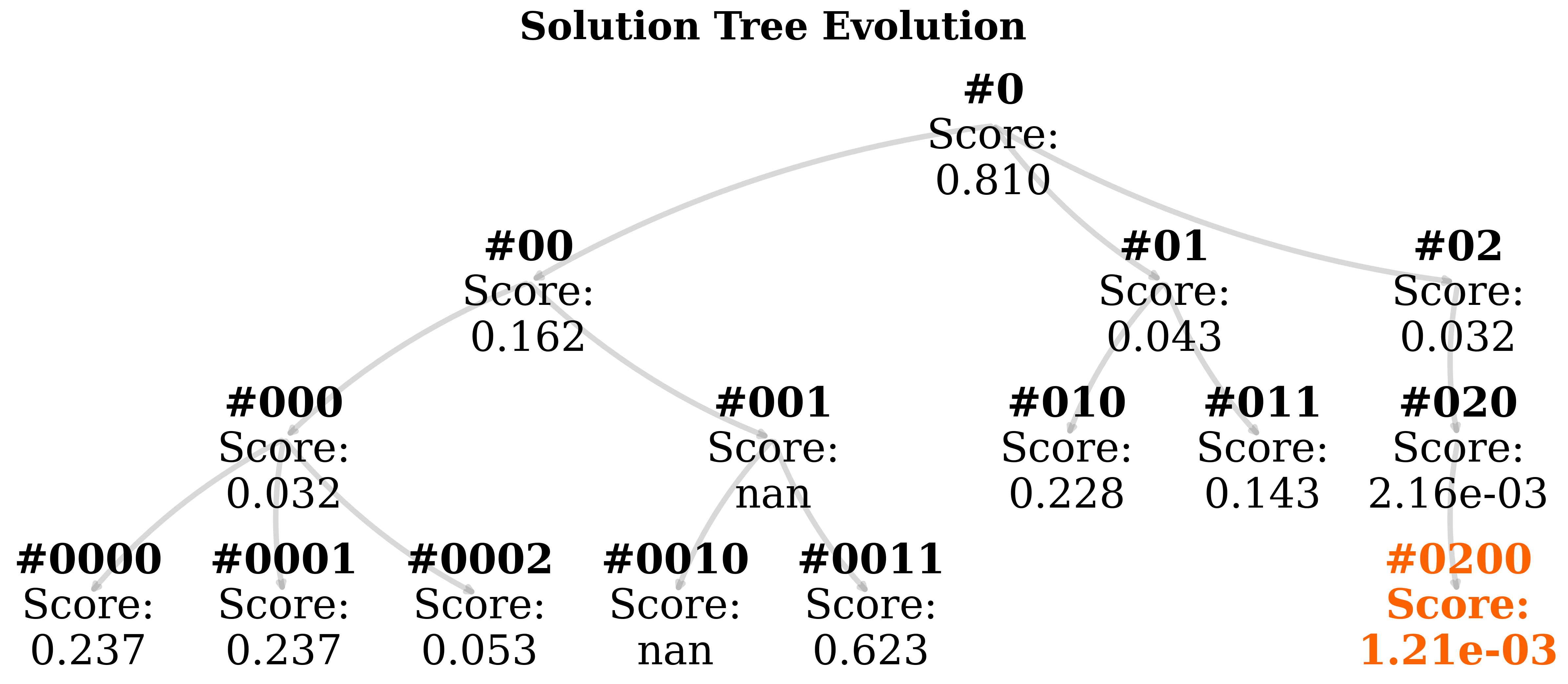}
\caption{Evolution of solution tree for antiderivative operator learning. Each node of the tree represents a solution implemented and evaluated by the agents. The score shows the average relative L2 error over the validation set. The best solution is colored in orange.}
\label{fig:antideriv_tree}
\end{figure}

\begin{figure}[h!]
\centering
\begin{subfigure}{0.48\linewidth}
\includegraphics[width=\linewidth]{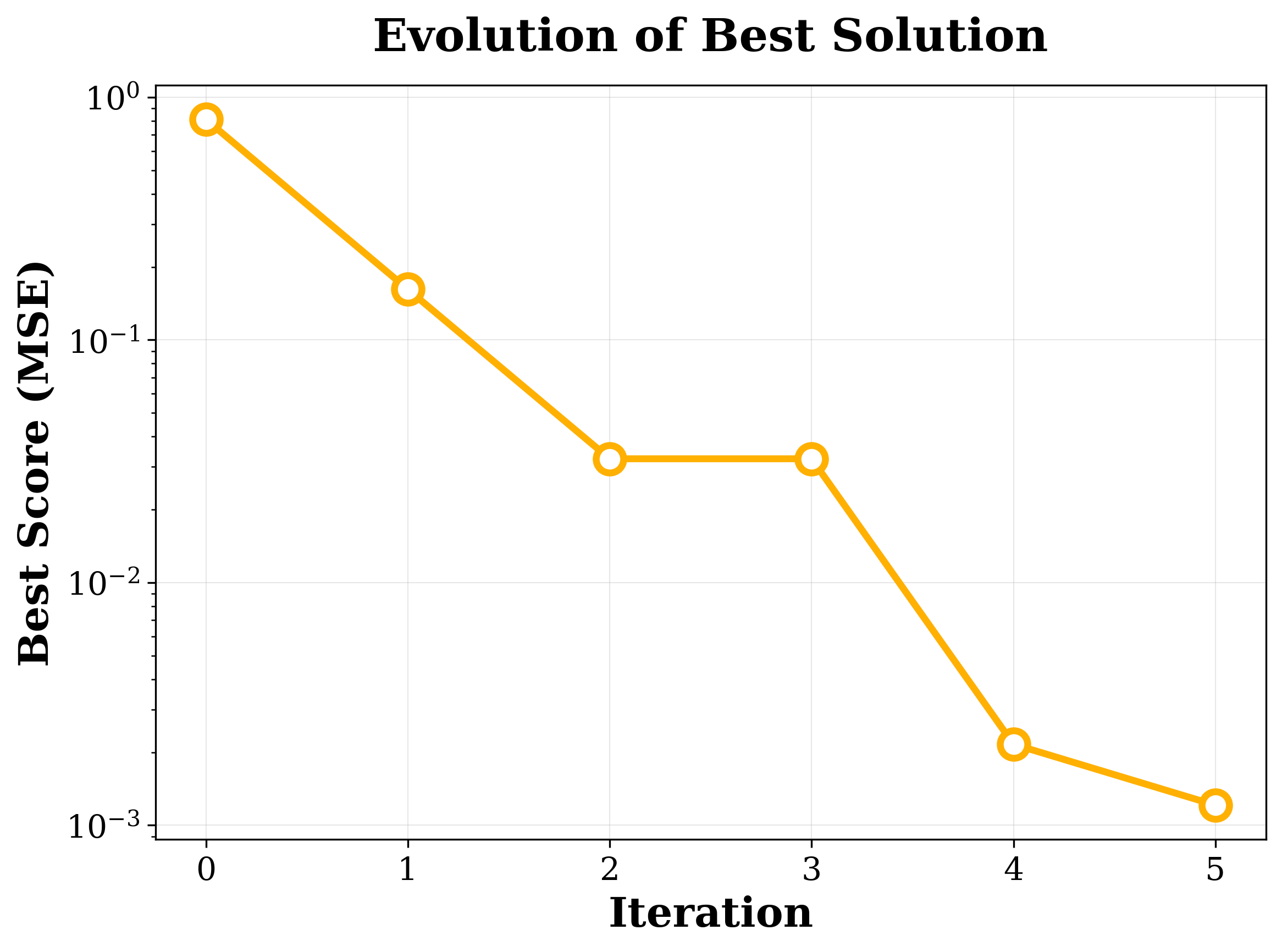}
\caption{Evolution of best solution}
\end{subfigure}
\hfill
\begin{subfigure}{0.48\linewidth}
\includegraphics[width=\linewidth]{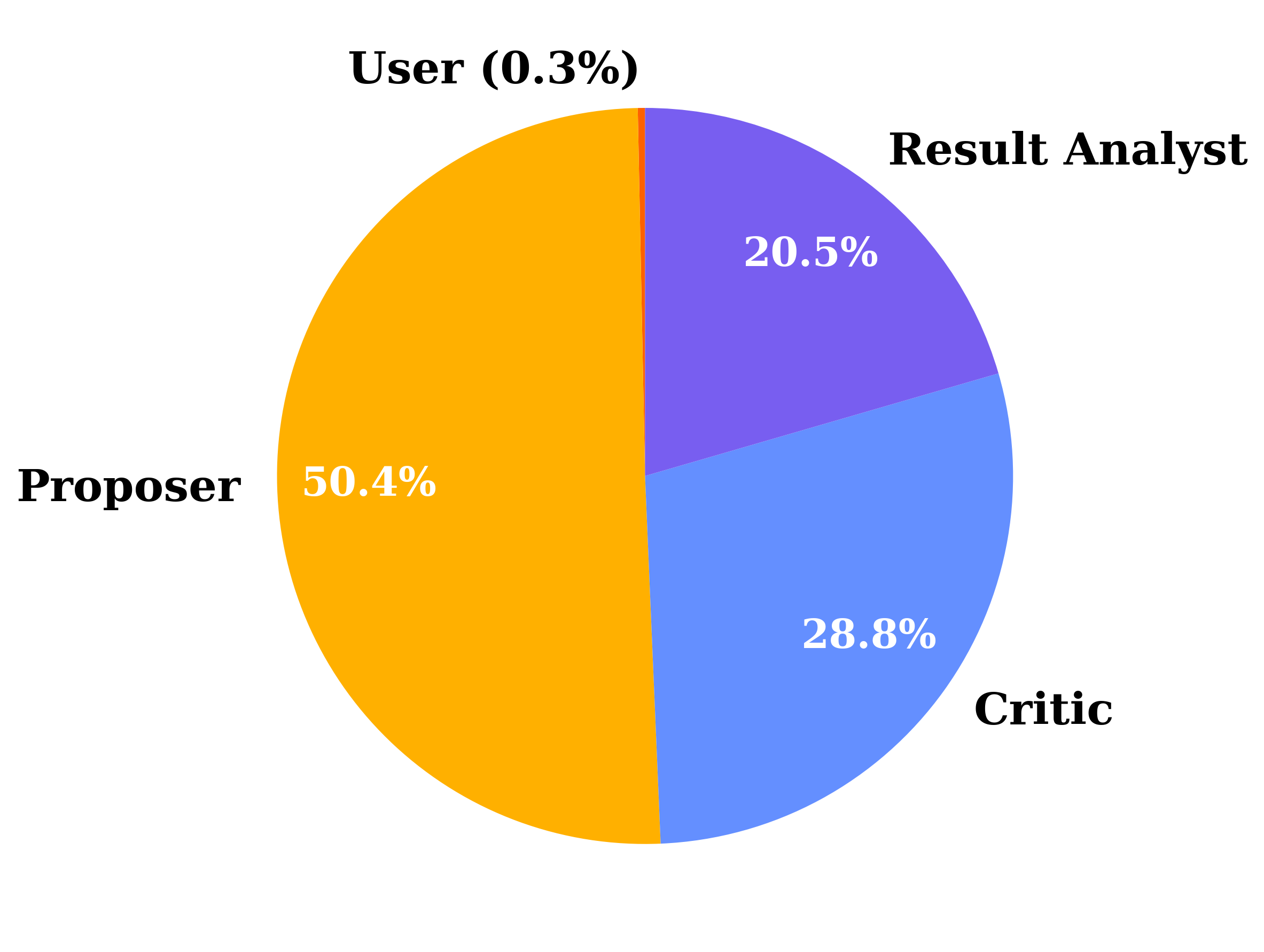}
\caption{Agent text contributions}
\end{subfigure}
\caption{Multi-agent evolution metrics for antiderivative operator learning. \textbf{Left:} Evolution of best solution score over iterations. \textbf{Right:} Text contributions by agent type, showing the proposer contributes the most while the human user contributes minimal text. Engineering agents and selector agents are excluded in this plot.}
\label{fig:antideriv_evolution}
\label{fig:antideriv_contrib}
\end{figure}

\begin{figure}[h!]
\centering
\includegraphics[width=0.95\linewidth]{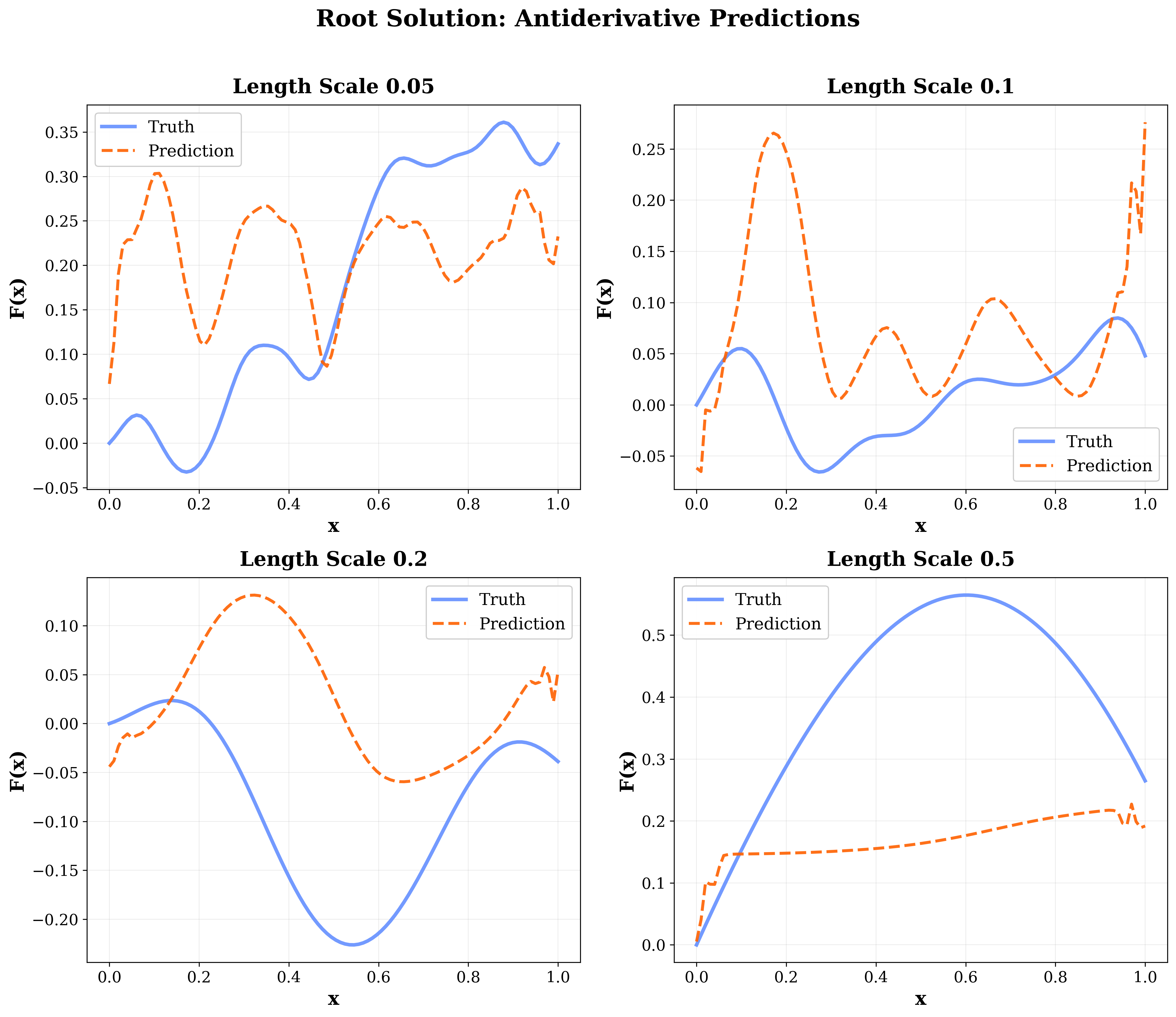}
\caption{Root solution predictions for antiderivative operator learning on the validation dataset where input functions are sampled using GRF with different length scales.}
\label{fig:antideriv_root_pred}
\end{figure}

\begin{figure}[h!]
\centering
\includegraphics[width=0.95\linewidth]{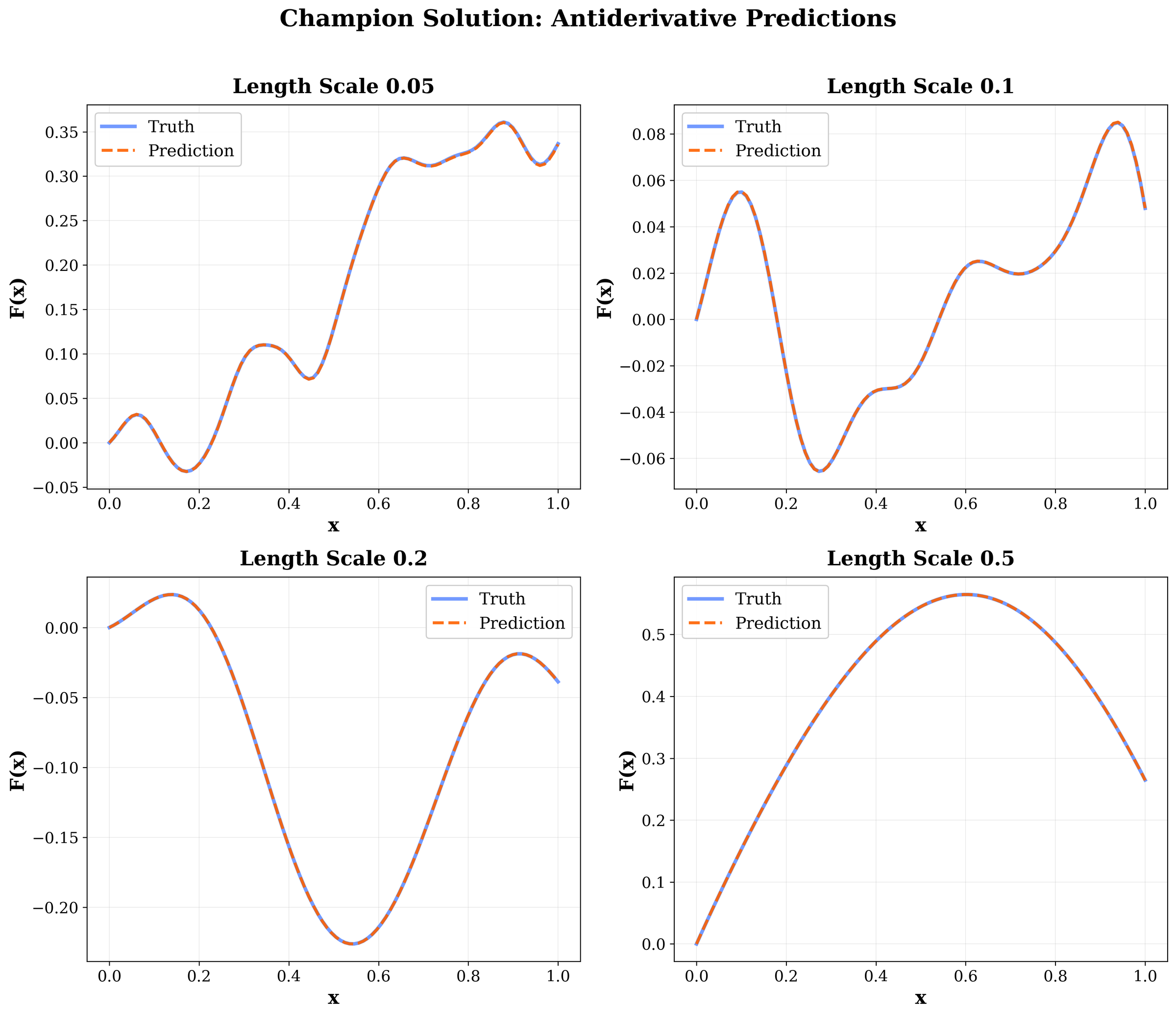}
\caption{Champion solution predictions for antiderivative operator learning on the validation dataset where input functions are sampled using GRF with different length scales.}
\label{fig:antideriv_champion_pred}
\end{figure}

\begin{figure}[h!]
\centering
\includegraphics[width=0.95\linewidth]{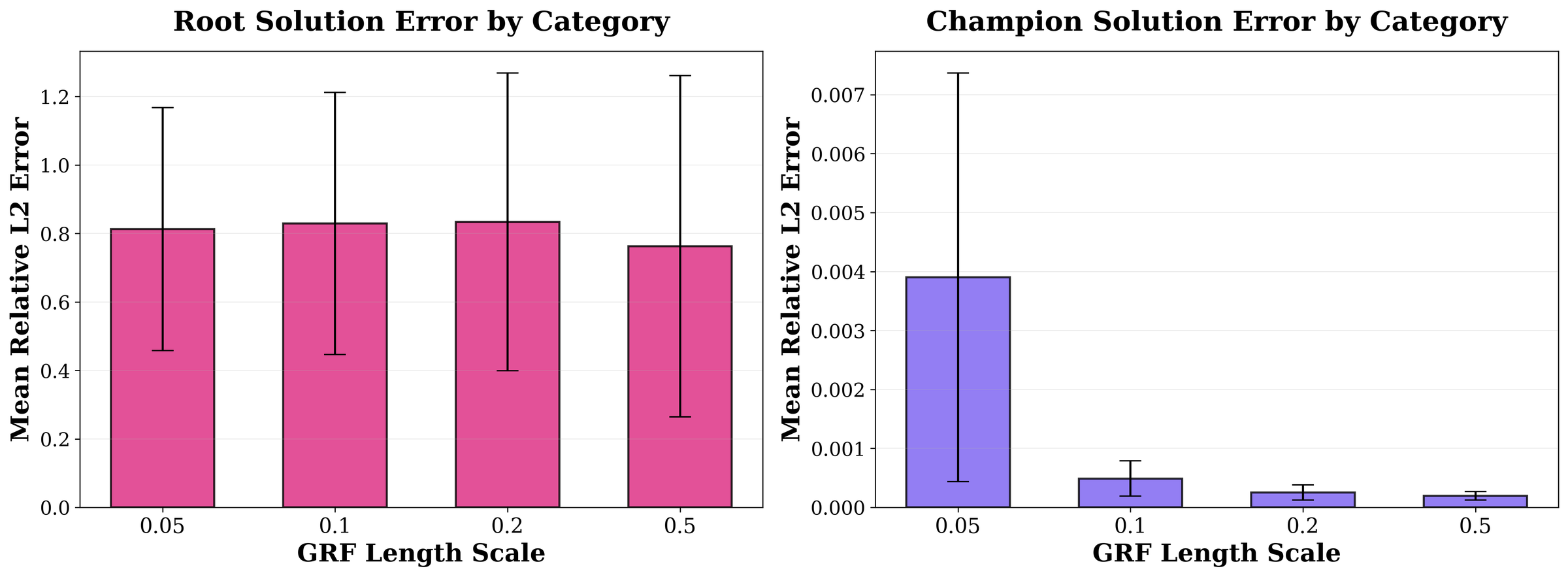}
\caption{Error comparison for antiderivative operator learning. \textbf{Left:} Root solution error distribution. \textbf{Right:} Champion solution error distribution. Champion solution exhibits significant improvement from the root across all frequency categories.}
\label{fig:antideriv_error}
\end{figure}

The evolutionary process generated 15 solutions over 5 iterations, achieving a 669$\times$ improvement from the root solution's relative L2 error of 0.810 to the champion solution's (solution\_0200) error of 0.00121. Figure~\ref{fig:antideriv_tree} visualizes the solution evolution tree showing all 15 solutions. Figure~\ref{fig:antideriv_evolution} shows the evolution of the best solution score and agent text contributions. Figure~\ref{fig:antideriv_root_pred} shows the root solution's predictions across four GRF length scales (0.05, 0.1, 0.2, 0.5), and Figure~\ref{fig:antideriv_champion_pred} shows the champion solution's predictions. Figure~\ref{fig:antideriv_error} compares mean relative L2 error across GRF length scales: the root solution shows high errors with high variance, while the champion solution achieves low errors, with functions generated from small length scale (0.05) being the most challenging case.

\subsubsection{Novelty of the Champion Solution} \label{sec:antideriv_novelty}

The solution uses a DeepONet consisting of two networks: a linear, bias-free branch network that maps the input function to a 128-dimensional latent space, and a trunk network with 4 hidden layers of 192 neurons each using AdaptiveGELU activations that maps spatial coordinates to the same 128-dimensional latent space. During the analysis phase, the proposer noted that 
\begin{quote}
  ``The integration operator $\mathcal{G}$ is linear. For any functions $f_1, f_2$ and scalars $\alpha, \beta$:
    $$
    \mathcal{G}(\alpha f_1 + \beta f_2) = \alpha \mathcal{G}(f_1) + \beta \mathcal{G}(f_2)
    $$
  A successful neural operator should respect this property.''
\end{quote}

Therefore, it decides to implement the unconventional choice of a linear branch network to enforce the linearity. Additionally, ``physics-informed DeepONet'' \cite{wang2021learning} retrieved from the knowledge base inspires the incorporation of a physics-informed loss
\begin{align*}
  \mathcal{L}_{\text{physics}} = \frac{1}{N \cdot 100} \sum_{i=1}^{N} \sum_{j=1}^{100} \left( \frac{dF^{(i)}}{dx}(x_j) - f^{(i)}(x_j) \right)^2
\end{align*}
into the training to enforce the differential relationship $F' = f$. 

Training employs a scheduled weighting strategy where the physics-informed loss weight starts at 0 and linearly increases to its final value of 0.1 over the first 50 epochs according to $w_{\text{physics}}(\text{epoch}) = 0.1 \cdot \min(1.0, \frac{\text{epoch} + 1}{50})$. This schedule is intended to let the model first establish a data fit before gradually introducing the physics constraint. The model is trained for 250 epochs using the AdamW optimizer with learning rate $10^{-3}$, weight decay $10^{-5}$, batch size 128, and a cosine annealing learning rate scheduler.

The agents generated their own training data using GRFs with length scales sampled from a log-uniform distribution on $[0.05, 0.5]$, giving more samples at smaller scales.

\subsubsection{Token and Cost Analysis}

\begin{table}[htbp]
\centering
\caption{Token and cost analysis for Antiderivative Operator Learning (neural operator, 5 iterations). GPU training time exceeds LLM time by 4.2$\times$; the proposer has the highest cost share (50\%) of any experiment.}
\label{tab:cost_antideriv}

\begin{tabular}{lrrrr}
\toprule
\textbf{Agent (Model)} & \textbf{API Calls} & \textbf{Tokens (M)} & \textbf{Cost (\$)} & \textbf{Time (s)} \\
\midrule
\multicolumn{5}{l}{\textit{Overall}} \\
\quad Total             & 130 & 3.29         &  6.75         & 22{,}906 \\
\quad LLM (multi-agent) & 130 & 3.29 &  6.75 &  4{,}361 \\
\quad GPU Training      & --  & --           & --            & 18{,}545 \\
\midrule
\multicolumn{5}{l}{\textit{Agent Breakdown}} \\
\quad Proposer (Gemini-2.5-Pro)  & 45 & 1.24 (38\%) & 3.36 (50\%) & 1{,}939 (44\%) \\
\quad Analyst (Gemini-2.5-Pro)   & 15 & 0.70 (21\%) & 1.39 (21\%) &   597 (14\%) \\
\quad Engineer (Claude-Haiku)    & 20 & 0.38 (12\%) & 0.91 (13\%) &   747 (17\%) \\
\quad Retriever (Gemini-2.5-Pro) & 15 & 0.37 (11\%) & 0.80 (12\%) &   391  (9\%) \\
\quad Critic (GPT-4o-Mini)       & 30 & 0.54 (16\%) & 0.26  (4\%) &   592 (14\%) \\
\quad Debugger (GPT-4o-Mini)     &  5 & 0.05  (2\%) & 0.03  (0\%) &    94  (2\%) \\
\bottomrule
\end{tabular}

\end{table}

GPU training time (5.15\,h) far exceeds LLM API time (1.21\,h) by 4.2$\times$. With only 5 iterations, the proposer contributes 50\% of LLM cost --- the highest proposer share across all experiments. Debugger calls remain light (5 calls), indicating stable neural operator code.

\clearpage
\subsection{Multiple-input Operator Learning for Reaction-Diffusion Equation} \label{sec:multiop}

This problem tests the agents' ability to design a neural operator that takes multiple input functions. We follow the same setup as \cite{jin2022mionet}, but this work is not included in the knowledge base. 

\subsubsection{Problem Setup}

The user provided the following structured prompt to the system:

\begin{tcolorbox}[colback=problemcolor!10, colframe=problemcolor, title=Problem]
Learn operator mapping two input functions to spatiotemporal solution.

PDE: $\frac{\partial u}{\partial t}= \frac{\partial }{\partial x} (k(x) \frac{\partial u}{\partial x}) + 0.01 \cdot u^2 + f(x)$,
$x \in [0,1]$, $t \in [0,1]$, $u(x,0) = 0$, $u(0,t) = u(1,t) = 0$.

\textit{Task:} Input two functions $k(x)$ and $f(x)$ (each 100 spatial points), output solution $u(x,t)$ on 100$\times$100 grid.

Note: This differs from standard single-input operator learning by having two input functions. Design a model that can handle multiple inputs.
\end{tcolorbox}

\begin{tcolorbox}[colback=requirementscolor!10, colframe=requirementscolor, title=Requirements]
Use PyTorch.
\end{tcolorbox}

\begin{tcolorbox}[colback=evaluationcolor!10, colframe=evaluationcolor, title=Evaluation]
Average relative L2 error: Score = $(1/N) \sum_i \|u_{\text{pred}} - u_{\text{true}}\|_{L^2} / \|u_{\text{true}}\|_{L^2}$
\end{tcolorbox}

\begin{tcolorbox}[colback=datacolor!10, colframe=datacolor, title=Data, breakable]
\textbf{training\_set:}
\begin{itemize}
\item filename: train\_data.npz
\item description: 1000 samples for diffusion-reaction operator learning. PDE: u\_t = (k(x)u\_x)\_x + 0.01·u² + f(x) on [0,1]×[0,1] with zero BC and IC. Both k(x) and f(x) are time-independent spatial functions from GRF (length scale 0.2). Shapes: k (1000, 100), f (1000, 100), u (1000, 100, 100), x (100,), t (100,). Solution u is in the shape of (num\_samples, time\_points, spatial\_points).
\item loading\_instructions: Use np.load('train\_data.npz'). Access via data['k'], data['f'], data['u'], data['x'], data['t']
\end{itemize}

\textbf{validation\_set:}
\begin{itemize}
\item filename: val\_data.npz
\item description: 200 samples for diffusion-reaction operator learning. PDE: u\_t = (k(x)u\_x)\_x + 0.01·u² + f(x) on [0,1]×[0,1] with zero BC and IC. Both k(x) and f(x) are time-independent spatial functions from GRF (length scale 0.2). Shapes: k (200, 100), f (200, 100), u (200, 100, 100), x (100,), t (100,). Solution u is in the shape of (num\_samples, time\_points, spatial\_points).
\item loading\_instructions: Use np.load('val\_data.npz'). Access via data['k'], data['f'], data['u'], data['x'], data['t']
\end{itemize}
\end{tcolorbox}

\subsubsection{Agents Setup}

Table~\ref{tab:multiop_config} summarizes the agent configurations and evolutionary parameters used in this experiment.

\begin{table}[h!]
\centering
\caption{Agent configurations for multi-input operator experiment}
\label{tab:multiop_config}
\small
\begin{tabular}{lll}
\toprule
\textbf{Agent} & \textbf{Model} & \textbf{Temperature} \\
\midrule
root engineer & GPT-5 Mini & 0.0 \\
\midrule
data analyst & Gemini 2.5 Flash & 0.1 \\
evaluator & Claude Haiku 4.5 & 0.0 \\
retriever & Gemini 2.5 Flash & 0.0 \\
proposer & Gemini 2.5 Flash & 0.9 \\
critic & GPT-5 Mini & 0.3 \\
engineer & Claude Haiku 4.5 & 0.0 \\
debugger & GPT-5 Mini & 0.0 \\
result analyst & Gemini 2.5 Flash & 0.1 \\
selector & GPT-5 Mini, Grok-4 Fast, Gemini 2.5 Flash & - \\
\midrule
\multicolumn{3}{l}{\textbf{Evolutionary Parameters}} \\
\midrule
Max Iterations & 8 & \\
Parallel Mutations & 4 & \\
\bottomrule
\end{tabular}
\end{table}

\subsubsection{Results}

\begin{figure}[h!]
\centering
\includegraphics[width=0.95\linewidth]{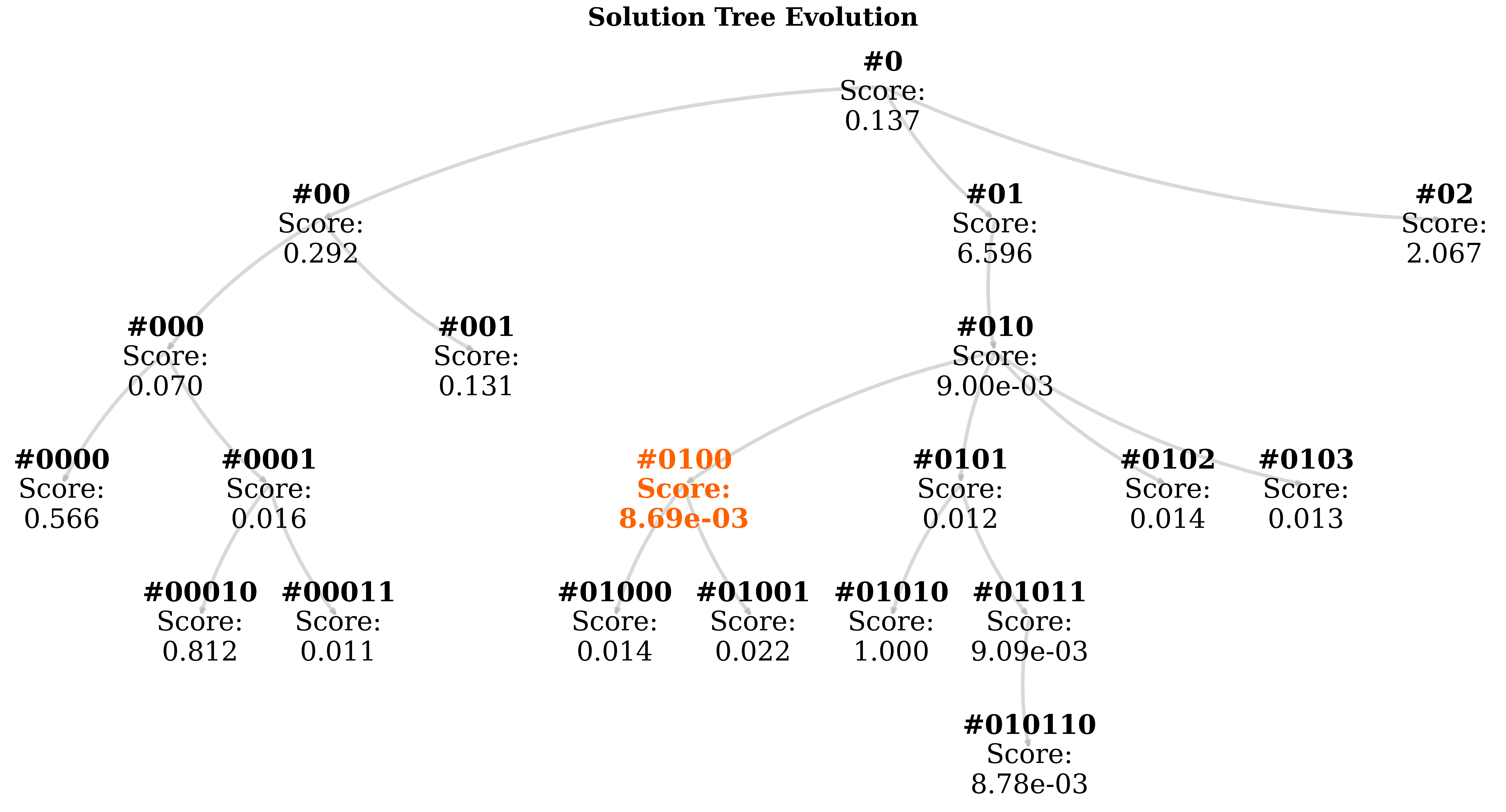}
\caption{Evolution of solution tree for multi-input operator learning for reaction-diffusion equation. Each node of the tree represents a solution implemented and evaluated by the agents. The score shows the average relative L2 error over the validation set. The best solution is colored in orange.}
\label{fig:multiop_tree}
\end{figure}

\begin{figure}[h!]
\centering
\begin{subfigure}{0.48\linewidth}
\includegraphics[width=\linewidth]{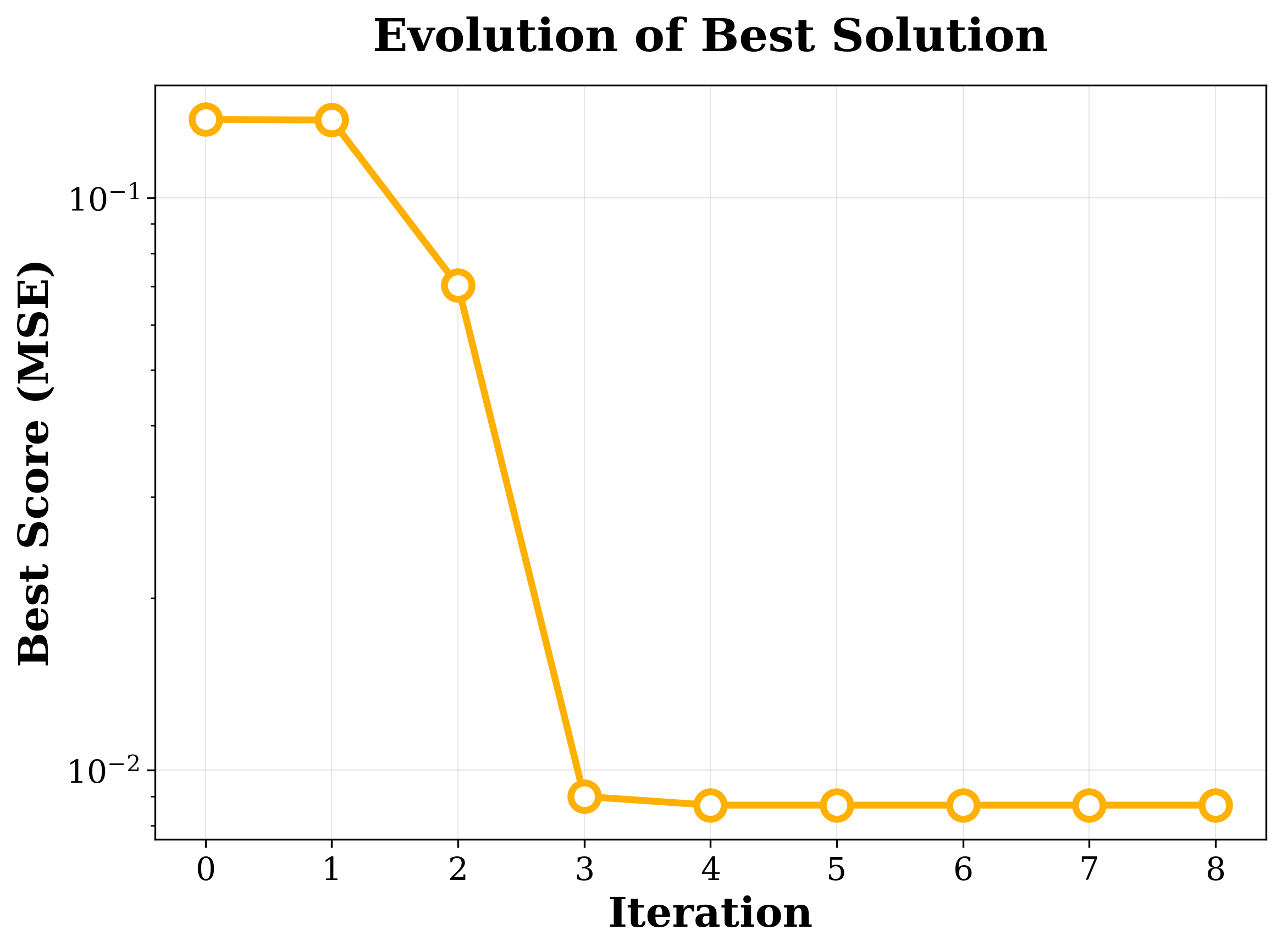}
\caption{Evolution of best solution}
\end{subfigure}
\hfill
\begin{subfigure}{0.48\linewidth}
\includegraphics[width=\linewidth]{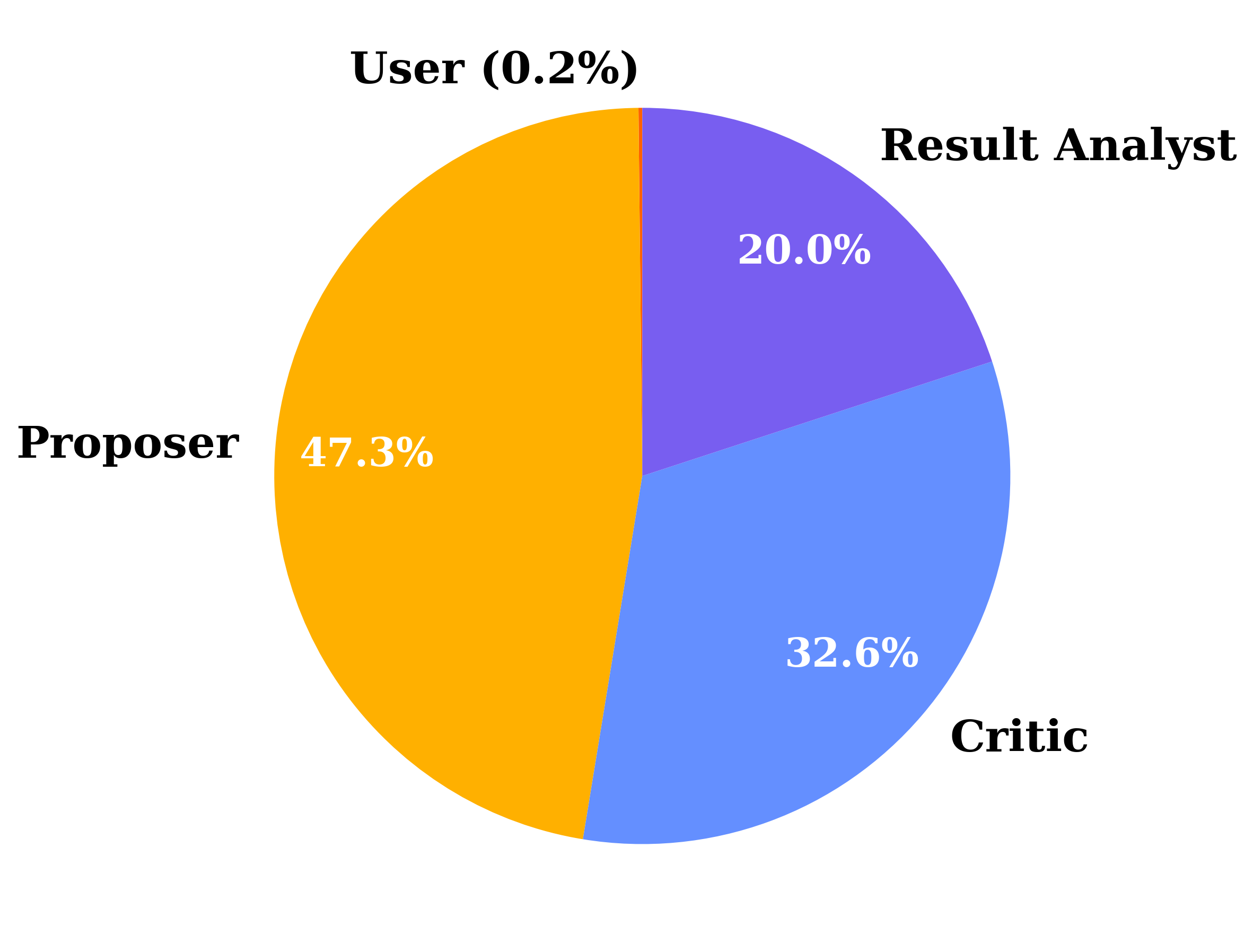}
\caption{Agent text contributions}
\end{subfigure}
\caption{Multi-agent evolution metrics for multi-input operator learning for reaction-diffusion equation. \textbf{Left:} Evolution of best solution score over iterations. \textbf{Right:} Text contributions by agent type, showing the proposer contributes the most while the human user contributes minimal text. Engineering agents and selector agents are excluded in this plot.}
\label{fig:multiop_evolution}
\label{fig:multiop_contrib}
\end{figure}

\begin{figure}[h!]
\centering
\includegraphics[width=0.95\linewidth]{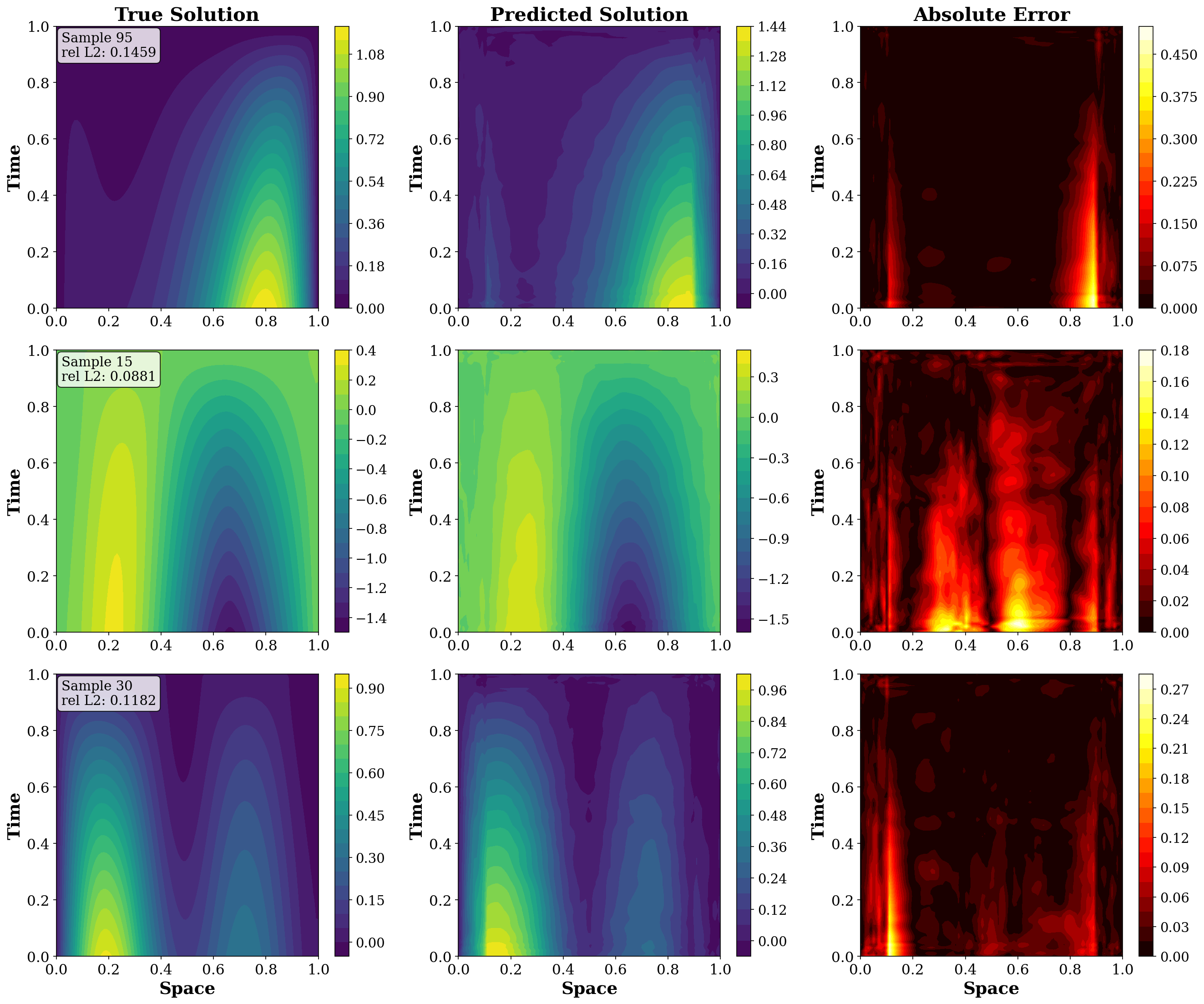}
\caption{Root solution predictions for multi-input operator learning on the validation dataset. The contours and magnitudes of the prediction are not accurate compared to the ground truth.}
\label{fig:multiop_root}
\end{figure}

\begin{figure}[h!]
\centering
\includegraphics[width=0.95\linewidth]{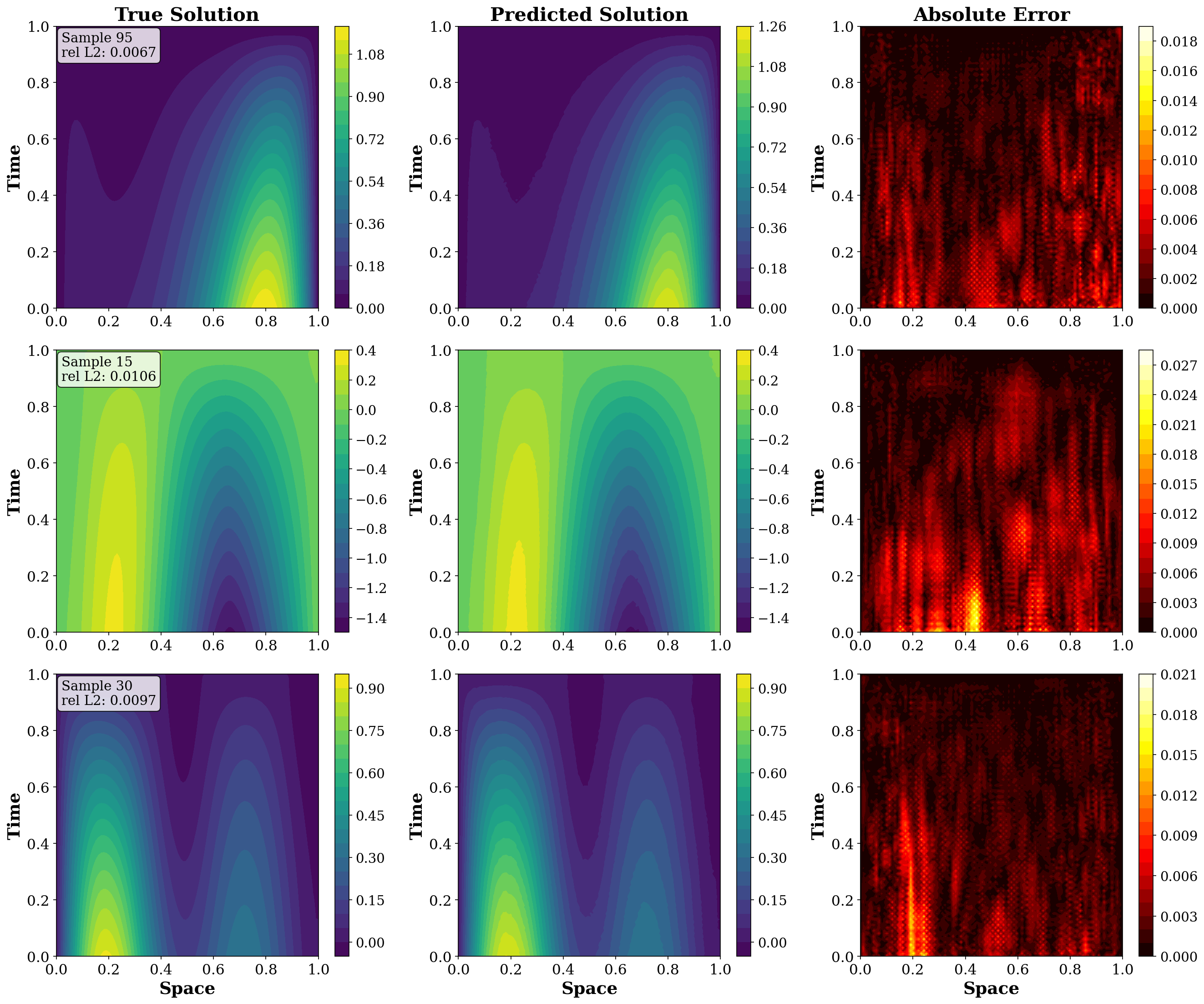}
\caption{Champion solution predictions for multi-input operator learning on the validation dataset. The prediction is very close to the ground truth.}
\label{fig:multiop_champion}
\end{figure}

The evolutionary process generated 20 solutions over 7 iterations, achieving a 15.6$\times$ improvement from the root solution's relative L2 error of 0.137 to the champion solution's (solution\_010110) error of 0.00878. Figure~\ref{fig:multiop_tree} shows the solution evolution tree showing all 20 solutions. Figure~\ref{fig:multiop_evolution} shows the evolution of the best score across iterations and agent text contributions. Figure~\ref{fig:multiop_root} shows the root solution's predictions on three validation samples. Figure~\ref{fig:multiop_champion} shows the champion solution's predictions on the same samples. Note that the immediate children of the root solution all degraded in performance. However, further mutations led to significant improvements, as the agents learn from failures and refine their designs. This shows the importance of not discarding solutions that initially perform poorly but keeping exploring their potential through further mutations.

\subsubsection{Novelty of the Champion Solution} \label{sec:multiop_novelty}

The solution employs a Fourier Neural Operator (FNO2d) architecture with 4 input channels ($k(x)$, $f(x)$, spatial coordinates $x$, and temporal coordinates $t$), 16 Fourier modes in both time and space dimensions, and 4 spectral layers. To handle the two input functions $k(x)$ and $f(x)$, the model expands each 1D spatial function into 2D spatiotemporal grids by replicating across the time dimension, then concatenates them with coordinate grids to form a 4-channel input tensor of shape $(N, 4, 100, 100)$. 

Inspired by a retrieved knowledge base entry on the training strategy of U-FNO for capturing sharp saturation fronts in CO2 storage simulations \cite{wen2022u}, the proposer introduced a derivative loss term to encourage the model to learn solutions more accurately in regions of high spatial variation. The spatial derivative $\partial u/\partial x$ is approximated using a central finite difference scheme $(\partial u/\partial x)_i \approx (u_{i+1} - u_{i-1}) / (2\Delta x)$. The derivative loss is computed as the relative loss
\begin{align*}
  L_{\text{deriv}} = \frac{\sum_{i=1}^{N_{\text{interior}}} [(\partial u_{\text{pred}}/\partial x)_i - (\partial u_{\text{true}}/\partial x)_i]^2}{\sum_{i=1}^{N_{\text{interior}}} (\partial u_{\text{true}}/\partial x)_i^2 + 10^{-8}}
\end{align*}
where the summation is over all interior spatial points (excluding boundaries) across all samples and time steps. The total loss function is given as 
\begin{align*}
  \mathcal{L} = \frac{1}{N_{\text{data}}}\sum_{i=1}^{N_{\text{data}}} [u_{\text{pred},i} - u_{\text{true},i}]^2 + 0.5 \cdot L_{\text{deriv}}
\end{align*}
The training process consists of 100 epochs with batch size 32, Adam optimizer with learning rate $10^{-3}$, and ReduceLROnPlateau scheduler (factor 0.5, patience 5). Boundary and initial conditions are hard-enforced by directly setting $u(x, 0) = 0$, $u(0, t) = 0$, and $u(1, t) = 0$ on the FNO output before returning predictions.

\subsubsection{Token and Cost Analysis}

\begin{table}[htbp]
\centering
\caption{Token and cost analysis for Multi-input Operator Learning for Diffusion-Reaction PDE (neural operator, 8 iterations). GPU training time dominates at 10.7\,h vs.\ 2.1\,h LLM time; the Gemini-2.5-Flash model enables cost-effective scaling to 8 iterations at \$6.73 total.}
\label{tab:cost_multiop}

\begin{tabular}{lrrrr}
\toprule
\textbf{Agent (Model)} & \textbf{API Calls} & \textbf{Tokens (M)} & \textbf{Cost (\$)} & \textbf{Time (s)} \\
\midrule
\multicolumn{5}{l}{\textit{Overall}} \\
\quad Total             & 204 & 8.10         &  6.73         & 46{,}248 \\
\quad LLM (multi-agent) & 204 & 8.10 &  6.73 &  7{,}713 \\
\quad GPU Training      & --  & --           & --            & 38{,}535 \\
\midrule
\multicolumn{5}{l}{\textit{Agent Breakdown}} \\
\quad Proposer (Gemini-2.5-Flash)  & 72 & 2.68 (33\%) & 2.72 (40\%) & 3{,}644 (47\%) \\
\quad Analyst (Gemini-2.5-Flash)   & 18 & 2.96 (36\%) & 1.10 (16\%) &   502  (7\%) \\
\quad Engineer (Claude-Haiku)      & 33 & 0.80 (10\%) & 1.96 (29\%) & 1{,}597 (21\%) \\
\quad Retriever (Gemini-2.5-Flash) & 24 & 0.70  (9\%) & 0.49  (7\%) &   573  (7\%) \\
\quad Critic (GPT-4o-Mini)         & 48 & 0.85 (11\%) & 0.42  (6\%) & 1{,}277 (17\%) \\
\quad Debugger (GPT-4o-Mini)       &  9 & 0.11  (1\%) & 0.05  (1\%) &   122  (2\%) \\
\bottomrule
\end{tabular}

\end{table}

At 8 iterations and 204 LLM calls, this is the longest run; GPU training time (10.71\,h) far exceeds LLM API time (2.14\,h). The analyst incurs 36\% of all tokens --- the highest analyst share --- due to multimodal outputs for the multi-input setting. Despite the longest run length, total LLM cost (\$6.73) remains moderate, enabled by the lightweight Gemini-2.5-Flash model.

\clearpage

\subsection{Reconstruction of 2D Cylinder Wake Vorticity Field from Sparse Observations} \label{sec:cylinder_wake}

This problem evaluates the agents' capability to design a model architecture that can reconstruct a physical field from sparse and noisy sensor measurements. We use the set up and dataset from \cite{zhao2024recfno} and \cite{fukami2021global}, but these works are not included in the agents' knowledge base. 

\subsubsection{Problem Setup}

The user provided the following structured prompt to the system:

\begin{tcolorbox}[colback=problemcolor!10, colframe=problemcolor, title=Problem]
Reconstruct the complete 2D vorticity field from sparse noisy sensor measurements in unsteady cylinder wake flow.

\textit{Physical System:} The cylinder wake is a well-known benchmark in fluid mechanics characterized by periodic laminar flow vortex shedding. The flow field is obtained by solving the incompressible Navier-Stokes equations at Reynolds number Re = 100. The computational domain is rectangular with the cylinder body positioned within it, and the dataset contains 5000 snapshots spanning four vortex shedding periods.

\textit{Task:} Input: Noisy vorticity measurements from 4 fixed sensor locations distributed across the flow field. Output: Complete vorticity field reconstructed at all 112$\times$192 grid points in the spatial domain. Objective: Learn the mapping from sparse noisy observations (4 points) to the global physical field (21,504 grid points), achieving accurate reconstruction that captures the complex vortex dynamics despite measurement noise.

\textit{Challenge:} Sparse Sensors: Only 4 measurement points to reconstruct 21,504 grid points (5,376$\times$ expansion ratio). Measurement Noise: Additive White Gaussian Noise (AWGN) with SNR=30dB is added to sensor observations.
\end{tcolorbox}

\begin{tcolorbox}[colback=requirementscolor!10, colframe=requirementscolor, title=Requirements]
Use PyTorch.
\end{tcolorbox}

\begin{tcolorbox}[colback=evaluationcolor!10, colframe=evaluationcolor, title=Evaluation]
The reconstruction performance is evaluated using the relative L2 error: Relative L2 Error = $\|u - u_{\text{pred}}\|_2 / \|u\|_2$. Final Score: Average relative L2 error over all test snapshots (750 snapshots).
\end{tcolorbox}

\begin{tcolorbox}[colback=datacolor!10, colframe=datacolor, title=Data, breakable]
\textbf{training\_set:}
\begin{itemize}
\item filename: train\_data.npy
\item description: 3500 samples for noisy cylinder wake vorticity reconstruction. Flow at Re=100 with periodic vortex shedding. AWGN noise with SNR=30dB added to all data. 4 fixed sensors in wake region at positions [56,37], [53,40], [59,43], [50,47] in (h,w) coordinates. Shapes: sparse\_observations (3500, 1, 112, 192) with noisy values at 4 sensor locations, full\_fields (3500, 1, 112, 192) complete noisy vorticity fields, coordinates (2, 112, 192) where dim0=[x\_grid, y\_grid] normalized to [0,1], sensor\_positions (4, 2) array of [h\_idx, w\_idx] for each sensor.
\item loading\_instructions: Use np.load('train\_data.npy', allow\_pickle=True).item(). Returns dictionary with keys: 'sparse\_observations', 'full\_fields', 'coordinates', 'sensor\_positions'. Access via data['sparse\_observations'][i] for i-th sample's noisy sparse field, data['full\_fields'][i] for noisy ground truth. Coordinates are shared: data['coordinates'][0] is x-grid, data['coordinates'][1] is y-grid. Input to model: concatenate [sparse\_observations[i], coordinates] along channel dimension to get (3, 112, 192) where channels are [sparse\_noisy\_vorticity, x\_coords, y\_coords]. Target: data['full\_fields'][i, 0] shape (112, 192) noisy field.
\end{itemize}

\textbf{validation\_set:}
\begin{itemize}
\item filename: val\_data.npy
\item description: 750 samples for noisy cylinder wake vorticity reconstruction. Flow at Re=100 with periodic vortex shedding. AWGN noise with SNR=30dB added to all data. 4 fixed sensors in wake region at positions [56,37], [53,40], [59,43], [50,47] in (h,w) coordinates. Shapes: sparse\_observations (750, 1, 112, 192) with noisy values at 4 sensor locations, full\_fields (750, 1, 112, 192) complete noisy vorticity fields, coordinates (2, 112, 192) where dim0=[x\_grid, y\_grid] normalized to [0,1], sensor\_positions (4, 2) array of [h\_idx, w\_idx] for each sensor.
\item loading\_instructions: Use np.load('val\_data.npy', allow\_pickle=True).item(). Returns dictionary with keys: 'sparse\_observations', 'full\_fields', 'coordinates', 'sensor\_positions'. Access via data['sparse\_observations'][i] for i-th sample's noisy sparse field, data['full\_fields'][i] for noisy ground truth. Coordinates are shared: data['coordinates'][0] is x-grid, data['coordinates'][1] is y-grid. Input to model: concatenate [sparse\_observations[i], coordinates] along channel dimension to get (3, 112, 192) where channels are [sparse\_noisy\_vorticity, x\_coords, y\_coords]. Target: data['full\_fields'][i, 0] shape (112, 192) noisy field.
\end{itemize}
\end{tcolorbox}

\subsubsection{Agents Setup}

Table~\ref{tab:cylinder_config} summarizes the agent configurations and evolutionary parameters used in this experiment.

\begin{table}[h!]
\centering
\caption{Agent configurations for cylinder reconstruction experiment}
\label{tab:cylinder_config}
\small
\begin{tabular}{lll}
\toprule
\textbf{Agent} & \textbf{Model} & \textbf{Temperature} \\
\midrule
root engineer & Claude Haiku 4.5 & 0.0 \\
\midrule
data analyst & Gemini 2.5 Flash & 0.1 \\
evaluator & Claude Haiku 4.5 & 0.0 \\
retriever & Gemini 2.5 Flash & 0.0 \\
proposer & Grok-4 Fast Reasoning & 0.1 \\
critic & GPT-5 Mini & 0.1 \\
engineer & Claude Haiku 4.5 & 0.0 \\
debugger & GPT-5 Mini & 0.0 \\
result analyst & Gemini 2.5 Flash & 0.1 \\
selector & GPT-5 Mini, Grok-4 Fast, Gemini 2.5 Flash & - \\
\midrule
\multicolumn{3}{l}{\textbf{Evolutionary Parameters}} \\
\midrule
Max Iterations & 5 & \\
Parallel Mutations & 4 & \\
\bottomrule
\end{tabular}
\end{table}

\subsubsection{Results}

\begin{figure}[h!]
\centering
\includegraphics[width=0.75\linewidth]{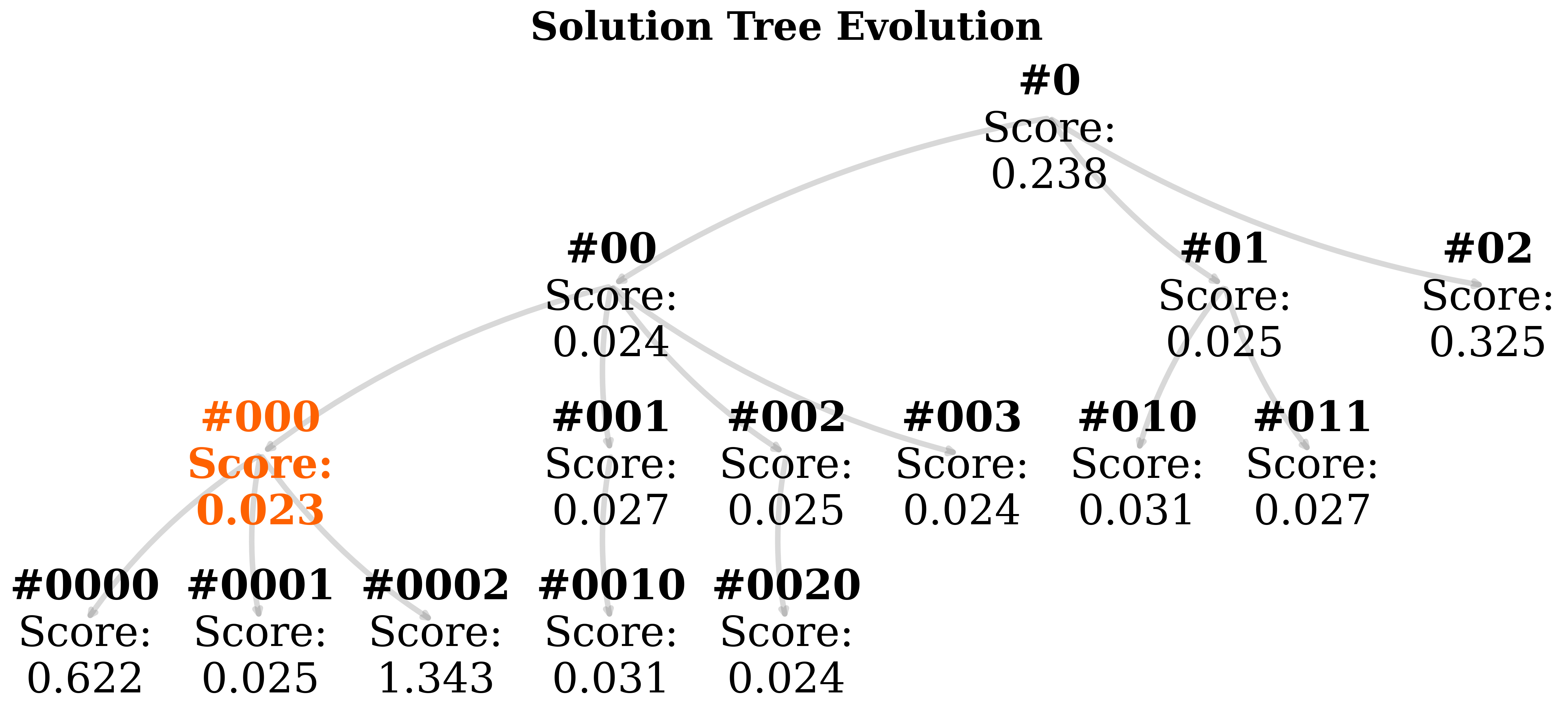}
\caption{Evolution of solution tree for cylinder wake vorticity field reconstruction from sparse noisy observations. Each node of the tree represents a solution implemented and evaluated by the agents. The score shows the relative L2 error on the validation set. The best solution is colored in orange.}
\label{fig:cylinder_tree}
\end{figure}

\begin{figure}[h!]
\centering
\begin{subfigure}{0.48\linewidth}
\includegraphics[width=\linewidth]{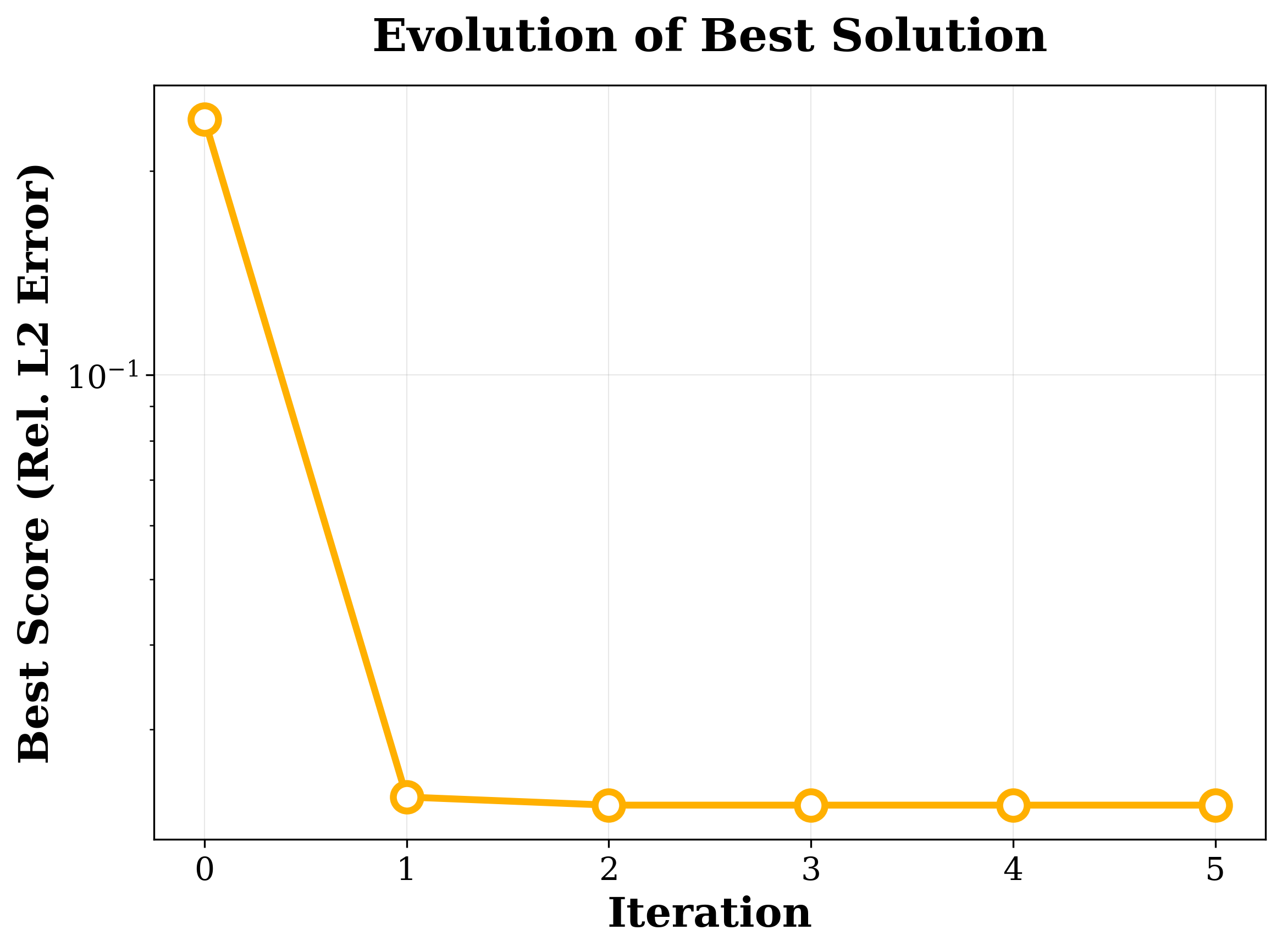}
\caption{Evolution of best solution}
\end{subfigure}
\hfill
\begin{subfigure}{0.48\linewidth}
\includegraphics[width=\linewidth]{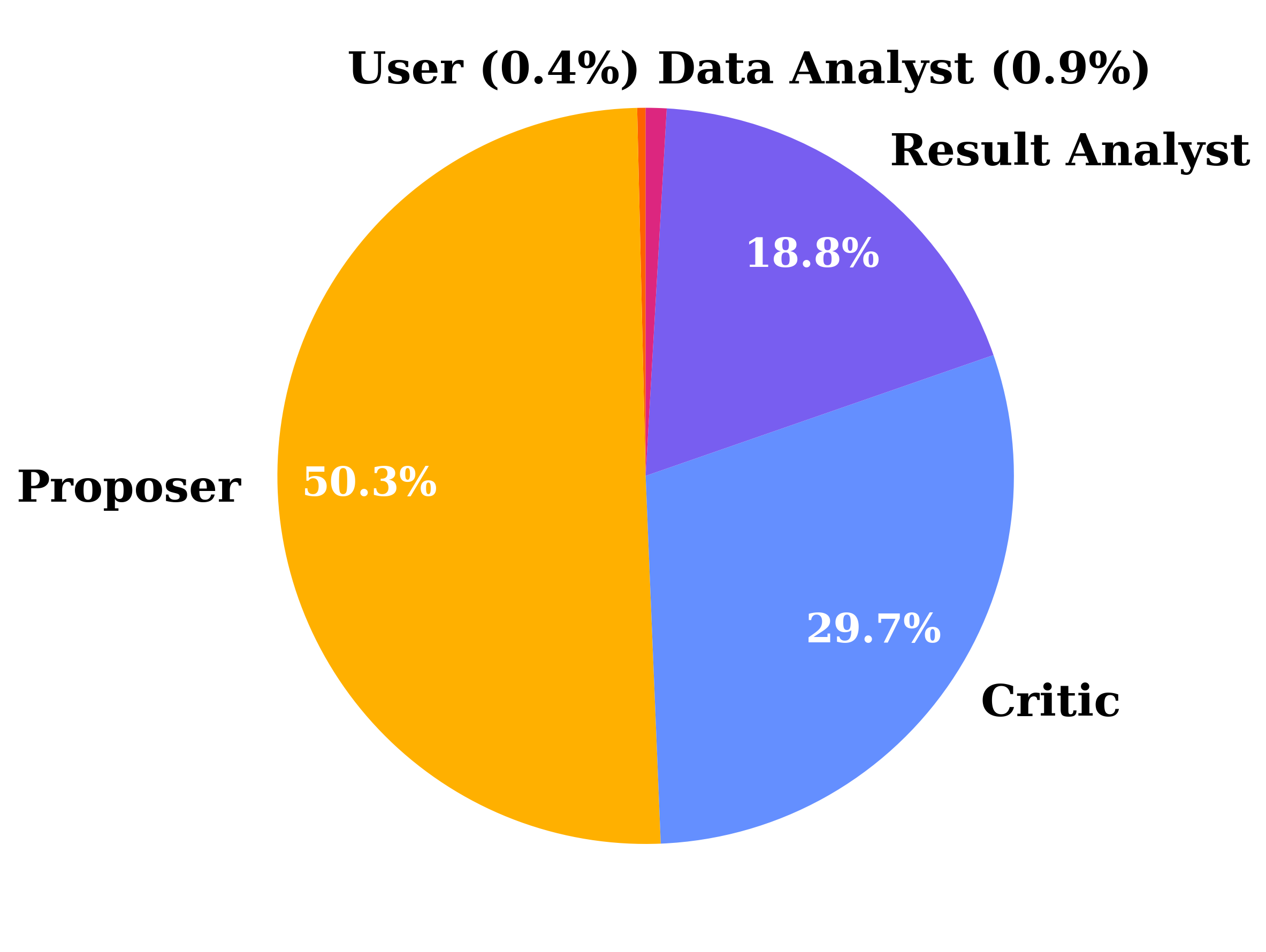}
\caption{Agent text contributions}
\end{subfigure}
\caption{Multi-agent evolution metrics for cylinder wake vorticity field reconstruction from sparse noisy observations. \textbf{Left:} Evolution of best solution score over iterations. \textbf{Right:} Text contributions by agent type, showing the proposer contributes the most while the human user contributes minimal text. Engineering agents and selector agents are excluded in this plot.}
\label{fig:cylinder_evolution}
\label{fig:cylinder_contrib}
\end{figure}

\begin{figure}[h!]
\centering
\begin{subfigure}{0.60\linewidth}
\includegraphics[width=\linewidth]{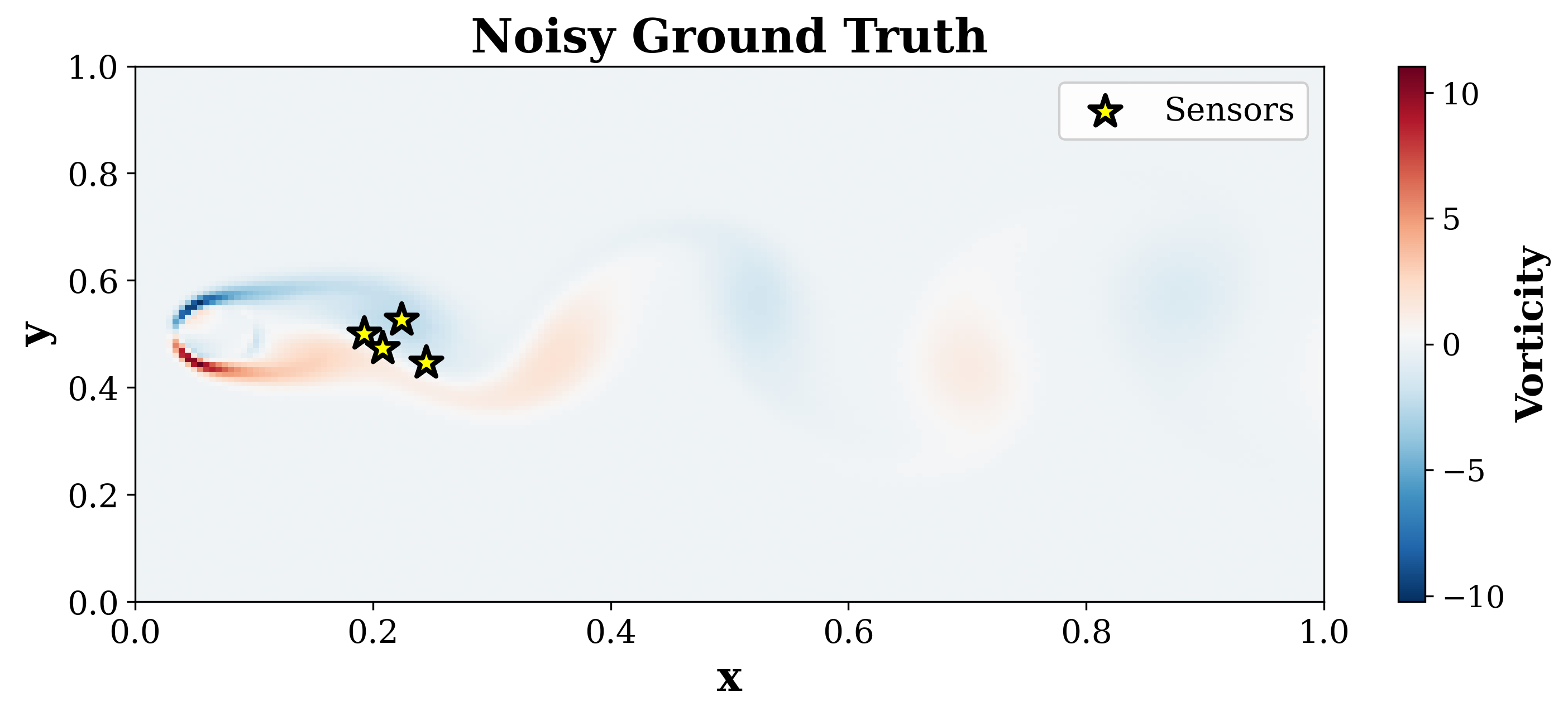}
\caption{Noisy ground truth with sensor positions}
\end{subfigure}

\vspace{0.5cm}

\begin{subfigure}{0.95\linewidth}
\includegraphics[width=\linewidth]{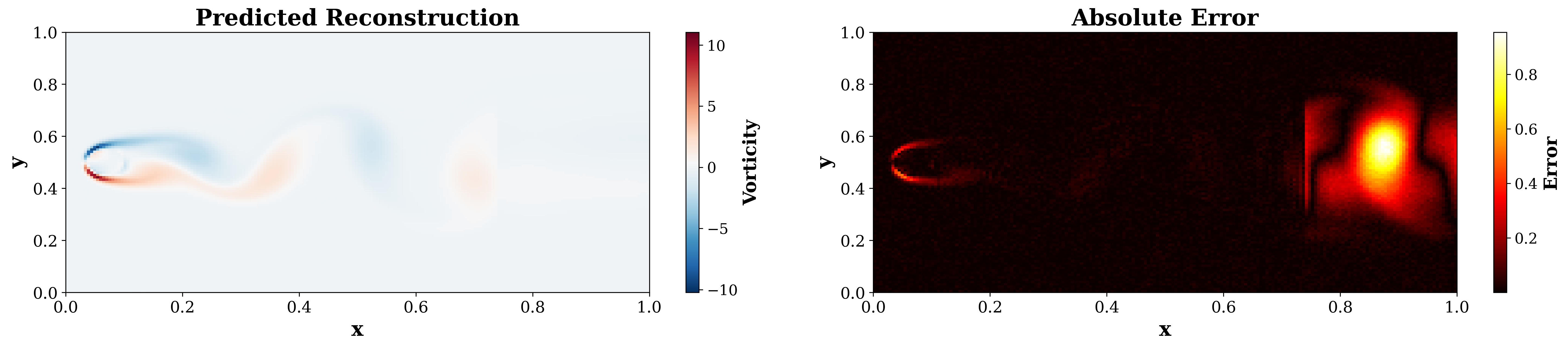}
\caption{Root solution}
\end{subfigure}

\vspace{0.5cm}

\begin{subfigure}{0.95\linewidth}
\includegraphics[width=\linewidth]{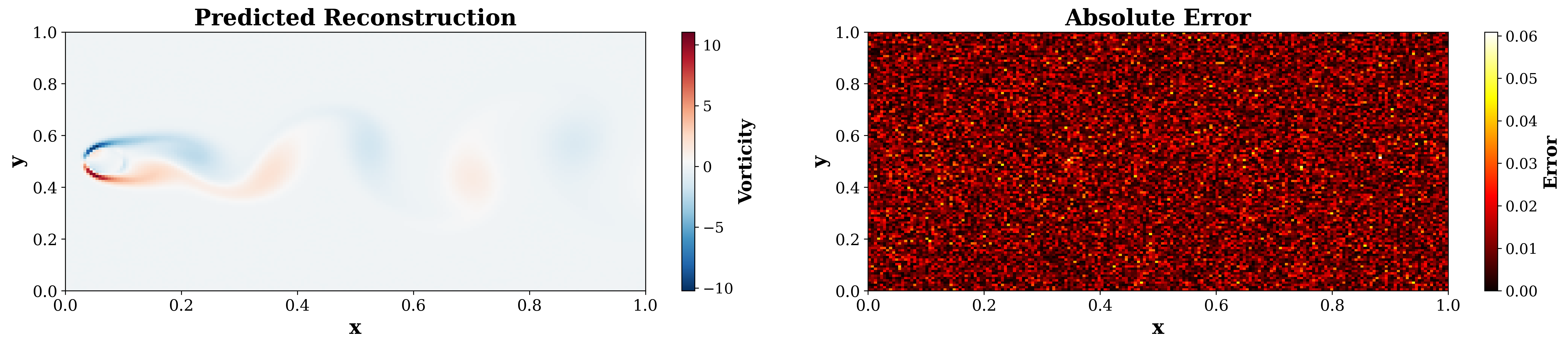}
\caption{Champion solution}
\end{subfigure}
\caption{Comparison of ground truth, root, and champion solutions for cylinder wake vorticity field reconstruction.}
\label{fig:cylinder_compare}
\end{figure}

The evolutionary process generated 15 solutions over 4 iterations, achieving a 10.3$\times$ improvement from the root solution's relative L2 error of 0.238 to the champion solution's (solution\_000) error of 0.0232. Figure~\ref{fig:cylinder_tree} visualizes the solution evolution tree showing all 15 solutions. Figure~\ref{fig:cylinder_evolution} shows the evolution of the best solution score across iterations and agent text contributions. Figure~\ref{fig:cylinder_compare} shows the ground truth and the root and champion predictions for a validation sample. 

\subsubsection{Novelty of the Champion Solution} \label{sec:cylinder_novelty}

The solution combines the U-FNO architecture \cite{wen2022u} with a bandlimit-preserving filter in the U-Net component inspired by Convolutional Neural Operators (CNO) \cite{raonic2023convolutional} to mitigate aliasing effects during decoder upsampling. This strategy is different from the ones in \cite{zhao2024recfno} and \cite{fukami2021global} which inspired this benchmark problem. 

Specifically, the U-FNO modifies the standard FNO by appending a mini U-Net path to the Fourier layer. The model takes 3 input channels (sparse noisy vorticity observations, $x$-coordinate grid, and $y$-coordinate grid) and reconstructs the full 112$\times$192 vorticity field. The architecture consists of a lifting layer projecting inputs to width 32, two pure Fourier blocks with SpectralConv2d operations, and two U-Fourier fusion layers that concatenate spectral output, bias projection, and U-Net output through 1$\times$1 convolutions before a final projection to the output channel. 

The key modification is a filtering implemented in the U-Net's upsampling stage, inspired by the Convolutional Neural Operator. It is intended to address aliasing when upsampling operations reconstruct fine-scale features from coarse representations. The bandlimit-preserving activation is implemented in the U-Net through four stages: upsample by factor 2 using nearest-neighbor interpolation, apply LeakyReLU with negative slope 0.1, convolve with a Gaussian low-pass filter ($g(x,y) = \exp(-(x^2+y^2)/(2\sigma^2))$ with kernel size $7\times7$ and $\sigma=1.0$, ), and downsample back to original resolution using area interpolation.

The loss function combines three weighted terms:
\begin{align*}
  \mathcal{L} = 1.0 \cdot \frac{\|u_{\text{pred}} - u_{\text{true}}\|_2}{\|u_{\text{true}}\|_2} + 0.1 \cdot \frac{1}{N}\sum_{i=1}^{N} |u_{\text{pred},i} - u_{\text{true},i}| + 0.3 \cdot \frac{\|\nabla u_{\text{pred}} - \nabla u_{\text{true}}\|_2}{\|\nabla u_{\text{true}}\|_2}
\end{align*}
where $N = 112 \times 192$ is the total number of grid points. The L1 term is designed to provide robustness to noise (SNR=30dB) and outliers. The gradient computation uses central finite differences instead of automatic differentiation. The proposer agent decides to add this gradient term to help capture sharp vorticity gradients at vortex core boundaries.

\subsubsection{Token and Cost Analysis}

\begin{table}[htbp]
\centering
\caption{Token and cost analysis for Cylinder Wake Flow Reconstruction (inverse problem, 5 iterations). The engineer accounts for 71\% of LLM cost due to the complexity of the reconstruction pipeline; the debugger is invoked 31 times --- the most of any experiment.}
\label{tab:cost_cylinder}

\begin{tabular}{lrrrr}
\toprule
\textbf{Agent (Model)} & \textbf{API Calls} & \textbf{Tokens (M)} & \textbf{Cost (\$)} & \textbf{Time (s)} \\
\midrule
\multicolumn{5}{l}{\textit{Overall}} \\
\quad Total             & 182 & 4.99         &  4.88         & 27{,}929 \\
\quad LLM (multi-agent) & 182 & 4.99 &  4.88 &  5{,}390 \\
\quad GPU Training      & --  & --           & --            & 22{,}539 \\
\midrule
\multicolumn{5}{l}{\textit{Agent Breakdown}} \\
\quad Proposer (Grok-4)            & 45 & 1.14 (23\%) & 0.27  (5\%) & 1{,}008 (19\%) \\
\quad Analyst (Gemini-2.5-Flash)   & 15 & 0.90 (18\%) & 0.41  (8\%) &   305  (6\%) \\
\quad Engineer (Claude-Haiku)      & 46 & 1.40 (28\%) & 3.44 (71\%) & 2{,}769 (51\%) \\
\quad Retriever (Gemini-2.5-Flash) & 15 & 0.46  (9\%) & 0.30  (6\%) &   345  (6\%) \\
\quad Critic (GPT-4o-Mini)         & 30 & 0.67 (14\%) & 0.30  (6\%) &   647 (12\%) \\
\quad Debugger (GPT-4o-Mini)       & 31 & 0.42  (8\%) & 0.16  (3\%) &   316  (6\%) \\
\bottomrule
\end{tabular}

\end{table}

The engineer accounts for 71\% of total LLM cost and 51\% of LLM time across 46 calls --- a sharp reversal of the proposer-dominant pattern in all other experiments --- reflecting the difficulty of setting up the inverse reconstruction pipeline. The debugger is invoked 31 times, confirming the cylinder codebase required the most iterative repair. The Grok-4 proposer again generates a large token share (23\%) at minimal cost (5\%), consistent with the Burgers' experiment.

\clearpage

\section{Knowledge Base} \label{sec:kb}

\subsection{Construction of the Knowledge Base}

The knowledge base (KB) used in our experiments consists of 70 entries of SciML techniques. While it is generally advantageous to include \textit{only} the entries relevant to a specific problem context, we keep all entries in the KB throughout our experiments to test the retriever agent’s robustness, even though many entries are not directly relevant and can potentially lead to agents' confusion.

Each KB entry is constructed by synthesizing information from both research papers and their corresponding publicly available code on GitHub. The curation process is semi-automated using an LLM agent (Claude). Human researchers first identify relevant papers and code, and the agent then extracts key information including:

\begin{itemize}
    \item \textbf{Problem Setup:} The specific scientific computing problem being addressed, including equation types, boundary conditions, and domain characteristics.
    \item \textbf{Issues addressed:} Concrete symptoms and challenges the method mitigates (e.g., slow convergence, oscillations, discontinuities, noise sensitivity), so that agents can match methods to observed training difficulties.
    \item \textbf{Core method:} The key algorithmic or architectural innovation.
    \item \textbf{Implementation:} Annotated code snippets on the model architecture, training loop, and other algorithmic components.
    \item \textbf{Critical parameters:} Hyperparameters and design choices.
\end{itemize}

If a paper discussed multiple applications, all applications are included in the knowledge base as separate entries under the same method.

All entries are stored as structured Markdown files. A summary index in JSON format holds all the method names, descriptions, and file paths. The retriever agent will search through this index to identify relevant methods based on the problem context and training issues reported by other agents.

\begin{figure}[h!]
    \centering
    \includegraphics[width=0.95\linewidth]{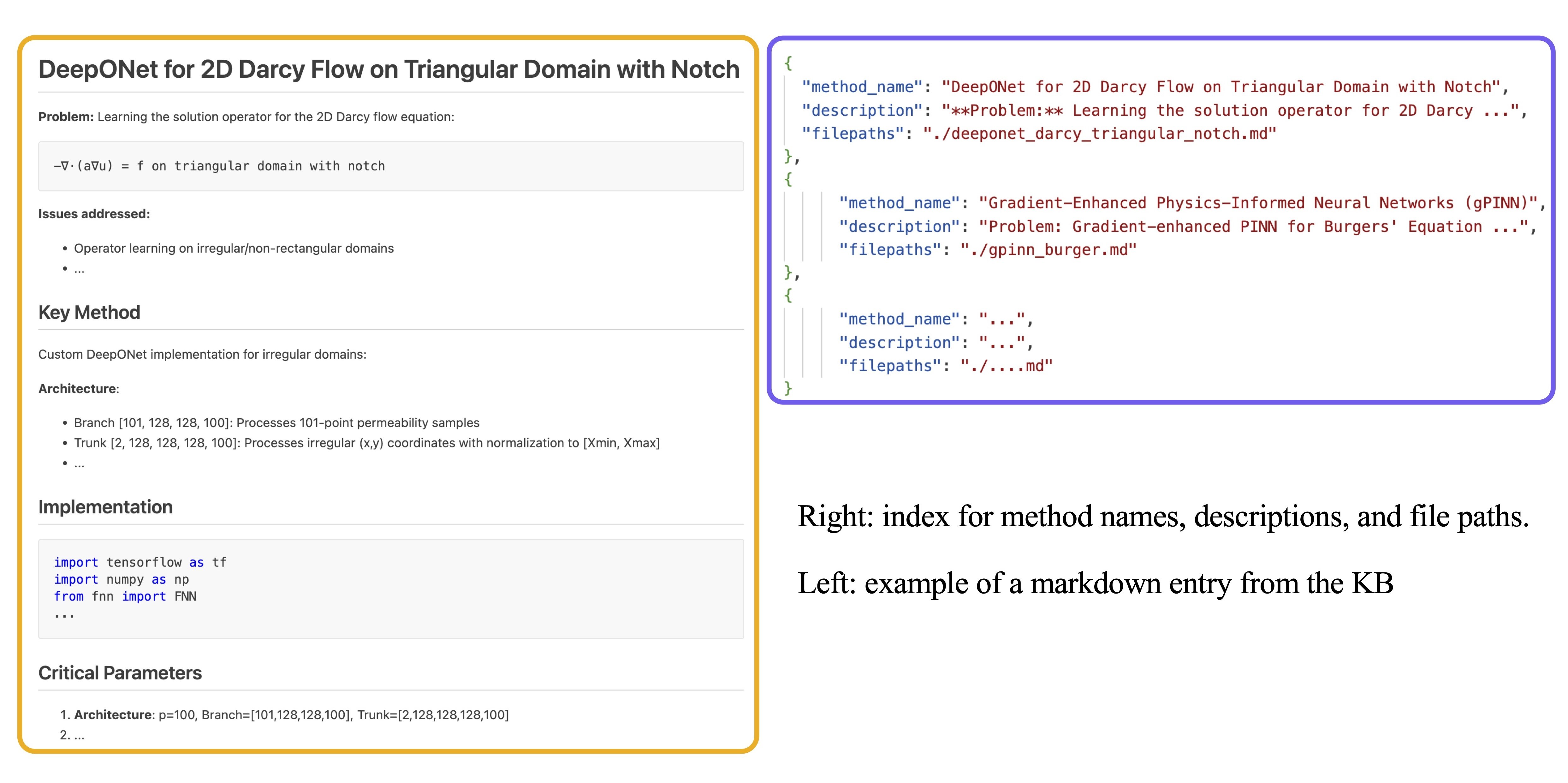}
    \caption{Example knowledge base entry and the JSON index summarizing method descriptions and file paths.}
    \label{fig:kb_example}
\end{figure}

An example knowledge base entry is provided in Figure~\ref{fig:kb_example}.

{\color{blue}
\subsection{Ablation Studies of the Knowledge Base}
\label{sec:kb_ablation}

We evaluate KB sensitivity on one controlled benchmark: discontinuous function approximation (Supplementary Section~\ref{sec:func_approx}). The objective is to test whether performance depends on retrieving relevant KB entries, and how the multi-agent reasoning trajectory changes when KB support is removed or made irrelevant.

Methodology.
We ran three settings with all non-KB factors fixed:
\begin{itemize}
    \item Full KB: standard retrieval from the 70-entry KB.
    \item No KB: retriever disabled (no KB context provided to proposer/critic).
    \item Random KB: retriever returns a random KB entry each mutation.
\end{itemize}
All three runs use the same root solution (\texttt{solution\_0}), the same evaluation contract (\texttt{evaluate.py}, \texttt{guidelines.md}), the same agent configuration in Table~\ref{tab:func_approx_config}, and the same evolutionary parameters (6 iterations, mutation batch size 4). For this benchmark, training is done on \texttt{train\_data.npz}; the score is always computed by the fixed evaluator on held-out \texttt{val\_data.npz}. All three settings start from the same root MSE, 0.2828776240.

\begin{table}[h!]
\centering
\color{blue}
\caption{KB ablation results for discontinuous function approximation (lower is better).}
\label{tab:kb_ablation_scores}
\small
\begin{tabular}{lcc}
\toprule
Setting & Best solution & Best MSE \\
\midrule
Full KB & \texttt{solution\_0100} & 0.0014631104 \\
No KB & \texttt{solution\_00101} & 0.0034207932 \\
Random KB & \texttt{solution\_00002} & 0.0303153563 \\
\bottomrule
\end{tabular}
\end{table}

Relative to Full KB, No-KB is \(2.338\times\) higher error and Random-KB is \(20.720\times\) higher error.

\begin{table}[h!]
\centering
\color{blue}
\caption{Round-3 proposer trajectory comparison across KB settings (function approximation only).}
\label{tab:kb_ablation_trajectory}
\small
\begin{tabular}{p{0.14\linewidth}p{0.23\linewidth}p{0.57\linewidth}}
\toprule
Setting & Best lineage & Observed method trajectory \\
\midrule
Full KB & \texttt{0 \(\rightarrow\) 01 \(\rightarrow\) 010 \(\rightarrow\) 0100} &
Trajectory converges to a stable MoE strategy: feature-augmented adaptive activations (\texttt{01}) \(\rightarrow\) explicit two-expert gating (\texttt{010}) \(\rightarrow\) bounded trainable gate sharpness and interface-focused weighting (\texttt{0100}). Retrieved entries are directly relevant, and later rounds refine rather than replace the core mechanism. \\
\midrule
No KB & \texttt{0 \(\rightarrow\) 00 \(\rightarrow\) 001 \(\rightarrow\) 0010 \(\rightarrow\) 00101} &
Agents still discover domain-decomposition and gating ideas from internal reasoning, but with more trial-and-error pivots: soft temperature gate \(\rightarrow\) hard detached gate \(\rightarrow\) Gumbel-softmax routing \(\rightarrow\) rollback to simpler asymmetric experts with a stable sigmoid gate and explicit jump term. Improvement is substantial, but less efficient and less stable than Full KB. \\
\midrule
Random KB & \texttt{0 \(\rightarrow\) 00 \(\rightarrow\) 000 \(\rightarrow\) 0000 \(\rightarrow\) 00002} &
Retrieved papers are frequently off-task (e.g., PINN/DeepONet/GNOT/ev-NSFnet entries from different PDE/operator regimes). The trajectory is less coherent: continuity-heavy and auxiliary-complex detours are introduced, then partially removed. Final recovery comes from simplifying back to a cleaner MoE-like formulation, but quality remains much worse than Full KB. \\
\bottomrule
\end{tabular}
\end{table}

These results indicate that success is not purely dependent on retrieving one exact ``right'' paper: even without KB, the system still iteratively improves candidate solutions from \(0.2829\) to \(0.00342\) through multi-agent reasoning alone, although the final result is less optimal than Full KB. At the same time, performance is strongly correlated with KB relevance. Relevant retrieval helps the proposal-critic loop converge to a coherent method and best score, while random/irrelevant retrieval degrades outcome quality and can misguide agent reasoning toward less relevant design choices.
}

\subsection{List of Knowledge Base Entries}
Table~\ref{tab:kb_entries} provides a comprehensive overview of all KB entries.

\begin{longtable}{p{0.05\linewidth}p{0.20\linewidth}p{0.55\linewidth}p{0.12\linewidth}}
\caption{Knowledge Base Entries: 70 Scientific Machine Learning Methods} \label{tab:kb_entries} \\
\toprule
\textbf{\#} & \textbf{Method} & \textbf{Description} & \textbf{Ref.} \\
\midrule
\endfirsthead

\multicolumn{4}{c}{\tablename\ \thetable\ -- Continued from previous page} \\
\toprule
\textbf{\#} & \textbf{Method} & \textbf{Description} & \textbf{Ref.} \\
\midrule
\endhead

\midrule
\multicolumn{4}{r}{Continued on next page} \\
\endfoot

\bottomrule
\endlastfoot

1 & Adaptive Act. & Adaptive activation functions for Burgers equation & \cite{jagtap2020adaptive} \\
2 & Local Adaptive Act. & Locally adaptive activation with slope recovery & \cite{jagtap2020locally} \\
3 & DeepONet & Antiderivative operator learning & \cite{lu2021learning,toscano2024deeponet} \\
4 & PI-DeepONet & Physics-informed antiderivative operator learning & \cite{wang2021learning,toscano2024deeponet} \\
5 & PI-DeepONet & Diffusion-reaction systems & \cite{wang2021learning,toscano2024deeponet} \\
6 & MoE-PINN & Burgers equation with mixture-of-experts & \cite{bischof2022mixture} \\
7 & MoE-PINN & Poisson equation on L-shaped domain & \cite{bischof2022mixture} \\
8 & Self-Adaptive PINN & Allen-Cahn equation & \cite{mcclenny2023self} \\
9 & Self-Adaptive PINN & Burgers equation & \cite{mcclenny2023self} \\
10 & Self-Adaptive PINN & Helmholtz equation & \cite{mcclenny2023self} \\
11 & PINN & Burgers equation forward problem & \cite{raissi2019physics} \\
12 & PINN & Burgers equation inverse problem & \cite{raissi2019physics} \\
13 & PINN & Diffusion equation with source term & \cite{raissi2019physics} \\
14 & PINN & Simple ordinary differential equation & \cite{raissi2019physics} \\
15 & PINN & 1D Poisson equation with Dirichlet BC & \cite{raissi2019physics} \\
16 & hp-VPINN & Advection-diffusion inverse problem & \cite{kharazmi2021hp} \\
17 & hp-VPINN & 1D Poisson equation & \cite{kharazmi2021hp} \\
18 & hp-VPINN & 2D Poisson equation & \cite{kharazmi2021hp} \\
19 & hp-VPINN & 2D Poisson equation on L-shaped domain & \cite{kharazmi2021hp} \\
20 & gPINN & Diffusion-reaction equation & \cite{yu2022gradient} \\
21 & Delta-PINN & Complex geometries & \cite{costabal2024delta} \\
22 & gPINN & Allen-Cahn equation with adaptive refinement & \cite{yu2022gradient} \\
23 & gPINN & Burgers' equation & \cite{yu2022gradient} \\
24 & gPINN & Inverse diffusion-reaction system & \cite{yu2022gradient} \\
25 & gPINN & Function approximation & \cite{yu2022gradient} \\
26 & gPINN & 1D Poisson equation & \cite{yu2022gradient} \\
27 & gPINN & 2D Poisson equation & \cite{yu2022gradient} \\
28 & NSFnet & Incompressible Navier-Stokes equations & \cite{jin2021nsfnets} \\
29 & ev-NSFnet & Entropy-viscosity regularized NSFnet & \cite{jin2021nsfnets} \\
30 & hPINN & Inverse design with hard constraints & \cite{lu2021physics} \\
31 & hPINN & Stokes flow topology optimization & \cite{lu2021physics} \\
32 & DeepONet & 2D advection equations & \cite{lu2022comprehensive,lu2021learning} \\
33 & FNO & 2D advection equations & \cite{lu2022comprehensive,li2020fourier} \\
34 & DeepONet & 1D Burgers' equation & \cite{lu2022comprehensive,lu2021learning} \\
35 & FNO & 1D Burgers' equation & \cite{lu2022comprehensive,li2020fourier} \\
36 & DeepONet & 2D Darcy flow with piecewise constant permeability & \cite{lu2022comprehensive,lu2021learning} \\
37 & FNO & 2D Darcy flow & \cite{lu2022comprehensive,li2020fourier} \\
38 & DeepONet & 2D Darcy flow on triangular domain with notch & \cite{lu2022comprehensive,lu2021learning} \\
39 & FNO & Irregular geometries & \cite{lu2022comprehensive,li2020fourier} \\
40 & POD-DeepONet & Instability wave propagation & \cite{lu2022comprehensive,lu2021learning} \\
41 & FNO & Wave propagation with non-uniform grids & \cite{lu2022comprehensive,li2020fourier} \\
42 & ICON & In-context operator networks & \cite{yang2023context} \\
43 & U-FNO & Gas saturation in CO2 storage & \cite{wen2022u} \\
44 & U-FNO & Pressure buildup in CO2 storage & \cite{wen2022u} \\
45 & CNO & Allen-Cahn equation & \cite{raonic2023convolutional} \\
46 & Fourier-DeepONet & Full waveform inversion (CurveFault-A) & \cite{zhu2023fourier} \\
47 & Fourier-DeepONet & Full waveform inversion (CurveVel-A) & \cite{zhu2023fourier} \\
48 & Fourier-DeepONet & Full waveform inversion (FlatVel-B) & \cite{zhu2023fourier} \\
49 & Fourier-DeepONet & Full waveform inversion (Style-A) & \cite{zhu2023fourier} \\
50 & CNO & Smooth transport equation & \cite{raonic2023convolutional} \\
51 & CNO & Darcy flow & \cite{raonic2023convolutional} \\
52 & CNO & Discontinuous transport equation & \cite{raonic2023convolutional} \\
53 & CNO & Compressible Euler equations (airfoil) & \cite{raonic2023convolutional} \\
54 & CNO & Navier-Stokes shear layer & \cite{raonic2023convolutional} \\
55 & CNO & Wave equation & \cite{raonic2023convolutional} \\
56 & CNO-FM & Foundation model for multiple PDEs & \cite{raonic2023convolutional} \\
57 & Oformer & Compressible flow around airfoils & \cite{li2022transformer} \\
58 & Oformer & Burgers' equation & \cite{li2022transformer} \\
59 & Oformer & Darcy flow & \cite{li2022transformer} \\
60 & Oformer & Electrostatics Poisson equation & \cite{li2022transformer} \\
61 & Oformer & Magnetostatics Poisson equation & \cite{li2022transformer} \\
62 & Oformer & 2D Navier-Stokes equations & \cite{li2022transformer} \\
63 & GNOT & 2D Darcy flow & \cite{hao2023gnot} \\
64 & GNOT & 2D elasticity & \cite{hao2023gnot} \\
65 & GNOT & 2D multi-scale heat conduction & \cite{hao2023gnot} \\
66 & GNOT & 3D multi-physics heatsink & \cite{hao2023gnot} \\
67 & GNOT & 2D electromagnetic inductor & \cite{hao2023gnot} \\
68 & GNOT & 2D transonic airfoil flow & \cite{hao2023gnot} \\
69 & GNOT & 2D Navier-Stokes steady-state & \cite{hao2023gnot} \\
70 & GNOT & 2D time-dependent Navier-Stokes & \cite{hao2023gnot} \\

\end{longtable}

\newpage
\section{Agent Prompts}

This section provides the core system prompts for each agent in the framework. Full prompt details will be available in the code repository.

\subsection{Data analyst agent}

\begin{verbatim}
ROLE: Exploratory data analyst with multimodal vision (Gemini)

TASK: Generate Python code for EDA, execute it, analyze
visualizations, and produce text-only report.

KEY INSTRUCTIONS:
- Generate code focusing on:
  (1) Mathematical properties (singularities, discontinuities)
  (2) Data quality (outliers, distributions)
  (3) Solution-relevant insights (sharp gradients, sampling issues)
- Create matplotlib visualizations and save all plots
- Execute code with self-debugging loop (max 3 iterations)
- Analyze generated plots using vision capabilities
- Write text-only report (downstream agents cannot see images)
- Use NumPy/matplotlib/scipy only (no PyTorch/JAX)
\end{verbatim}

\subsection{Evaluator agent}

\begin{verbatim}
ROLE: Testing contract generator

TASK: Generate evaluate.py and guidelines.md defining solution
interface and evaluation protocol.

KEY INSTRUCTIONS:
- evaluate.py: Self-contained test script with get_test_data(),
  compute_score(), and main() functions
- guidelines.md: Engineering contract specifying MODEL class
  interface, checkpoint format, data shapes
- Make ALL technical decisions (checkpoint format, data shapes)
- Ensure 100% consistency between evaluate.py and guidelines.md
- Specify WHAT to implement, NOT HOW (no architecture suggestions)
- Do NOT provide validation dataset details to engineer
\end{verbatim}

\subsection{Retriever agent}

\begin{verbatim}
ROLE: Knowledge base retriever

TASK: Search 70 KB entries to identify 0-1 relevant technique
that can improve current solution.

KEY INSTRUCTIONS:
- Evaluate relevance based on problem description, parent's
  performance, and parent's code
- Check if technique is already correctly implemented (skip if yes)
- Consider testing results (avoid regularization if underfitting)
- Select AT MOST 1 entry that will help and is feasible
- Provide detailed reasoning for selection/non-selection
\end{verbatim}

\subsection{Proposer agent}

\begin{verbatim}
ROLE: Mutation proposal generator (4-round debate with critic)

TASK: Generate implementation-ready proposals through structured
debate (2 reasoning + 1 synthesis + 1 finalization rounds).

ROUNDS:
- Rounds 1-2 (Reasoning): Explore improvements by analyzing:
  (1) Parent's weaknesses from analysis report
  (2) KB entry techniques
  (3) Insights from relatives in Analysis Bank
  (4) selector's reasoning for choosing this parent
  Focus on diagnosing issues and proposing targeted fixes.

- Round 3 (Synthesis): Synthesize critic feedback into concrete
  mutation plan. Specify exact changes with hyperparameters.

- Round 4 (Finalization): Generate complete proposal with sections:
  Motivation, Core Changes, Implementation Plan, Expected Outcomes.

KEY INSTRUCTIONS:
- Consider data analysis report for training data characteristics
- Balance incremental improvements and novel approaches
- Respond to critic's feedback constructively
\end{verbatim}

\subsection{Critic agent}

\begin{verbatim}
ROLE: Proposal critic (3-round debate with proposer)

TASK: Critique proposals through structured debate
(2 reasoning + 1 plan critique rounds).

ROUNDS:
- Rounds 1-2 (Reasoning Critique): Evaluate proposal for:
  (1) Correctness (mathematical validity, feasibility)
  (2) Relevance (addresses parent's actual weaknesses)
  (3) Feasibility (compatible with existing code)
  (4) Novelty vs safety trade-off
  Challenge assumptions and identify potential issues.

- Round 3 (Plan Critique): Review synthesized plan for
  completeness. Ensure all implementation details specified.
  Verify consistency with requirements and testing contract.

KEY INSTRUCTIONS:
- Balance constructive criticism and reasonable exploration
- If parent selected for exploration, be permissive of risky ideas
\end{verbatim}

\subsection{Engineer agent}

\begin{verbatim}
ROLE: Solution code implementer

TASK: Implement solution.py following proposals and guidelines.

MODES:
- Root Solution (Phase 2): Generate complete solution.py
  implementing MODEL class from scratch. No parent/proposal.

- Child Solution (Phase 3): Implement proposal's changes on
  top of parent's code. Use parent as foundation, modify per
  proposal. Include all hyperparameters from proposal.

- Debugging (All phases): Fix errors based on debugger feedback.
  Preserve working components, only modify problematic sections.

KEY INSTRUCTIONS:
- Implement both --mode=validate (1 epoch) and --mode=train (full)
- Save checkpoint to ./MODEL_CHECKPOINT per guidelines
- Print training progress to stdout
- Follow guidelines exactly (API, data shapes, checkpoint format)
\end{verbatim}

\subsection{Debugger agent}

\begin{verbatim}
ROLE: Implementation error analyzer

TASK: Diagnose errors from failed validation/training runs and
provide debugging suggestions.

KEY INSTRUCTIONS:
- Analyze error traceback and error message
- Review problematic code sections
- Identify root cause (syntax, runtime, logic errors)
- Provide specific fix suggestions with code snippets
- Consider common pitfalls (shape mismatches, device issues, etc.)
- Keep fixes minimal and targeted to preserve working components
\end{verbatim}

\subsection{Result analyst agent}

\begin{verbatim}
ROLE: Performance analyzer with multimodal vision

TASK: Generate comprehensive analysis reports for completed solutions.

INPUTS:
- Solution code (complete, no truncation)
- Training log (complete, no truncation)
- Testing log (complete, no truncation)
- Final evaluation score
- Proposal markdown (for child solutions)
- Parent analysis (for child solutions)
- Plot images (if available)

REPORT STRUCTURE:
## Summary of Approach
- Model architecture/algorithm
- Key techniques and innovations (focus on novel, non-trivial)
- Hyperparameters (network size, loss weights, etc.)
- Training setup (epochs, batch size, etc.)
- Optimization details (optimizer, learning rate, scheduler)

## Training Dynamics
- Convergence behavior (fast/slow/stagnant)
- Loss components (which dominates)
- Training stability (stable/oscillating/diverging)
- Overfitting/underfitting behavior
- Numerical issues or anomalies

## Performance Breakdown
- Final score and component metrics
- Comparison with parent (for child solutions)
- Algorithmic/mathematical bottlenecks
- NOT hardware bottlenecks

## Plot Analysis (if plots provided)
- Describe what plots show
- Key observations and patterns
- How plots relate to performance

## Comparison with Parent (for child solutions)
- What changed from parent
- Why change helped or hurt performance
- Lessons learned

KEY INSTRUCTIONS:
- Be specific and technical (not generic praise)
- Focus on actionable insights for downstream mutations
- Use multimodal vision to analyze plots when available
\end{verbatim}

\subsection{Selector agent}

\begin{verbatim}
ROLE: Solution selector for ensemble-guided mutation

TASK: Select top-K solutions from candidate pool for next iteration.
Uses ensemble voting (GPT + Grok + Gemini).

SELECTION CRITERIA:
HIGH POTENTIAL (select):
- New architecture with moderate losses (may need tuning)
- Innovative techniques with fixable implementation flaws
- Strong theory but imbalanced loss terms (needs rebalancing)
- Novel approach not yet converged (needs more training)
- Underexplored branch (fewer children, unexplored search space)
- Good ideas with suboptimal hyperparameters (easy wins)

LOW POTENTIAL (avoid):
- Traditional architecture with performance plateau
- Repeatedly failed mutations with no clear improvement path
- Fundamental design flaws incompatible with requirements
- Overexplored branches (many children, likely exhausted)

EXPLOITATION VS EXPLORATION:
- Select 1-2 top performers for incremental refinement
- Select 1-2 interesting failures with fixable issues
- Balance is key for effective search

OUTPUT:
- Exactly K solution IDs
- 2-3 bullet points per selection explaining potential
- Explicit consideration of exploitation-exploration trade-off
\end{verbatim}

\newpage
\bibliographystyle{unsrt}  
\bibliography{references}

\end{document}

%% file: Introduction.tex
\section{Introduction}

Scientific Machine Learning (SciML) integrates data-driven learning with numerical simulation and physical modeling to solve problems in scientific computing that are analytically intractable or computationally expensive. Recent developments such as physics-informed neural networks (PINNs) \cite{raissi2019physics}, neural operators \cite{li2020fourier,lu2021learning}, and domain-decomposition methods \cite{jagtap2020conservative,meng2020ppinn} have enabled progress across fluid mechanics, inverse problems, materials modeling, biophysics, and geoscience. However, the construction of effective SciML models still requires substantial manual effort: practitioners must select architectures, enforce physical constraints, tune loss weightings, design sampling curricula, and determine proper optimization strategies. This iterative search is labor-intensive and problem-specific, forming a bottleneck for scalable scientific computing.

Recent LLM-based agent systems have motivated interest in automating portions of the SciML workflow \cite{ren2025towards,wei2025ai}. Notable examples include the \emph{Virtual Lab} \cite{swanson2025virtual}, in which domain-specialized agents collaborate to plan experiments and interpret results, and ChemCrow, an agent-based framework for molecular synthesis planning \cite{bran2023chemcrow}. These approaches demonstrate the utility of role-specialization and iterative refinement, but they focus primarily on managing experimental or model-selection workflows. More broadly, agentic systems have been applied to open-ended scientific discovery \cite{lu2024ai,gottweis2025towards,du2025accelerating,zhou2025autonomous}, mathematical exploration and algorithm synthesis \cite{georgiev2025mathematicalexplorationdiscoveryscale,romera2024mathematical,yu2025alpharesearch}, scientific equation discovery \cite{shojaee2024llm}, automated data science and code exploration \cite{ou2025automind,jiang2025aide}, numerical computation \cite{press2025algotune}, symbolic regression \cite{cranmer2020discovering,udrescu2020ai}, and expert-level scientific software development \cite{aygun2025ai,ma2024llm}, while verification and reliability have emerged as key challenges in such pipelines \cite{cornelio2025need,miao2025paper2agent}. In notable frameworks such as AlphaEvolve \cite{novikov2025alphaevolve} and ShinkaEvolve \cite{lange2025shinkaevolve}, evolutionary and program-synthesis approaches are employed to iteratively improve solutions, leading to potentially new scientific and practical discoveries. In parallel, for machine learning research, AutoML systems and neural architecture search (NAS) frameworks \cite{liu2018darts,real2019regularized} optimize hyperparameters or network topologies, with recent work extending this to agent-driven architecture discovery \cite{liu2025alphago}.  Within SciML specifically, recent LLM-based systems have been proposed for automated PINN construction and PDE surrogate modeling \cite{wuwu2025pinnsagent,he2025lang}. However, these systems do not generally produce \emph{new SciML modeling strategies}, such as novel PDE-constrained architectures, problem-specific loss formulations, or adaptive decomposition schemes. Instead, they typically search within a predetermined hypothesis space or rely on a fixed modeling paradigm.

In contrast, scientific computing often requires reasoning about model structure, discretization, and physics-informed inductive bias. The space of possible SciML solution strategies is therefore combinatorial and structured, involving architectural motifs, physics constraints, numerical solvers, and multi-objective optimization. A system capable of exploring this space must not only tune parameters but also \emph{compose and critique modeling ideas}.

To address this need, drawing inspiration from recent agentic systems for iterative scientific software refinement \cite{aygun2025ai,ma2024llm}, we introduce \textbf{AgenticSciML}, a collaborative multi-agent framework for the \emph{emergent discovery of SciML modeling strategies}. The framework coordinates multiple specialized agents---including proposers, critics, engineers, retrievers, and evaluators---that iteratively generate, analyze, and refine candidate SciML solutions. A persistent knowledge base stores prior solutions, ablation outcomes, error analyses, and textual reports. Retrieval agents surface relevant methodological precedents, allowing new strategies to build upon earlier ones. A structured debate process, grounded in evidence that multi-agent debate and ensemble voting improve decision quality \cite{choi2025debate}, encourages agents to justify, challenge, and revise modeling decisions, while an ensemble-guided evolutionary search balances exploitation of promising directions with exploration of alternative approaches.

Across benchmarks in PDE-constrained learning, operator learning, and inverse problems, AgenticSciML produces solution strategies that outperform single-agent LLM systems, sometimes by 1000$\times$,  automated PINN frameworks \cite{zhang2023auto}, and NAS/AutoML baselines. Moreover, the system discovers SciML strategies that do not appear in the knowledge base or in standard formulations, including (1) adaptive domain-decomposed PINNs for multi-scale PDEs, (2) physics-informed operator learning architectures with constraint-conditioned branches, and (3) dynamically weighted loss schedules derived from residual-flow structure. These results indicate that structured multi-agent collaboration can enable \emph{emergent innovation in SciML model construction}, beyond what can be achieved through direct prompting or search alone.

The remainder of this paper is organized as follows. In Section \ref{sec:methods} we describe our multi-agent architecture, including agent roles, knowledge retrieval mechanisms, and evolutionary search strategies. In Section \ref{sec:results} we detail experiment setups and findings across benchmark SciML problems. Finally, in Section \ref{sec:summary} we summarize our contributions and discuss future research directions.